\documentclass{article} %

\usepackage{hyperref}
\usepackage{url}

\usepackage{microtype}
\usepackage{graphicx}
\usepackage{subcaption}
\usepackage{booktabs} %
\usepackage[most]{tcolorbox}
\usepackage{xspace}
\tcbuselibrary{listings}
\usepackage{hyperref}
\usepackage{microtype}
\usepackage{hyperref}
\usepackage{url}
\usepackage{xurl}
\usepackage{amsmath}
\usepackage{amssymb}
\usepackage{mathtools}
\usepackage{amsthm}
\usepackage[most]{tcolorbox}
\usepackage{tabularx}
\usepackage{xcolor}
\usepackage{algorithm}
\usepackage{algpseudocode}
\usepackage{wrapfig,booktabs,tabularx}
\usepackage[table]{xcolor}
\usepackage[utf8]{inputenc}
\usepackage[english]{babel}
\usepackage[authoryear]{arxiv/natbib}
\usepackage{etoc}
\usepackage[T1]{fontenc}

\usepackage{amsmath,amsfonts,bm}

\def\eqref#1{equation~\ref{#1}}

\def\1{\bm{1}}

\DeclareMathAlphabet{\mathsfit}{\encodingdefault}{\sfdefault}{m}{sl}
\SetMathAlphabet{\mathsfit}{bold}{\encodingdefault}{\sfdefault}{bx}{n}

\theoremstyle{definition}

\theoremstyle{remark}

\usepackage{thmtools}
\usepackage{thm-restate}

\usepackage{listings}
\definecolor{ourblue}{HTML}{40aae3}
\usepackage{hyperref}

\definecolor{emerald}{RGB}{34,139,34} %

\usepackage{arxiv/ssArxiv}
\hypersetup{
    colorlinks,
    linkcolor=ourblue,
    citecolor=ourblue,
    urlcolor=ourblue
}

\newtcblisting{promptbox}[1][]{arc=0mm, outer arc=0mm, colback=gray!5, colframe=gray!50, fonttitle=\bfseries, left=1mm, right=1mm, top=1mm, bottom=1mm, listing only, breakable, listing options={numbers=none, frame=none, breaklines=true, basicstyle=\fontfamily{zi4}\selectfont\footnotesize\color{black}, columns=fullflexible, keepspaces=true, showstringspaces=false, literate={*}{{\raisebox{0.3ex}{*}}}1}, #1}
\newcommand{\transcripttag}[1]{\begingroup\setlength{\fboxsep}{2pt}\colorbox{blue!12}{\normalfont\ttfamily\footnotesize\color{black}\strut[#1]}\endgroup}
\newtcblisting{transcriptbox}[1][]{arc=0mm, outer arc=0mm, colback=gray!5, colframe=gray!50, fonttitle=\bfseries, left=1mm, right=1mm, top=1mm, bottom=1mm, listing only, breakable, listing options={moredelim={[il][\normalfont\ttfamily\bfseries]{(*B*)}}, escapeinside={(*@}{@*)}, moredelim={[il][\color{green!40!black}]{(*A*)}}, moredelim={[il][\color{red!70!black}]{(*D*)}}, numbers=none, frame=none, breaklines=true, basicstyle=\ttfamily\footnotesize\color{black}, columns=fullflexible, keepspaces=true, showstringspaces=false, literate={*}{{\raisebox{0.3ex}{*}}}1 {’}{{\textquoteright}}1 {“}{{\textquotedblleft}}1 {”}{{\textquotedblright}}1 {–}{{\textendash}}1 {—}{{\textemdash}}1 {−}{{$-$}}1 {→}{{$\rightarrow$}}1 {×}{{$\times$}}1}, #1}

\title{Do LLM Agents Still Need to Be Told How to Self-Improve?}

\author{
  Shobhita Sundaram$^1$\thanks{Correspondence: \texttt{shobhita@mit.edu}} ,
  Cansu Sancaktar $^{2}$,
  Phillip Isola $^1$ \\
  $^1$ MIT CSAIL,  $^2$ University of T\"ubingen \\
\\
}

\begin{document}

 \maketitle

\maketitle
\vspace{-25pt}
\begin{abstract}
LLM agents increasingly drive their own learning. Self-improvement methods delegate various components of model optimization to the models themselves.
As models become more capable, this motivates a basic question: do these increasingly capable models still need us to specify how they should self-improve?
In this paper we study an autonomous paradigm in which, given basic primitives, resources, and a general goal for self-improvement, agents construct their own self-learning algorithm for context-based optimization. 
Using the setting of gameplay as a testbed, a frozen LLM agent is given a compute budget and tools for collecting experience and updating text documents, and allowed to freely experiment without an externally imposed loop or access to real-world data. We find that sufficiently capable models 
do self-improve under these conditions, demonstrating strategies that align with classic cognitive frameworks for self-regulated learning and constructs from machine learning literature on self-generated curricula. Furthermore, we study the extent to which the learned text artifacts transfer to external benchmarks, observing improvement across different game horizons, mechanics, and 3D navigation.
\textbf{Project page:} \url{https://ssundaram21.github.io/self-org/}.
\end{abstract}

\vspace{-10pt}
\begin{figure*}[h]
    \centering
    \includegraphics[width=0.95\linewidth]{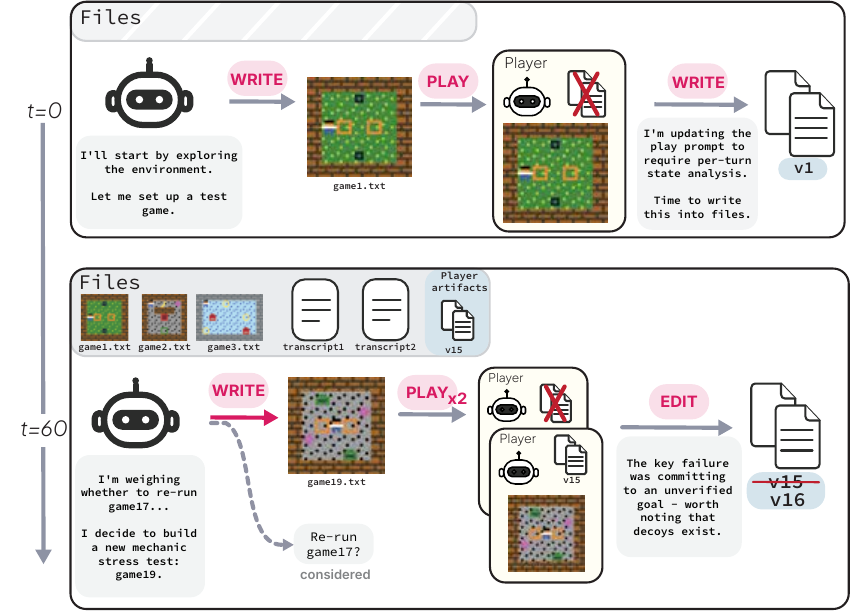}
    \caption{\small \textbf{Do capable models specify their own self-improvement?} We give a frozen LLM (here, \texttt{Claude Opus-5}) the general goal of improving itself at gameplay, and place it in a constrained workspace. We do not provide real games to practice on, or prescribe a particular self-improvement loop. We instead provide a toolset \texttt{\{Read, Write, Bash, Gameplay\}}, and allow the teacher to generate its own games and organize its own learning procedure. 
    At the start ($t=0$) the teacher uses its tools to build a test game, spawn a player session to play that game, analyze the result, and then update the player context (e.g., skill files). Later ($t=60$), the teacher decides to write an adversarial game and observe the player attempt it with and without the optimized context. The teacher then uses the observed failure mode to make another context update.    }\label{fig:teaser}
\end{figure*}

\section{Introduction}

Imagine giving an agent a computer, and asking it to improve its ability to play puzzle games. What will it do? The computer has no pre-installed games; the agent must write its own games if it wants to practice. It must decide what sorts of games to try, which lessons to retain, and how to measure its progress. 

These decisions concerning the design of learning procedures are typically made by human practitioners. Yet as LLM agent capabilities improve, we increasingly delegate the machine learning process to the learners themselves. 
Various works leverage models to generate supervised data, construct interactive environments, and even rewrite their own tools and harnesses \citep{zhang2026darwin,Liu2026SPADESI,Zelikman2022STaRBR,Wang2022SelfInstructAL}. 
In parallel with efforts to develop better methods, there are growing concerns about the need for safeguards against misaligned self-improvement. Both directions motivate a basic question: 

\begin{center}
    \textit{Do capable models still need us to specify how they should self-improve?}
\end{center}

We study the natural behavior of an LLM agent when provided with the primitive tools and resources for self-improvement via context manipulation, but without being wrapped in an explicit self-improvement algorithm. 
Rather than designing a loop for experience generation and learner updates, we allow the agent to compose actions into a self-learning procedure. The agent does not have access to real-world tasks or validation scores. It must experiment with different ways of generating tasks, reflecting on failures, and assessing progress in a constrained sandbox. 
Our setting is important for understanding both \textit{capability}, where models may no longer require prescribed self-optimization algorithms, and \textit{safety}, where the absence of a specified loop does not necessarily prevent self-improvement.

We find that \textit{sufficiently capable models successfully self-organize learning procedures and self-improve on held-out tasks.} We examine this phenomenon in the setting of gameplay with limited information. A \textit{player} agent plays turn-based games with undisclosed mechanics under strict compute and game-horizon budgets. This task requires diverse reasoning skills including efficient exploration, information acquisition, multi-step planning, and world modeling. 
An instance of that same agent serves as the \textit{teacher}. The teacher is given a general goal to improve the player through text (e.g., editing the prompt, creating skill files), a compute budget, and a set of tools for observing the player attempt self-generated games and editing player context. We do not instruct the teacher when to use the tools; the teacher freely experiments with the player. 

Across the set of effective self-learners, models from different families employ similar recurring, recognizable learning strategies, such as generating games with calibrated difficulty, running controlled A/B comparisons, and developing validation suites to measure progress. These strategies align to behaviors described by social-cognitive frameworks for self-regulated learning \citep{zimmerman2000attaining,zimmermanBecoming}. We also find that agents spend more effort on harder tasks, resembling learnability objectives in asymmetric self-play \citep{vygotsky1978mind, wang2022in, rutherford2024no}. Furthermore, the self-constructed learning procedures generalize: Learned text files transfer across multiple benchmarks that require reasoning over different game horizons, engines, and representations. We study how providing information on evaluation settings to the teacher, via context or environment constraints, shapes this downstream generalization. 

Our main findings are:
\vspace{-5pt}
\begin{enumerate}
    \item Given just primitive tools for self-improvement via context manipulation, a sufficiently capable agent constructs an effective learning procedure that improves its own performance on held-out gameplay. 
    \item Models from different families implement recognizable strategies that align to self-regulated learning frameworks, and machine learning literature on self-generated curricula.
    \item The learned text generalizes across game benchmarks and representations. Evaluation-setting awareness biases the teacher towards relevant tasks, however can encourage overfitting. 
\end{enumerate}
\vspace{-7pt}

\section{Preliminaries: Gameplay with limited information.}\label{sec:gameplay}

Our goal is to study agent self-improvement in interactive settings requiring multi-step reasoning. As a testbed, we use the setting of gameplay with undisclosed mechanics. The agent plays a turn-based game with limited knowledge of the rules and objectives. To succeed it must efficiently acquire information, plan, and execute within the given horizon. This setting exercises diverse capabilities, including adaptive efficiency, goal discovery, planning, and world modeling \citep{foundation2026arc}.

We consider a partially observable Markov decision process setting. For each game $g$, let $s_t \in \mathcal{S}$ denote the state at turn $t$, $o_t \in \mathcal{O}$ the observation received by the player, and $a_t \in \mathcal{A}$ the player's action. The game engine determines the transition dynamics $s_{t+1} \sim P_g(\cdot \mid s_t, a_t)$ and corresponding observation $o_{t+1} \sim O_g(\cdot \mid s_{t+1})$. The reward is binary, $r_t = R_g(s_t, a_t, s_{t+1}) \in \{0,1\}$; the player receives reward $1$ when reaching a winning state and $0$ otherwise. The episode terminates upon success or after at most $H$ turns.
The player $\pi_\theta$ does not have access to $P_g$ or $R_g$ and infers the game mechanics through interaction. It acts based on its interaction history $h_t = (o_1, a_1, \ldots, o_t)$ and context $c$ (e.g., a prompt), with $a_t \sim \pi_\theta(\cdot \mid h_t, c)$.

\section{A controlled self-improvement setting}\label{sec:setting}

\begin{wrapfigure}[10]{r}{0.46\linewidth}
\vspace{-2.4\baselineskip}
\begin{minipage}{\linewidth}
\begin{algorithm}[H]
\caption{Self-organized self-improvement}
\label{alg:autonomous-context-self-play}
\small
\begin{algorithmic}[1]
\Require Budget $B$, teacher context $I$
\State $S \gets \textsc{InitSandbox}(c_0, I, B)$
\State $\mathcal{A}_T \gets \{\textsc{Read}, \textsc{Write}, \textsc{Bash}, \textsc{Gameplay}\}$
\While{$B > 0$}
    \State $a \sim \pi_{teacher}(\cdot \mid S)$, \ $a \in \mathcal{A}_T$
    \State $S \gets \textsc{Execute}(S, a)$
\EndWhile
\State \Return player context $c_{final}$
\end{algorithmic}
\end{algorithm}
\end{minipage}
\vspace{-15pt}
\end{wrapfigure}
Do LLM agents, given primitive tools for text editing and experience generation, compose those tools into a useful self-optimization procedure? And can this be achieved in a \textit{data-free} setting with no access to human-written games? We study these questions in a controlled environment: A frozen model (the \textbf{\textit{teacher}}) is given a budget, a prompt, and a limited set of tools, and allowed to experiment freely. The teacher is given a general goal to optimize text (e.g., a prompt, skill files) that improves a \textbf{\textit{player}}'s performance on unseen games, where the player is another instance of the same model. We measure success through performance on held-out game benchmarks.

Our setting has notable similarities to asymmetric self-play, in which a teacher generates tasks for a student to solve \citep{sukhbaatar2017asymmetric}. When applied to LLMs the same model often takes on both roles \citep{zhao2025absolute}. These methods prescribe procedures for sampling tasks and rollouts, evaluating task learning value, and updating the student and teacher. In contrast we do not mandate how the teacher must use its budget and tools, and optimize in text-space rather than weight-space. The teacher's role extends beyond task generation and encompasses constructing the learning procedure itself. 
\vspace{-0.5em}
\subsection{Teacher-Player Setup}\label{sec:teacher-player-setup}

Let $\pi$ denote an LLM with frozen weights $\theta$. The LLM is instantiated as a \textbf{player agent} $\pi_{player}$ that has context $c=(c_P, c_L)$ where $c_P$ is the player prompt, and $c_L$ is auxiliary text files (e.g., skill files). The player policy induced by $c$ is then $\pi_{player}(a_{t} \mid h_t, c)$ for a game trajectory $h_t$. Another instance of the same LLM, the \textbf{teacher agent} $\pi_{teacher}$, modifies $c$. We define an \textbf{update} as \textit{a single edit tool-call to one file in $c$}. Let $c_0$ be the initial 
context state (e.g., a minimal prompt) and $c_{final}$ be the state of $c$ after the final teacher update.

Context optimization allows us to study closed-source models, but also has useful properties. Gradient-based learning is in continuous space and tied to predefined rewards. In text optimization, the teacher can reflect on player rollouts and dynamically decide to generate tasks,  experience, or knowledge updates. Those updates being in text preserves a readable explanation for changes in player behavior.

\subsection{Teacher environment}\label{sec:teacher-env}
The teacher is an agent in a persistent, sandboxed harness session (Claude Code, Pi). 
The harness is minimal; we disable the default system prompt, automatic memory, web search, subagents and other such features (Appendix \ref{sec:app-teacher-env}). The teacher has access only to the sandbox files and features that we explicitly provide: context, a specified toolset, and a budget.

\textbf{Tools.} The teacher is granted \textbf{Read/Write} tools for sandbox files, \textbf{Bash} for terminal commands (subject to sandbox constraints), and a \textbf{Gameplay} tool. The Gameplay tool allows the teacher to observe a player play a game with teacher-granted context. The teacher calls the tool with a game source file and, optionally, a path to a directory holding player text artifacts. The tool executes the game in a fresh, external player session, and returns to the teacher the player score and transcript.

In our main experiments the Gameplay tool uses \textit{PuzzleScript}, a domain-specific language for 2D grid-based puzzle games. PuzzleScript admits a diverse set of games that share basic structures and action spaces, enabling a controlled setting for studying generalization. The teacher issues games as PuzzleScript text files; the Gameplay tool rejects games that do not compile or where the win state is verifiably unreachable. We ablate the PuzzleScript constraint in experiments that instead use a Python-based Gameplay tool.

These tools can serve as a mechanism for giving the teacher information. For instance, we can allow the teacher to specify the game horizon (turn budget) in the Gameplay tool, or impose such a budget ourselves --- effectively communicating the evaluation game horizon via the constraint. 
\begin{figure*}[t]
    \centering
    \includegraphics[width=0.9\linewidth]{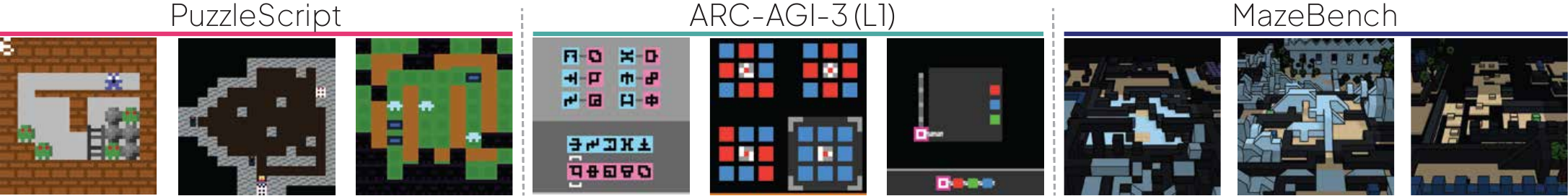}
    \caption{\small \textbf{Examples of evaluation benchmarks.} Sample games from our transfer benchmark, ordered in increasing distance from our canonical self-optimization setting: (1) \textsc{PuzzleScript} (same game engine, different horizon splits), (2) \textsc{ARC-AGI-3} (different engine and representation), (3) \textsc{MazeBench} (3D navigation and mechanics in isometric ASCII projections).}\label{fig:game-examples}
    \vspace{-5pt}
\end{figure*}
 
\textbf{Context.} 
The teacher's system prompt specifies a \textit{main goal}: to improve a player's performance on  single-player games by writing text artifacts. We also specify a \textit{generalizability goal}: the artifacts should be useful for unseen games with different tasks or mechanics. We do not guide how to structure the artifacts, instead allowing the teacher to freely create and modify player text files, including the prompt. When granted the PuzzleScript Gameplay tool, the teacher also receives official PuzzleScript documentation. Full prompts are in Appendix \ref{sec:app-prompts}.

\textbf{Budget.} The teacher has a two-part resource budget. The \textit{update budget} is the number of updates that may be applied to the player context $c$, designated as a directory in the teacher sandbox. 
Each edit tool-call to $c$ is one update. The \textit{gameplay budget} is the number of gameplays that the teacher may observe.
The teacher session ends when either budget is exhausted. If the teacher ends its action sequence without consuming either budget fully, we prompt it to continue.

\subsection{Measuring self-improvement through generalization}\label{sec:measuring}
We assess if an agent has structured its learning well via transfer to held-out benchmarks. 
Given player context $c$, we evaluate that context by how well the induced player policy performs on an unseen test set $\mathcal{D}_{test}$. For each game, we instantiate a fresh player session with $c$ and rollout the game at a fixed game horizon $H$ (turn budget) and output token budget. The game ends at turn $T \leq H$ when (1) the player achieves the objective, (2) the turn budget is exhausted, or (3) the player's output tokens (summed over turns so far) exceeds the game token budget. The reward on $\mathcal{D}_{test}$ is
\vspace{-2pt}
\begin{equation}
    J(c) \;=\; \frac{1}{|\mathcal{D}_{\text{test}}|}\sum_{g \in \mathcal{D}_{\text{test}}}\;
                  \left[\mathbb{E}_{\pi_{player}(\cdot \mid h_t, c)}\;\bigl[\sum_{t=1}^T r_t\bigr]\right].
  \label{eq:objective}
\end{equation}
In practice we estimate the inner expectation by averaging the rewards of independent Monte-Carlo rollouts (Table \ref{tab:app-eval-settings}).
The policy is evaluated across a distribution of games rather than initial states of a single game environment.
We measure self-improvement as the difference in player performance with and without the teacher-optimized context: $J(c_{final})-J(c_0) > 0$.

\begin{figure}[b]
    \centering
    \begin{minipage}{0.44\linewidth}
        \centering
        \includegraphics[width=\linewidth]{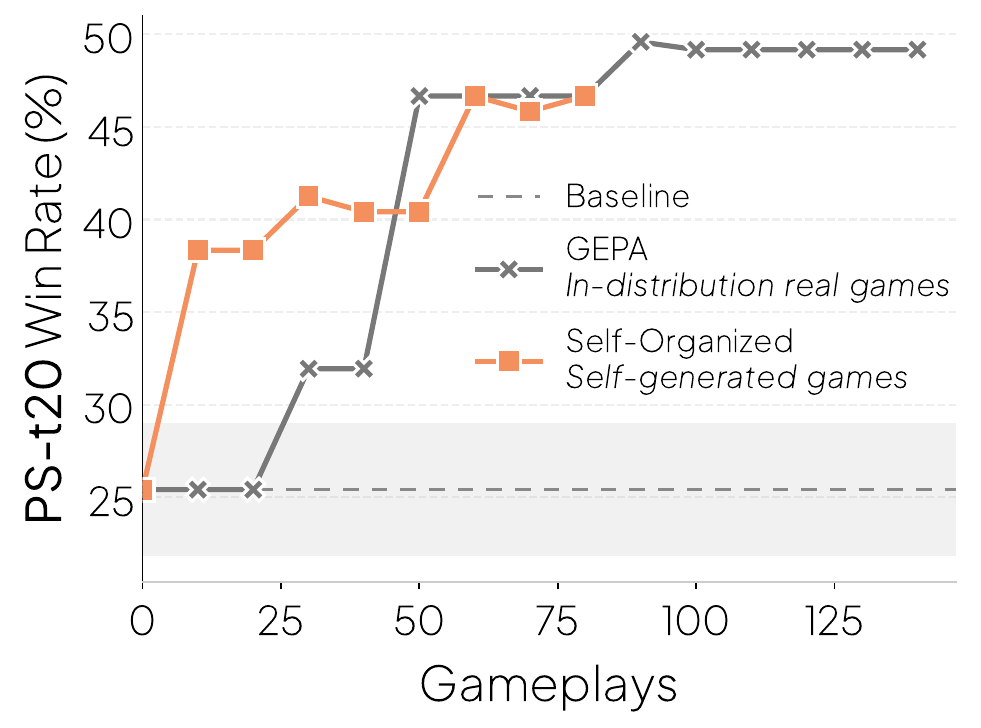}
    \end{minipage}%
    \hfill
    \begin{minipage}{0.47\linewidth}
        \caption{\small
            \textbf{Motivating observation: Self-organized learning with self-generated games achieves comparable gains to prescribed optimization on real games.} \texttt{Claude Opus-5} optimizes text artifacts that achieve similar performance on \textsc{PuzzleScript-t20}  as GEPA \citep{agrawal2026gepa} optimizing over 20 in-distribution real games. Results are averaged over 3 seeds; individual seeds are shown in Figure \ref{fig:gepa-seeds}.
        }
        \label{fig:gepa}
    \end{minipage}
    \vspace{-10pt}
\end{figure}

\subsection{Experiment Setup.}\label{sec:experiment-setup}
We assess six models across three families: \texttt{Claude Opus-\{4.6, 5\}}, \texttt{GPT \{5.6-Sol, 6-Astra\}}, and \texttt{GLM-\{4.7, 5.3\}}. In all experiments both the teacher and player use low reasoning effort and per-turn 20k token budgets. We run three independent teacher sessions per model with an update budget of 20, gameplay budget of 150, and toolset of \texttt{\{Read/Write, Bash, Gameplay\}}. In our canonical setting the Gameplay tool plays PuzzleScript games with a 20-turn horizon and 25k full-game token budget. We ablate these constraints and experiment with Python-based Gameplay in Section \ref{sec:full-transfer}. Our external evaluation benchmarks are described below, with examples in Figure \ref{fig:game-examples} and construction details in Appendix \ref{sec:app-datasets}. 

\textbf{\textsc{PuzzleScript-t20} (PS-t20).} We draw a subset of 20 short-horizon game levels (optimal solution length $\leq$12) from a pool of human-authored PuzzleScript games \citep{Earle2025PuzzleJAXAB}. We evaluate with a 20-turn horizon and 25k token-budget, thus sharing constraints with the main self-learning setting. We next test transfer to benchmarks that share progressively fewer attributes.

\textbf{\textsc{PuzzleScript-t30} (PS-t30).} We draw 20 longer-horizon game levels (solution length 13-22) from the same base pool as \textsc{PuzzleScript-t20}, and evaluate with a 30-turn horizon and 31k token-budget. This retains the shared game engine but tests generalization to longer game horizons. 

\textbf{\textsc{ARC-AGI-3 (L1)}.} We test transfer to different 2D game engines by evaluating on Level 1 of games in ARC-AGI-3 \citep{foundation2026arc} (30-turn horizon, 31k token-budget).

\textbf{\textsc{MazeBench.}} MazeBench is a 3D open world in which an agent navigates rooms with obstacle challenges to retrieve gems \citep{pappas2026mazebench}. We adopt the benchmark's room-completion criteria,  and draw a random set of 16 rooms solvable within 50 turns. We evaluate with a 100-turn horizon and 75k token-budget. This benchmark tests generalization to longer horizons, 3D navigation via isometric-style ASCII projections, and 3D mechanics.

\section{What are the strategies of successful self-improvers?}\label{sec:strategies}

\begin{figure*}[t]
    \centering
    \includegraphics[width=0.9\linewidth]{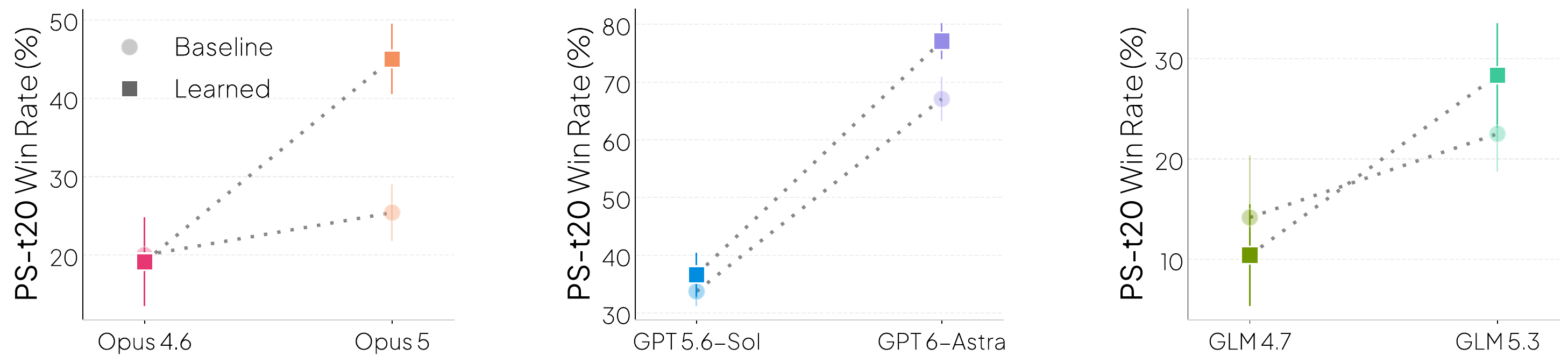}
    \caption{\small \textbf{Recent models self-improve.} Results on \textsc{PuzzleScript-t20} for two models in each family. Newer generations self-improve over the baseline while older generations fail to do so. Error bars are $\pm1$ SD over 3 seeds.}\label{fig:generation-results}
    \label{fig:zimmerman}
\end{figure*}

\begin{figure*}[t]
    \centering
\includegraphics[width=1.0\linewidth]{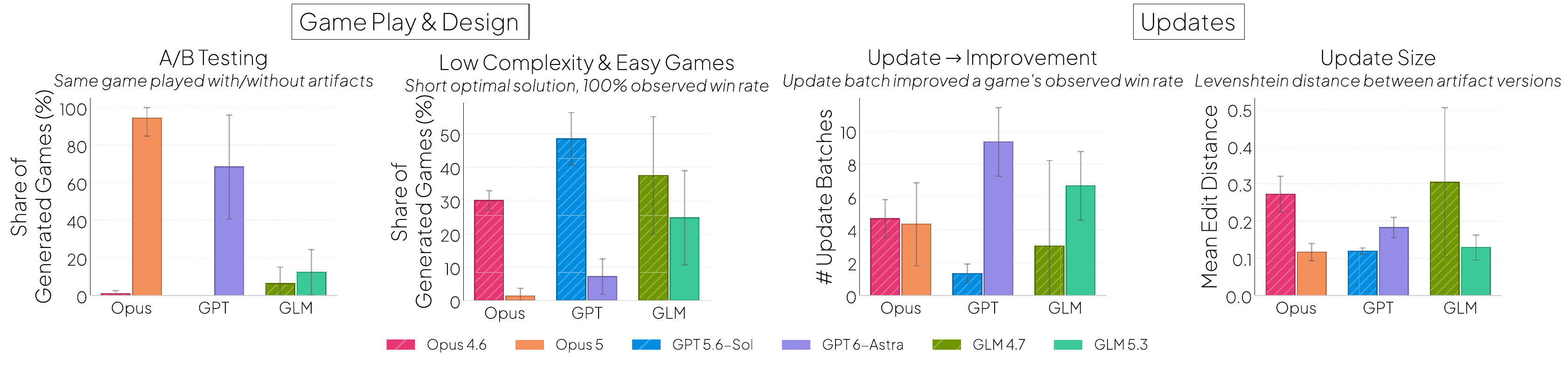}
    \vspace{-2em}
    \caption{\small \textbf{Algorithmically measured behaviors.} \textbf{(Left: Game play and design)} Newer models more often use A/B Testing (generated games played with and without teacher-written artifacts) and produce fewer games that have low complexity and low difficulty (optimal solution length $<$ 5 and 100\% teacher-observed win rate). \textbf{(Right: Updates)} For \texttt{GPT} and \texttt{GLM}, newer generations exhibit more updates that improve the observed win rate on at least one self-generated game. For \texttt{Opus} and \texttt{GLM}, newer models prefer smaller context updates. 
    }\label{fig:alg-measures}
    \label{fig:zimmerman}
    \vspace{-1pt}
\end{figure*}
\begin{figure*}[t]
    \centering
\includegraphics[width=1.0\linewidth]{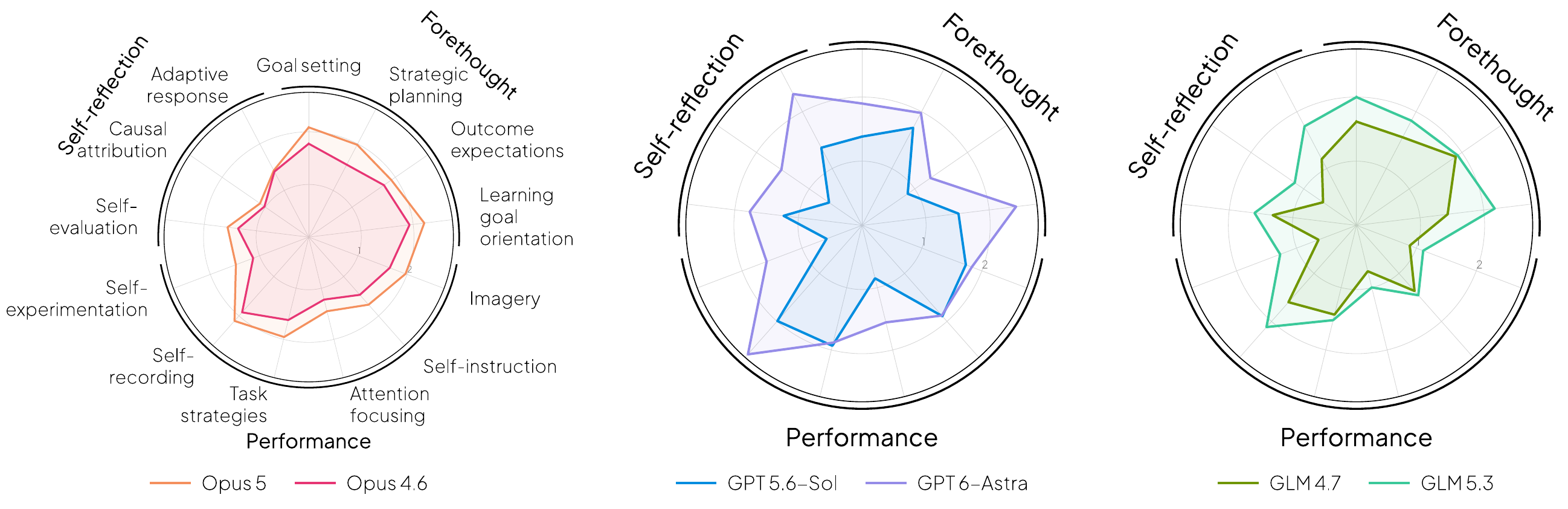}
    \vspace{-2em}
    \caption{\small \textbf{Self-improving models display human-recognizable learning strategies.} Alignment scores to 13 subprocesses of Zimmerman's self-regulated learning cycle, annotated by an LLM judge and averaged over 3 teacher sessions per model. Models that self-improve score higher across most subprocesses. Average over seeds; individual seeds in Appendix \ref{sec:app-srl-seeds}.}\label{fig:srl}
    \label{fig:zimmerman}
    \vspace{-1pt}
\end{figure*}

\begin{figure*}[h]
    \centering
\includegraphics[width=1.0\linewidth]{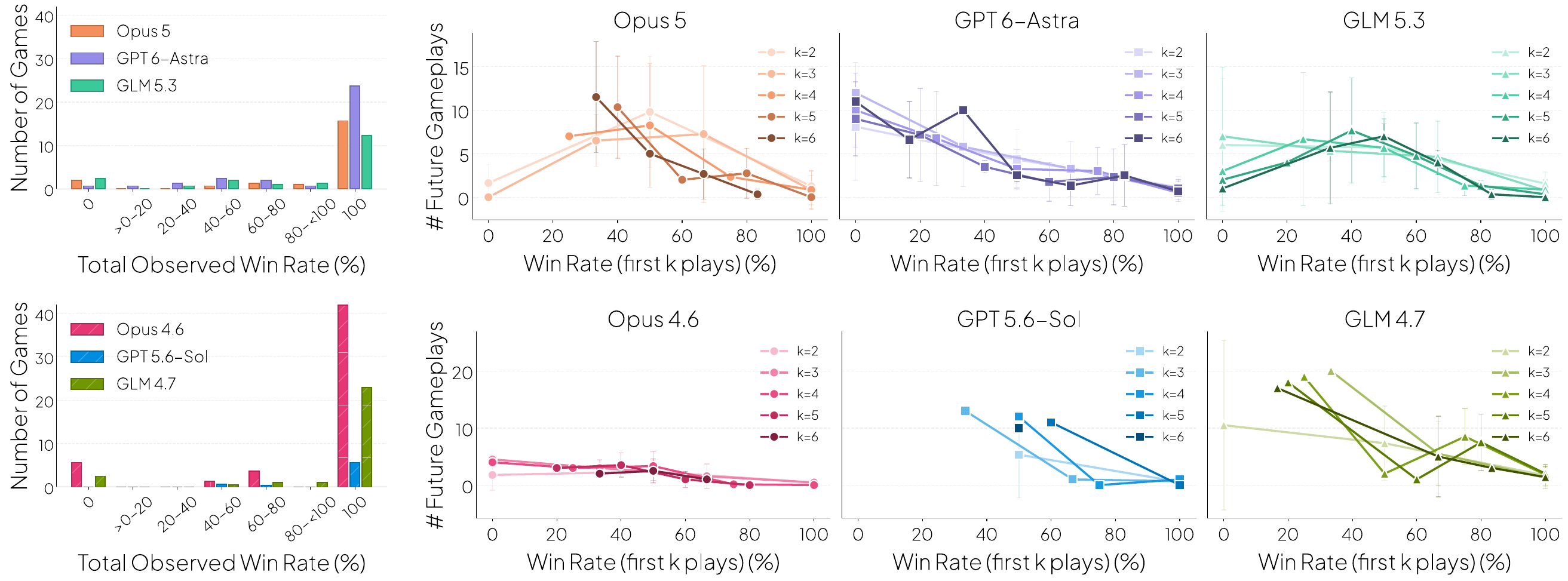}
    \vspace{-2em}
    \caption{\small \textbf{Self-improving agents use learnability-seeking task selection}. \textbf{(left)} Across models, most generated games are easy. \textbf{(right)} Newer model generations spend the most effort (\# replays) on games with low to medium win rates. Observed win rates are computed over the first $k$ plays for each game ($k \in [2,6]$). This can lead to an inverted-U resembling learnability objectives in asymmetric self-play and UED.}\label{fig:task-reward}
\end{figure*}
\vspace{-5pt}

\begin{figure*}[t]
    \centering
    \includegraphics[width=1.0\linewidth]{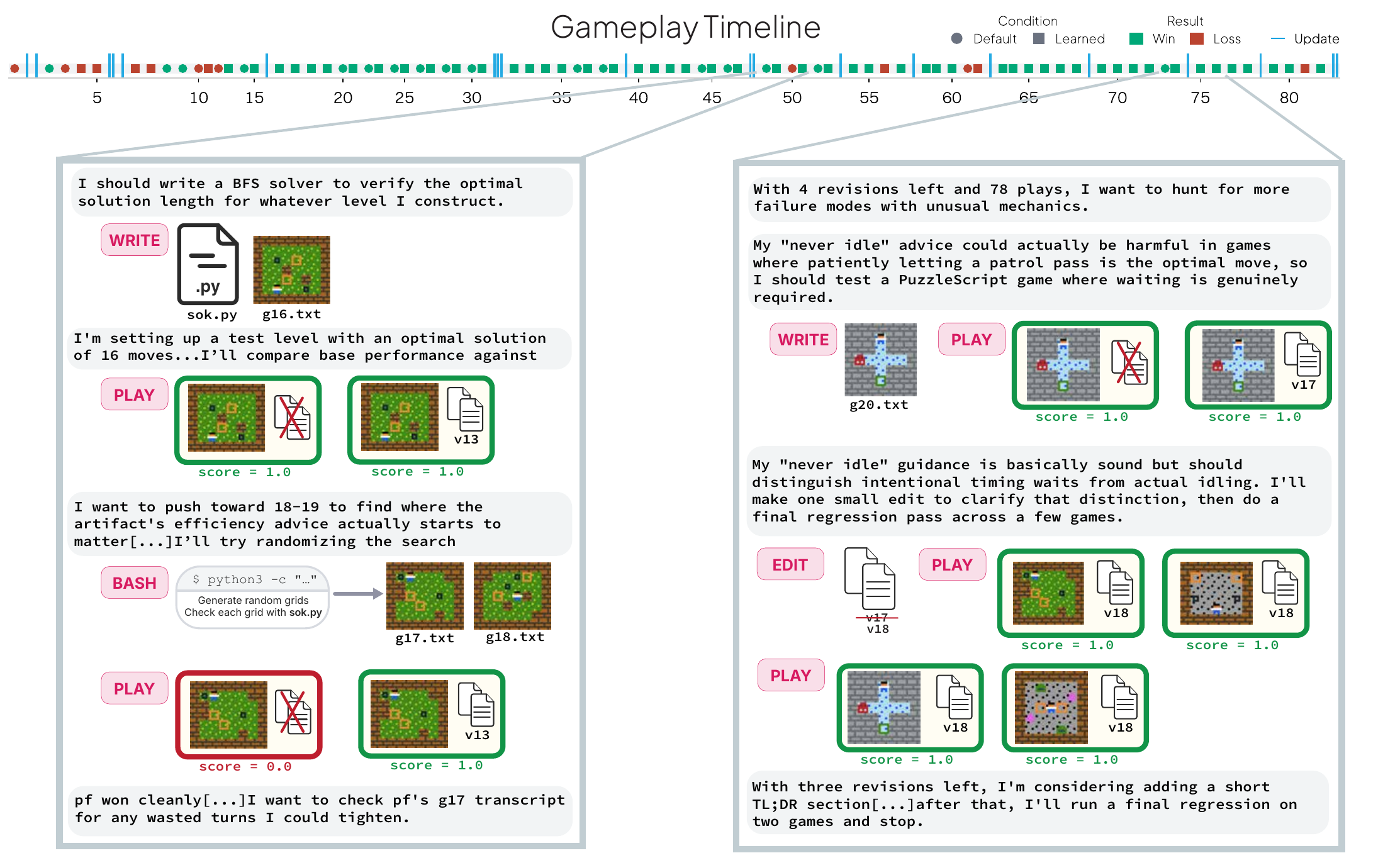}
    \vspace{-1em}
    \caption{\small \textbf{Excerpts from a \texttt{Claude Opus-5} teacher session with high alignment to self-regulated learning subprocesses.} In the first example (\textit{task-strategies}) the teacher writes a random level generator and BFS solver to filter generated games for difficulty and employs A/B testing to identify where its optimized text artifacts are useful. In the second (\textit{learning goal orientation}) the teacher designs an adversarial game to test its advice and runs validation tests across earlier games to identify regressions.}
    \label{fig:timeline}
    \vspace{-1em}
\end{figure*}

We begin with a motivating observation: For \texttt{Claude Opus-5}, under matched game-rollout budgets, \textit{self-organized learning} in our \textit{data-free} setting achieves comparable performance to a \textit{prescribed} optimization loop (GEPA, \citep{agrawal2026gepa}) that hill-climbs on 20 \textit{real in-distribution games} (Figure \ref{fig:gepa}). This calibrates the scale of expected performance gains with context optimization. Agents can construct useful learning procedures and tasks; which agents do so and what learning strategies do they employ?

\subsection{Newer model generations teach themselves better.}\label{sec:performance}
We first ask which models self-improve in our setting.
Evaluation results on \texttt{PuzzleScript-t20} are in Figure \ref{fig:generation-results}. Across model families, the three more recently released and broadly capable generations improve over the default-context player baseline (+20\% for \texttt{Claude Opus-5}, +10\% for \texttt{GPT 6-Astra}, +6\% for \texttt{GLM 5.3}). Critically however, this effect is not evident for all models; older generations show little or no improvement. The gap indicates a potential trajectory where self-organized self-learning  is increasingly evident in successive model generations. 
\vspace{-5pt}

\subsection{LLMs follow recognizable self-improvement strategies.}\label{sec:srl}
\vspace{-5pt}
The previous section shows a performance gap between older and newer models, however does not explain which strategies drive effective self-learning. 
Here, we show that the behaviors of self-improving agents align better to recognizable self-learning strategies across cognitive and machine-learning literature.

\paragraph{Quantifiable gameplay and update behaviors. } We first examine teacher behaviors that can be algorithmically measured (Figure \ref{fig:alg-measures}). Metrics for game design and play are consistent across model families. Models that self-improve frequently test the same game with and without text artifacts to directly measure player performance differences (A/B testing). These models also generate fewer trivial games (both short solution lengths and high win-rates), corroborated by histograms over solution lengths of generated games (Figure \ref{fig:opt-sol-len}). Metrics for teacher updates are less universal across families. \texttt{Claude Opus-5} and \texttt{GLM 5.3} exhibit a preference for smaller text updates. \texttt{GPT 6-Astra} and \texttt{GLM 5.3} exhibit more batched updates that improve observed performance on at least one generated game, versus their older generations. In Appendix \ref{sec:app-teacher-obs-improvement} we verify that empirically-measured change in performance agrees with teacher-observed performance gains.  
 
\paragraph{Alignment to Self-Regulated Learning.} Beyond directly quantifiable behaviors, we assess self-organized learning procedures using Zimmerman's cyclical phases model, a classic social-cognitive framework for self-regulated learning (SRL)
\citep{zimmermanBecoming}. The framework organizes SRL into subprocesses under \textit{forethought} (goal setting and planning), \textit{performance} (executing strategies and self-monitoring), and \textit{self-reflection} (evaluating performance and guiding subsequent learning). We functionally adapt this framework into a behavioral rubric for scoring the occurrence and quality of 13 subprocesses on a 0-3 scale, treating the teacher-player system as the unit of analysis. We use \texttt{GPT 6-Astra} as a judge. 
Scores for different teacher sessions of the same model align, indicating stable behavior across runs (Appendix \ref{sec:app-srl-seeds}). 

Models that effectively self-improve show higher SRL alignment (Figure \ref{fig:srl}). This is not necessarily a causal explanation, but quantifies that better self-learners exhibit human-recognizable task-selection, evaluation, and reflection strategies. Figure \ref{fig:timeline} illustrates examples: the teacher uses A/B testing to evaluate its text artifacts (\textit{self-experimentation}), writes a procedural game generator with BFS-based difficulty filtering (\textit{task strategies}), and designs adversarial games to stress-test its notes (\textit{learning goal orientation}). 
Further examples are in Appendix \ref{sec:app-qualitative}. Several subprocesses also overlap with the behaviors quantified in Figure \ref{fig:alg-measures} (e.g., self-experimentation, causal attribution).
\label{sec:generalization}
\begin{figure*}[t]
    \centering
    \includegraphics[width=1.0\linewidth]{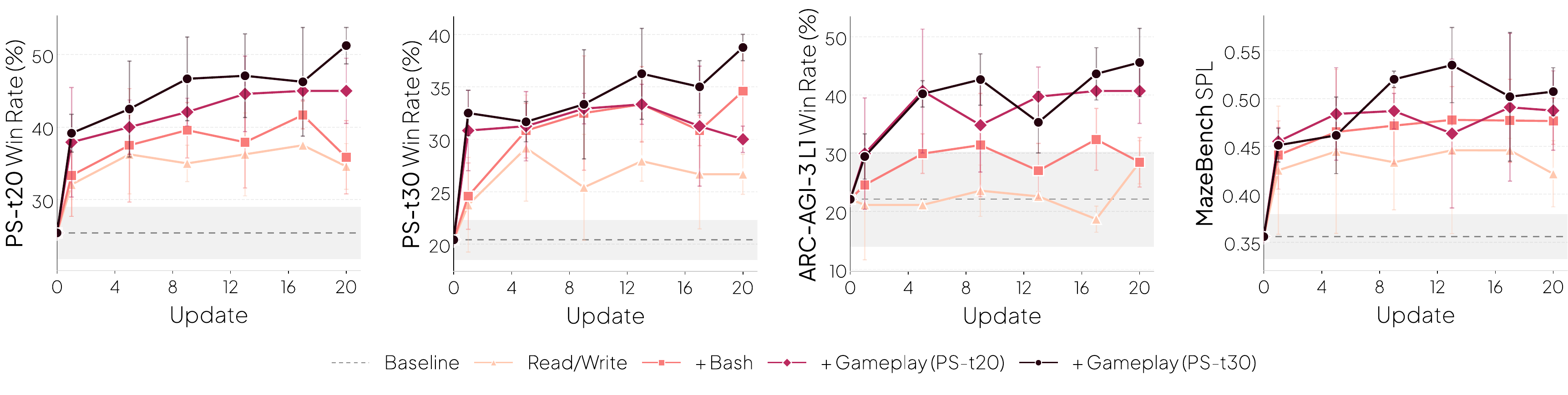}
    \vspace{-2em}
    \caption{\small \textbf{Improvements from self-organized learning transfer across benchmarks}. Player performance is evaluated at intermediate context updates along the self-learning trajectory on \textsc{PuzzleScript-t20}, \textsc{PuzzleScript-t30}, \textsc{ARC-AGI-L1} and \textsc{MazeBench}. The context optimized under our canonical setting, \textit{Gameplay (PS-t20)}, improves over the unoptimized baseline (dashed line) on PuzzleScript games with both 20- and 30-turn horizons. These gains transfer to different game engines and representations in \textsc{ARC-AGI-L1} and \textsc{MazeBench}. We also compare progressively richer teacher toolsets, starting from Read/Write, adding Bash, and then PuzzleScript-Gameplay tools. Error bars show $\pm 1$ SD over three independent teacher seeds.}\label{fig:task-generalization}
    \vspace{-1em}
\end{figure*}

\paragraph{Alignment to learnability.} 
As a complementary view of teacher behavior, we show in Figure \ref{fig:task-reward} that models that effectively self-improve allocate effort to tasks similarly. The majority of generated games are easy but quickly discarded. Games with a 100\% observed win rate have a low chance of being replayed. Replay rates are higher for games with a lower observed pass rate (below 50\%).
\begin{wrapfigure}[17]{r}{0.4\linewidth}
    \centering
    \includegraphics[width=\linewidth]{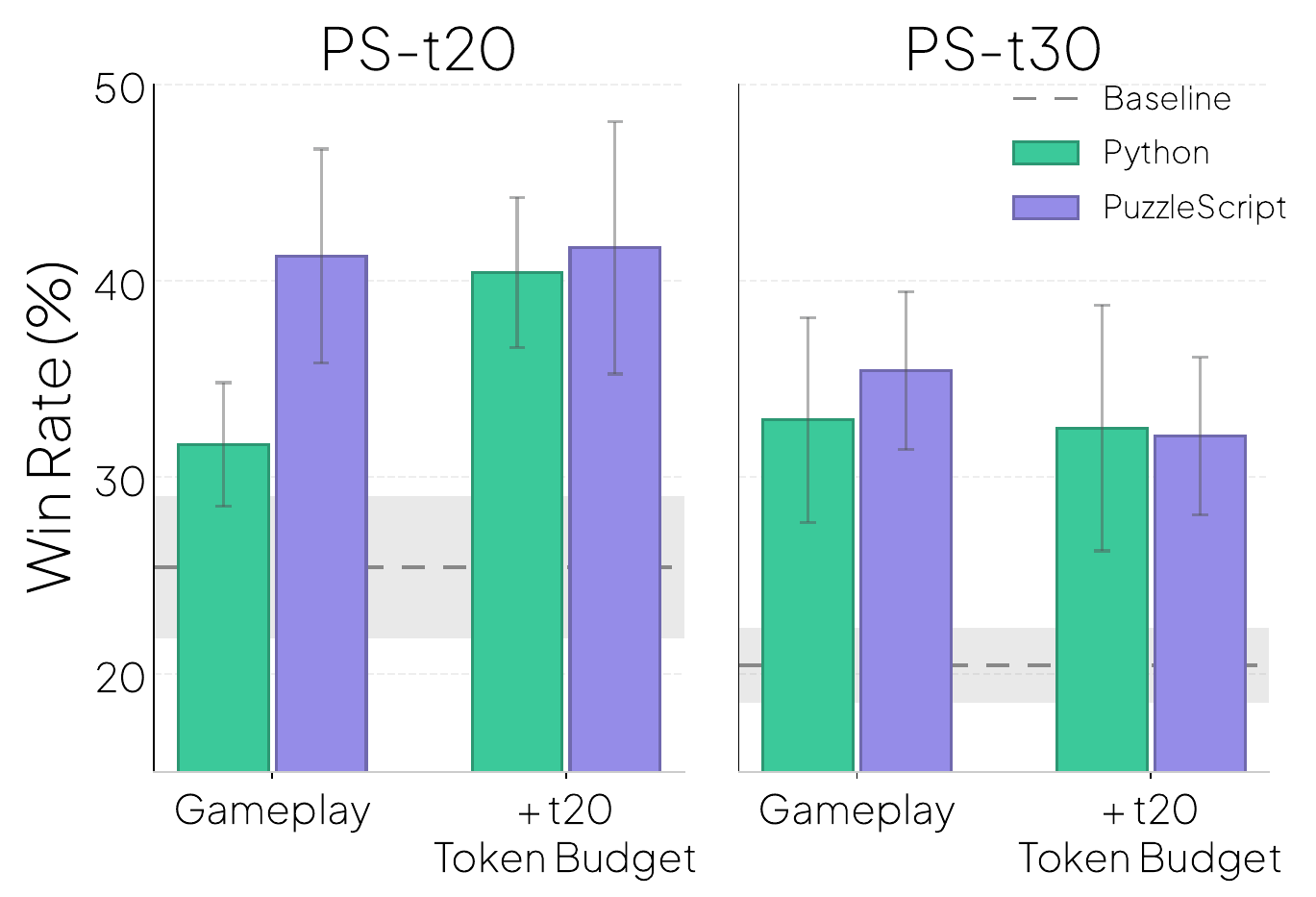}
    \caption{
    \textbf{PuzzleScript-specific gameplay is not essential for transfer.}
    We compare Python-Gameplay tool to the PuzzleScript-Gameplay tool with and without imposed token-budgets.\looseness=-1} 
    \label{fig:python-generalization}
    \vspace{-10pt}
\end{wrapfigure}
Particularly for \texttt{Claude Opus-5} and \texttt{GLM 5.3}, most replays concentrate in games with a 20-60\% observed win rate so far. Notably, these task-selection statistics are less consistent across models that fail to self-improve, partly because they replay games at a lower frequency. Full gameplay timelines for each game are in Appendix \ref{sec:app-full-timelines}. 
Some agents demonstrate an "inverted-U" that resembles the learnability task rewards used in asymmetric self-play and unsupervised environment design to encourage intermediate-difficulty generated tasks \citep{huang2026rzero, rutherford2024no}. 
Learnability-based task selection in machine learning is itself rooted in the "Zone of Proximal Development" concept in developmental psychology \citep{vygotsky1978mind, wang2022in}. We find that self-improving LLM agents prefer similar task selection without being steered to do so.

\section{When does self-organized self-improvement generalize?}\label{sec:full-transfer}
Section \ref{sec:performance} established that a class of capable models self-improve with minimal guidance on \textsc{PuzzleScript-t20}, a benchmark that shares key attributes with the self-learning setting. Now, using \texttt{Claude Opus-5}, we evaluate \textit{generalization} from three perspectives: transfer across environments, under different information regimes, and across models. 

\subsection{Self-learned artifacts transfer across environments.}\label{sec:env-transfer}
Figure \ref{fig:task-generalization} shows that learned context transfers to environments that differ increasingly from the self-optimization setting. We plot player performance at even intervals along the self-learning trajectory. At context update $k$ we evaluate a fresh player session with $c_k$ on our held-out benchmarks. The context optimized under 20-turn PuzzleScript gameplays (denoted \textit{Gameplay (PS-t20)}) improves \textsc{PuzzleScript-t20} performance by 20\%, \textsc{PuzzleScript-t30} by 9\%, \textsc{ARC-AGI-L1} by 21\%, and MazeBench SPL by 0.15. This demonstrates transfer across game engines, grid sizes, core mechanics, 3D spatial navigation, longer-horizon reasoning, and state representations. 

These trends also extend to different self-improvement settings. We run three comparisons, each with three independent teacher sessions at matched teacher budgets. (1) The teacher has only 
Read/Write tools (\textit{Read-Write}), (2) with Bash access (\textit{Read-Write-Bash}), and (3) with the Gameplay tool constrained to 30-turn PuzzleScript games. \textit{Gameplay (t-30)} performs best across benchmarks, even outperforming \textit{Gameplay (PS-t20)} on \textsc{PuzzleScript-t20}. \textit{Read-Write} outperforms the baseline, indicating that some of the observed improvement in \textit{Gameplay} settings stems from the 
\begin{figure*}[t]
    \centering
    \includegraphics[width=0.75\linewidth]{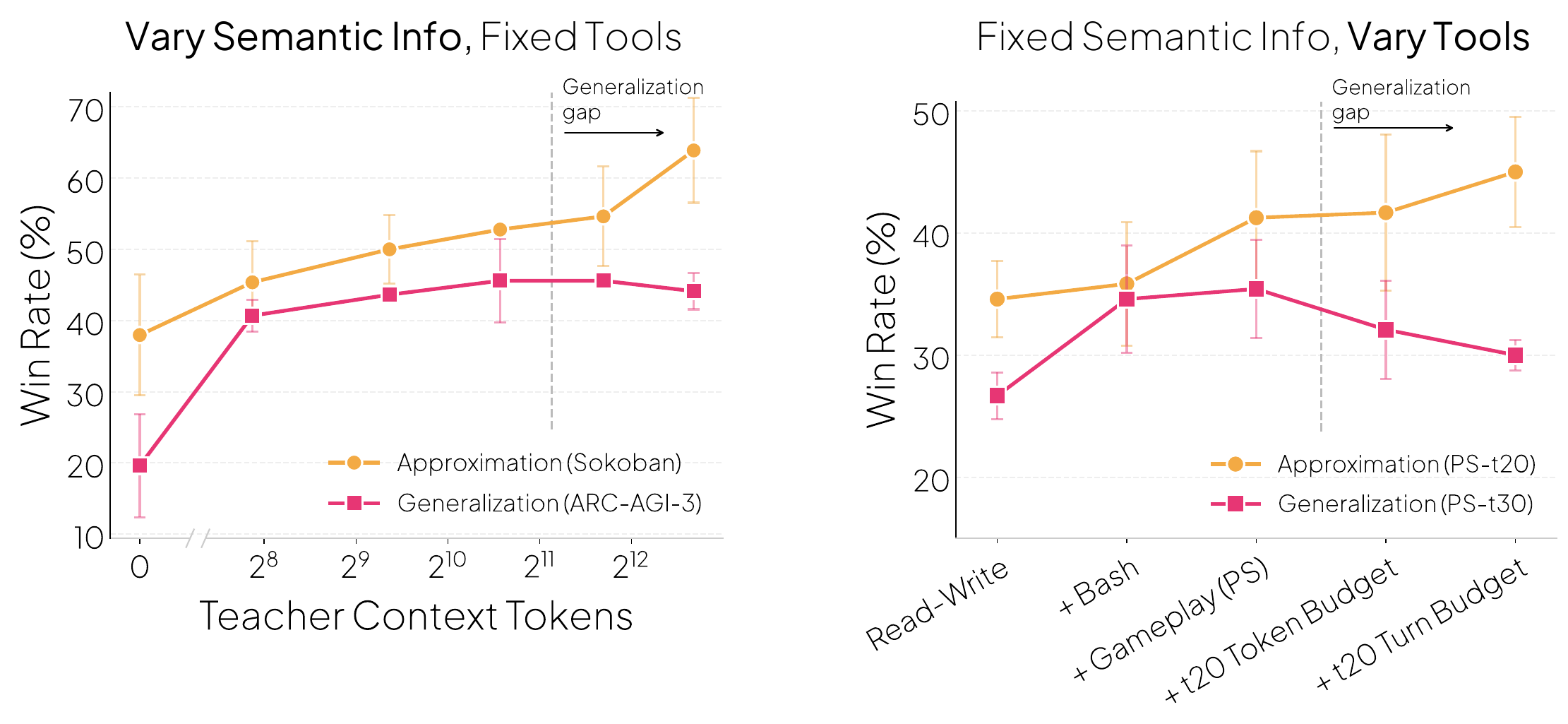}
    \vspace{-1em}
    \caption{\small \textbf{Generalization with respect to progressively richer information, via context and toolsets.} \textbf{(Left)} We run teacher sessions using Python Gameplay with increasingly detailed descriptions of \textsc{PuzzleScript-Narrow} games. Performance initially improves both on the same games, and \textsc{ARC-AGI-3} transfer, however at higher levels of detail transfer plateaus. \textbf{(Right)} We ablate the teacher's toolset and Gameplay constraints; performance initially improves on both \textsc{PuzzleScript-t20} and \textsc{PuzzleScript-t30} with richer toolsets, however declines on \textsc{PuzzleScript-t30} after adding t20-specific token and turn budgets. Error bars show $\pm1$ SD over three independent teacher seeds.
    }\label{fig:tool-generalization}
    \vspace{-1em}
\end{figure*}
teacher's self-reflection on prior knowledge. On all benchmarks besides \textsc{MazeBench}, \textit{Read-Write} recovers less than 50\% of the performance gain of \textit{Gameplay} teachers, indicating benefit from generated experience. \textit{Read-Write-Bash} outperforms \textit{Read-Write}, but fails to recover the full gains of separate gameplay. Notably, in one \textit{Read-Write-Bash} session the teacher still writes and runs game simulators in the sandbox to test its advice despite no mention of gameplay in its prompt (Appendix \ref{sec:app-qualitative}).
For a sufficiently strong teacher with Bash, an explicit Gameplay tool may be unnecessary.

We also evaluate allowing the teacher to issue Python games; Python gameplay recovers nearly all the performance gain of PuzzleScript gameplay in 3 out of 4 cases (Figure \ref{fig:python-generalization}), suggesting that PuzzleScript-awareness is not strictly necessary for the teacher to learn transferable context. We further investigate  Python-based generation in the next section.

\subsection{Teacher information shapes generalization.}\label{sec:information}
How does overfitting arise in our setting? We do not provide external tasks or validation feedback, however we communicate information about the evaluation setting via the teacher's \textit{context} and its \textit{tools}. Figure \ref{fig:tool-generalization} illustrates the relationship between teacher information and generalization under two regimes: semantic information via the teacher prompt, and mechanic information via tools.

\textbf{Information via context.} We run teacher sessions with increasingly-detailed generated summaries of \textsc{PuzzleScript-Narrow}, a set of nine Sokoban-style games (details in Appendix \ref{sec:app-context-ablations}). Teacher sessions use the\textbf{ Python Gameplay} tool, ensuring that the prompt is the only source of semantic game information. With Python, the teacher can flexibly generate games that align with or that differ significantly from PuzzleScript. We evaluate \textit{approximation} (\textsc{PuzzleScript-Narrow}) and \textit{generalization} (\textsc{ARC-AGI-3}). Richer context initially improves performance on both, as the teacher is steered towards 2D grid-based games. At highest token counts however, \textsc{PuzzleScript-Narrow} performance improves while \textsc{ARC-AGI-3} performance plateaus and slightly declines. 

\textbf{Information via tools.} We ablate the teacher's awareness of the \textsc{PuzzleScript-t20} token and turn budgets, and plot \textit{approximation} (\textsc{PuzzleScript-t20} performance) versus \textit{generalization} (\textsc{PuzzleScript-t30} performance). Setup details are in Appendix \ref{sec:app-eval-awareness-ablations}. Performance rises as tools and constraints relevant to both benchmarks are added, such as a PuzzleScript Gameplay tool with teacher-selected turn and token budgets. Imposing Gameplay turn- and token-budgets that match the t20 setting, however, improves \textsc{PuzzleScript-t20} performance while \textsc{PuzzleScript-t30} transfer falls. 

These results yield an intuitive takeaway: Increased awareness of the evaluation setting biases the teacher towards generating relevant experience, but can also encourage task-specific specialization.

\subsection{Self-learned artifacts transfer between models.}\label{sec:cross-model}

\begin{wraptable}[8]{r}{0.40\textwidth}
\vspace{-1.7\baselineskip}
\centering
\footnotesize
\newcommand{\sd}[1]{\,{\color{black!60}\scriptsize$\pm$\,#1}}
\renewcommand{\arraystretch}{1.08}
\setlength{\tabcolsep}{3pt}

\begin{tabularx}{\linewidth}{lXr}
\toprule
Player & Text artifact source & Win rate (\%) \\
\midrule
Opus 5 & None & 25.4\sd{3.6} \\
\rowcolor{black!5}
       & GPT-6 Astra & \textbf{47.5}\sd{4.5} \\
\midrule
GPT-6 Astra & None & 67.1\sd{3.8} \\
\rowcolor{black!5}
            & Opus 5 & \textbf{72.1}\sd{4.0} \\
\bottomrule
\end{tabularx}

\vspace{-.4em}
\caption{\small \textbf{Cross-model transfer of self-optimized context}
on PuzzleScript-t20. Mean $\pm$ std over three seeds.}
\vspace{-12pt}
\label{tab:transfer}
\end{wraptable}

Does self-optimized text contain general lessons that another agent might benefit from, or is that text overfit to the target agent's generated experience? We evaluate \texttt{Claude Opus-5} on \textsc{PuzzleScript-t20} using $c_{final}$ self-optimized by \texttt{GPT 6-Astra} and vice versa. The learned text transfers in both directions, even despite \texttt{GPT 6-Astra} having higher baseline performance (Table \ref{tab:transfer}).

\section{Related Work}\label{sec:related-work}
For an extended view of related work, see Appendix \ref{sec:app-related-work}. We categorize self-improvement methods based on which components of the learning process remain externally specified, how learning takes place, and the scope of generalization that is studied. 

\textbf{Asymmetric self-play}. In asymmetric self-play a teacher generates challenging problems for a student to solve, resulting in an emergent, adaptive curriculum \citep{sukhbaatar2017asymmetric}. Such methods have found success in post-training reasoning LLMs \citep{zhao2025absolute, huang2026rzero, poesia2024_minimo, dong2025_stp, liu2025_spice, zhang2025better,teodorescu2023codeplay}. In these methods the teacher self-generates the source of experience by synthesizing student problems, while learning proceeds in weight-space. The optimization procedure (e.g., update schedules, teacher rewards, the number of sampled tasks in a batch) remains externally specified. Such methods can also suffer from diversity or performance collapse \citep{chae2025understandingselfplay}, motivating additional mechanisms for assessing the learning value of generated problems. Various methods use judges \citep{bailey2026scaling} or ground the agent in real-data hard questions as goalposts \citep{Jana2026GASPGA}. These mechanisms help stabilize self-play but introduce additional externally specified choices about what constitutes a useful learning experience. Collapse can also be addressed by rewarding the teacher based on whether its generated problems induce student improvement on external hard problems \citep{sundaram2026teachingmodelsteachthemselves}. In this case, the feedback is externally obtained while still relying on self-generated data to drive learning. Recent works have the teacher generate interactive environments rather than question-answer tasks \citep{Liu2026SPADESI} but retain externally specified optimization and task-selection procedures.

\textbf{Context and harness optimization}. A separate line of work considers self-improvement through updates to the model's context or agent harness  \citep{suzgun2025dynamic, wang2025agent, hu2025automated, zhang2025aflow}. GEPA uses natural-language reflection over execution trajectories to optimize model prompts via evolutionary search \citep{agrawal2026gepa}, while ACE maintains an evolving playbook that accumulates strategies from previous interactions \citep{zhang2026agentic}. These works still leverage an externally specified task distribution and pre-specified context optimization procedure. 
MCE partly relaxes the latter by co-evolving the context with a context-engineering skill \citep{ye2026meta}. Meta-Harness expands the object of optimization further to executable harness code, allowing the model to change how context is managed, including what information is stored and retrieved \citep{lee2026metaharness}.
The optimization still uses externally provided tasks, and candidate harnesses are evaluated with an externally specified objective. Recent work on self-modifying agents further expands the components under the model's control. The Darwin Gödel Machine iteratively modifies its own agent code and evaluates the resulting agents on external benchmarks \citep{zhang2026darwin}, while Hyperagents make the meta-level procedure used to generate future improvements itself editable \citep{zhang2026hyperagents}. These approaches allow increasing parts of the self-improvement mechanism to change, while still operating within an externally specified task and evaluation setting.

\textbf{Self-constructed learning procedures.} \citet{xue2026rethinking} introduce Open-Ended Optimization, where the objective, available interactions, resource budget, data boundary and evaluation are fixed, but the model organizes the improvement process itself. Performance is evaluated on held-out examples from the same benchmark setting used for optimization. 
Closest to our setting, in concurrent work, \citet{wu2026aspire} introduce ASPIRE, a benchmark for self-evolution from vague capability goals. Agents 
(sometimes a stronger teacher  optimizing a smaller student) improve weights or harnesses with access to external data and, in some settings, aggregate scores from hidden evaluations. Performance is measured on hidden items from the same goal. In contrast, we study a model improving itself from self-generated experience alone, and evaluate transfer to different environments.
\vspace{-0.7em}

\section{Discussion and Limitations}
Our results align to the basic trend in deep learning that as models become more broadly capable, they require fewer human-imposed inductive biases. We study this phenomenon at the meta-optimization layer, showing that LLM agents compose their own self-improvement loops with minimal guidance on how to do so, and demonstrate recognizable learning strategies. 

\textbf{Models are improving at self-improvement.} The key implication is not necessarily that a single generation of agents effectively self-improves, but rather that \textit{this meta-capability improves across generations}. The agents that effectively self-improve in our experiments require far less scaffolding than older generations of agents, yet still benefit from having an explicitly provided tool for observing a sub-agent play games. The next generation of agents may require even less scaffolding. Our experiments point to this possibility; in our \texttt{Claude Opus-5} \textit{Read-Write-Bash} experiments, one teacher session tests its notes by running command-line game simulations, without any provided Gameplay tool or reference to simulating games in its prompt (Section \ref{sec:app-qualitative}). Indeed, a sufficiently capable model could simply execute games in its chain of thought, without a written program or separate agent instance. 

\textbf{Safety.} Another consequence of our findings is that self-organized improvement is not a discovery waiting to happen. We observe this capability in a controlled setting, over short horizons with relatively simple puzzle-based tasks. As this capability improves and potentially is observed in more complex settings, it is important to scientifically understand it and characterize model behavior.

\textbf{Beyond the plateau?} Our experiments suggest that continual revision can have diminishing returns (Figure \ref{fig:task-generalization}). In Appendix \ref{sec:app-compute} we show that extra teacher update and gameplay budgets can result in higher performance, however still observe a plateauing effect. This may reflect limitations in self-improvement capabilities, or even the limitations of context-based learning. We study self-improvement over 20 updates; what will it take for models to continue improving over 100 updates, or 1000? Traditional learning algorithms improve until they minimize the given objective under certain conditions; it remains unclear if the learning abilities of these models exhibit similar properties.

\paragraph{Limitations.} Our experiments show that several human-interpretable self-learning strategies are exhibited more frequently by effective self-improving models. However, these results do not fully disentangle a causal relationship between self-regulatory strategies and self-improvement. It may also be the case that less human-interpretable behaviors also shape agent performance; how to effectively measure and characterize such behaviors remains an open question.  

Another open question is whether our observed trends extend to more open-ended domains where models have weaker priors. Games have an extensive literature on curriculum design and learnable task generation that models have likely ingested. The use of PuzzleScript may hint at real-world environments or bias the model, though our Python-game ablations show that PuzzleScript is not a critical factor for self-improvement. 
Nonetheless, games provide a rich setting with several controllable axes of variation, including game mechanics, game horizon and observation representations.

\section{Acknowledgments}
We thank Nathan Cloos, Antonio Norelli, Ulyana Piterbarg, Souradip Chakraborty, and Sarvesh Sundaram for helpful discussions. This work was supported by an NSF GRFP fellowship to SS, by a Packard Fellowship to P.I., by ONR MURI grants N00014-22-1-2740 and N00014-26-1-2024, and by gifts from Hudson River Trading and Adobe.

\newpage 
\bibliographystyle{abbrvnat}
\bibliography{iclr2027/iclr2027_conference}

\begin{thebibliography}{40}
\providecommand{\natexlab}[1]{#1}
\providecommand{\url}[1]{\texttt{#1}}
\expandafter\ifx\csname urlstyle\endcsname\relax
  \providecommand{\doi}[1]{doi: #1}\else
  \providecommand{\doi}{doi: \begingroup \urlstyle{rm}\Url}\fi

\bibitem[Agrawal et~al.(2026)Agrawal, Tan, Soylu, Ziems, Khare, Opsahl-Ong, Singhvi, Shandilya, Ryan, Jiang, Potts, Sen, Dimakis, Stoica, Klein, Zaharia, and Khattab]{agrawal2026gepa}
L.~A. Agrawal, S.~Tan, D.~Soylu, N.~Ziems, R.~Khare, K.~Opsahl-Ong, A.~Singhvi, H.~Shandilya, M.~J. Ryan, M.~Jiang, C.~Potts, K.~Sen, A.~Dimakis, I.~Stoica, D.~Klein, M.~Zaharia, and O.~Khattab.
\newblock {GEPA}: Reflective prompt evolution can outperform reinforcement learning.
\newblock In \emph{The Fourteenth International Conference on Learning Representations}, 2026.
\newblock URL \url{https://openreview.net/forum?id=RQm2KQTM5r}.

\bibitem[Bailey et~al.(2026)Bailey, Wen, Dong, Hashimoto, and Ma]{bailey2026scaling}
L.~Bailey, K.~Wen, K.~Dong, T.~Hashimoto, and T.~Ma.
\newblock Scaling self-play with self-guidance.
\newblock \emph{arXiv preprint arXiv:2604.20209}, 2026.

\bibitem[Chae et~al.(2025)Chae, Alam, and Rastogi]{chae2025understandingselfplay}
J.~Y. Chae, M.~T. Alam, and N.~Rastogi.
\newblock Towards understanding self-play for {LLM} reasoning.
\newblock In \emph{The 5th Workshop on Mathematical Reasoning and AI at NeurIPS 2025}, 2025.
\newblock URL \url{https://openreview.net/forum?id=ENymEXXRVH}.

\bibitem[Dennis et~al.(2020)Dennis, Jaques, Vinitsky, Bayen, Russell, Critch, and Levine]{dennis2020paired}
M.~Dennis, N.~Jaques, E.~Vinitsky, A.~Bayen, S.~Russell, A.~Critch, and S.~Levine.
\newblock Emergent complexity and zero-shot transfer via unsupervised environment design.
\newblock In H.~Larochelle, M.~Ranzato, R.~Hadsell, M.~Balcan, and H.~Lin, editors, \emph{Advances in Neural Information Processing Systems}, volume~33, pages 13049--13061. Curran Associates, Inc., 2020.
\newblock URL \url{https://proceedings.neurips.cc/paper_files/paper/2020/file/985e9a46e10005356bbaf194249f6856-Paper.pdf}.

\bibitem[Dong and Ma(2025)]{dong2025_stp}
K.~Dong and T.~Ma.
\newblock Stp: Self-play llm theorem provers with iterative conjecturing and proving, 2025.
\newblock URL \url{https://arxiv.org/abs/2502.00212}.

\bibitem[Earle et~al.(2025)Earle, Todd, Li, Khalifa, Nasir, Jiang, Banburski-Fahey, and Togelius]{Earle2025PuzzleJAXAB}
S.~Earle, G.~Todd, Y.~Li, A.~Khalifa, M.~U. Nasir, Z.-H. Jiang, A.~Banburski-Fahey, and J.~Togelius.
\newblock Puzzlejax: A benchmark for reasoning and learning.
\newblock \emph{ArXiv}, abs/2508.16821, 2025.
\newblock URL \url{https://api.semanticscholar.org/CorpusID:280711515}.

\bibitem[Foundation(2026)]{foundation2026arc}
A.~Foundation.
\newblock Arc-agi-3: A new challenge for frontier agentic intelligence.
\newblock \emph{arXiv preprint arXiv:2603.24621}, 2026.

\bibitem[Hu et~al.(2025)Hu, Lu, and Clune]{hu2025automated}
S.~Hu, C.~Lu, and J.~Clune.
\newblock Automated design of agentic systems.
\newblock In \emph{The Thirteenth International Conference on Learning Representations}, 2025.
\newblock URL \url{https://openreview.net/forum?id=t9U3LW7JVX}.

\bibitem[Huang et~al.(2026)Huang, Yu, Wang, Zhang, Li, Li, Huang, Mi, and Yu]{huang2026rzero}
C.~Huang, W.~Yu, X.~Wang, H.~Zhang, Z.~Li, R.~Li, J.~Huang, H.~Mi, and D.~Yu.
\newblock R-zero: Self-evolving reasoning {LLM} from zero data.
\newblock In \emph{The Fourteenth International Conference on Learning Representations}, 2026.
\newblock URL \url{https://openreview.net/forum?id=96apU6YzSO}.

\bibitem[Jana et~al.(2026)Jana, Sancaktar, Danis, Martius, Orvieto, and Kolev]{Jana2026GASPGA}
S.~Jana, C.~Sancaktar, T.~Danis, G.~Martius, A.~Orvieto, and P.~Kolev.
\newblock Gasp: Guided asymmetric self-play for coding llms.
\newblock \emph{ArXiv}, abs/2603.15957, 2026.
\newblock URL \url{https://api.semanticscholar.org/CorpusID:286579725}.

\bibitem[Jiang et~al.(2020)Jiang, Grefenstette, and Rockt{\"a}schel]{Jiang2020PrioritizedLR}
M.~Jiang, E.~Grefenstette, and T.~Rockt{\"a}schel.
\newblock Prioritized level replay.
\newblock In \emph{International Conference on Machine Learning}, 2020.
\newblock URL \url{https://api.semanticscholar.org/CorpusID:222208809}.

\bibitem[Jiang et~al.(2021)Jiang, Dennis, Parker-Holder, Foerster, Grefenstette, and Rockt\"{a}schel]{jiang2021ued}
M.~Jiang, M.~Dennis, J.~Parker-Holder, J.~Foerster, E.~Grefenstette, and T.~Rockt\"{a}schel.
\newblock Replay-guided adversarial environment design.
\newblock In M.~Ranzato, A.~Beygelzimer, Y.~Dauphin, P.~Liang, and J.~W. Vaughan, editors, \emph{Advances in Neural Information Processing Systems}, volume~34, pages 1884--1897. Curran Associates, Inc., 2021.
\newblock URL \url{https://proceedings.neurips.cc/paper_files/paper/2021/file/0e915db6326b6fb6a3c56546980a8c93-Paper.pdf}.

\bibitem[Lee et~al.(2026)Lee, Nair, Zhang, Lee, Khattab, and Finn]{lee2026metaharness}
Y.~Lee, R.~S. Nair, Q.~Zhang, K.~Lee, O.~Khattab, and C.~Finn.
\newblock Meta-harness: End-to-end optimization of model harnesses.
\newblock In \emph{Third Conference on Language Modeling}, 2026.
\newblock URL \url{https://openreview.net/forum?id=tmbOUyFx3R}.

\bibitem[Liu et~al.(2025)Liu, Jin, Kim, Yuan, Zhao, Kulikov, Li, Sukhbaatar, Lanchantin, and Weston]{liu2025_spice}
B.~Liu, C.~Jin, S.~Kim, W.~Yuan, W.~Zhao, I.~Kulikov, X.~Li, S.~Sukhbaatar, J.~Lanchantin, and J.~Weston.
\newblock Spice: Self-play in corpus environments improves reasoning, 2025.
\newblock URL \url{https://arxiv.org/abs/2510.24684}.

\bibitem[Liu et~al.(2026)Liu, Yu, Jiang, Qu, Zhao, Liu, Kim, Zhou, Kim, Ren, Liu, Yu, Chen, Shi, Liang, Zettlemoyer, Choi, and Jaques]{Liu2026SPADESI}
B.~Liu, S.~Yu, Y.~Jiang, A.~Qu, A.~Zhao, Z.~Liu, J.~Kim, Z.~Zhou, S.~Kim, T.~Ren, M.~Liu, H.~Yu, Z.~Chen, W.~Shi, P.~Liang, L.~S. Zettlemoyer, Y.~Choi, and N.~Jaques.
\newblock Spade: Self-play in adaptive synthetic executable environments.
\newblock 2026.
\newblock URL \url{https://api.semanticscholar.org/CorpusID:291219356}.

\bibitem[Pappas and Pappas(2026)]{pappas2026mazebench}
J.~Pappas and D.~Pappas.
\newblock {Maze Bench}: Visual spatial reasoning in a 3d open world, 2026.
\newblock URL \url{https://mazebench.com}.
\newblock Project oversight by Florian Brand and Sebastian M{\"u}ller.

\bibitem[Parker-Holder et~al.(2022)Parker-Holder, Jiang, Dennis, Samvelyan, Foerster, Grefenstette, and Rockt{\"a}schel]{parker2022evolving}
J.~Parker-Holder, M.~Jiang, M.~Dennis, M.~Samvelyan, J.~Foerster, E.~Grefenstette, and T.~Rockt{\"a}schel.
\newblock Evolving curricula with regret-based environment design.
\newblock In \emph{International Conference on Machine Learning}, pages 17473--17498. PMLR, 2022.

\bibitem[Poesia et~al.(2024)Poesia, Broman, Haber, and Goodman]{poesia2024_minimo}
G.~Poesia, D.~Broman, N.~Haber, and N.~D. Goodman.
\newblock Learning formal mathematics from intrinsic motivation, 2024.
\newblock URL \url{https://arxiv.org/abs/2407.00695}.
\newblock NeurIPS 2024 (Oral).

\bibitem[Racaniere et~al.(2020)Racaniere, Lampinen, Santoro, Reichert, Firoiu, and Lillicrap]{racaniere2020settersolver}
S.~Racaniere, A.~Lampinen, A.~Santoro, D.~Reichert, V.~Firoiu, and T.~Lillicrap.
\newblock Automated curriculum generation through setter-solver interactions.
\newblock In \emph{International Conference on Learning Representations}, 2020.
\newblock URL \url{https://openreview.net/forum?id=H1e0Wp4KvH}.

\bibitem[Rutherford et~al.(2024)Rutherford, Beukman, Willi, Lacerda, Hawes, and Foerster]{rutherford2024no}
A.~Rutherford, M.~Beukman, T.~Willi, B.~Lacerda, N.~Hawes, and J.~N. Foerster.
\newblock No regrets: Investigating and improving regret approximations for curriculum discovery.
\newblock In \emph{The Thirty-eighth Annual Conference on Neural Information Processing Systems}, 2024.
\newblock URL \url{https://openreview.net/forum?id=iEeiZlTbts}.

\bibitem[Sukhbaatar et~al.(2018)Sukhbaatar, Lin, Kostrikov, Synnaeve, Szlam, and Fergus]{sukhbaatar2017asymmetric}
S.~Sukhbaatar, Z.~Lin, I.~Kostrikov, G.~Synnaeve, A.~Szlam, and R.~Fergus.
\newblock Intrinsic motivation and automatic curricula via asymmetric self-play.
\newblock In \emph{International Conference on Learning Representations}, 2018.
\newblock URL \url{https://openreview.net/forum?id=SkT5Yg-RZ}.

\bibitem[Sundaram et~al.(2026)Sundaram, Quan, Kwiatkowski, Ahuja, Ollivier, and Kempe]{sundaram2026teachingmodelsteachthemselves}
S.~Sundaram, J.~Quan, A.~Kwiatkowski, K.~Ahuja, Y.~Ollivier, and J.~Kempe.
\newblock Teaching models to teach themselves: Reasoning at the edge of learnability, 2026.
\newblock URL \url{https://arxiv.org/abs/2601.18778}.

\bibitem[Suzgun et~al.(2026)Suzgun, Y{\"{u}}ksekg{\"{o}}n{\"{u}}l, Bianchi, Jurafsky, and Zou]{suzgun2025dynamic}
M.~Suzgun, M.~Y{\"{u}}ksekg{\"{o}}n{\"{u}}l, F.~Bianchi, D.~Jurafsky, and J.~Zou.
\newblock Dynamic cheatsheet: Test-time learning with adaptive memory.
\newblock In V.~Demberg, K.~Inui, and L.~Marquez, editors, \emph{Proceedings of the 19th Conference of the European Chapter of the Association for Computational Linguistics, {EACL} 2026 - Volume 1: Long Papers, Rabat, Morocco, March 24-29, 2026}, pages 7080--7106. Association for Computational Linguistics, 2026.
\newblock URL \url{https://doi.org/10.18653/v1/2026.eacl-long.333}.

\bibitem[Teodorescu et~al.(2023)Teodorescu, Colas, Bowers, Carta, and Oudeyer]{teodorescu2023codeplay}
L.~Teodorescu, C.~Colas, M.~Bowers, T.~Carta, and P.-Y. Oudeyer.
\newblock Codeplay: Autotelic learning through collaborative self-play in programming environments.
\newblock In \emph{IMOL 2023-Intrinsically Motivated Open-ended Learning workshop at NeurIPS 2023}, pages 4--16, 2023.

\bibitem[Vygotsky(1978)]{vygotsky1978mind}
L.~S. Vygotsky.
\newblock Mind in society: The development of higher psychological processes, 1978.

\bibitem[Wang et~al.(2022{\natexlab{a}})Wang, Mu, Arumugam, Jaques, and Goodman]{wang2022in}
R.~E. Wang, J.~Mu, D.~Arumugam, N.~Jaques, and N.~Goodman.
\newblock In the {ZONE}: Measuring difficulty and progression in curriculum generation.
\newblock In \emph{Deep Reinforcement Learning Workshop NeurIPS 2022}, 2022{\natexlab{a}}.
\newblock URL \url{https://openreview.net/forum?id=SgBHmHMctfd}.

\bibitem[Wang et~al.(2022{\natexlab{b}})Wang, Kordi, Mishra, Liu, Smith, Khashabi, and Hajishirzi]{Wang2022SelfInstructAL}
Y.~Wang, Y.~Kordi, S.~Mishra, A.~Liu, N.~A. Smith, D.~Khashabi, and H.~Hajishirzi.
\newblock Self-instruct: Aligning language models with self-generated instructions.
\newblock In \emph{Annual Meeting of the Association for Computational Linguistics}, 2022{\natexlab{b}}.
\newblock URL \url{https://api.semanticscholar.org/CorpusID:254877310}.

\bibitem[Wang et~al.(2025)Wang, Mao, Fried, and Neubig]{wang2025agent}
Z.~Z. Wang, J.~Mao, D.~Fried, and G.~Neubig.
\newblock Agent workflow memory.
\newblock In \emph{Forty-second International Conference on Machine Learning}, 2025.
\newblock URL \url{https://openreview.net/forum?id=NTAhi2JEEE}.

\bibitem[Wu et~al.(2026)Wu, Zhang, Shi, Zhang, Lei, Zhou, Wang, Wu, Zhou, Wang, et~al.]{wu2026aspire}
Y.~Wu, J.~Zhang, J.~Shi, Y.~Zhang, X.~Lei, J.~Zhou, Z.~Wang, Y.~Wu, H.~Zhou, D.~Wang, et~al.
\newblock Aspire: Can models self-evolve from vague goals?
\newblock \emph{arXiv preprint arXiv:2608.31111}, 2026.

\bibitem[Xue and Yang(2026)]{xue2026rethinking}
H.~Xue and F.~Yang.
\newblock Rethinking self-evolving agents: Do we still need prescribed optimization pipelines?
\newblock \emph{arXiv preprint arXiv:2608.09629}, 2026.

\bibitem[Ye et~al.(2026)Ye, He, Arak, Dong, and Song]{ye2026meta}
H.~Ye, X.~He, V.~Arak, H.~Dong, and G.~Song.
\newblock Meta context engineering via agentic skill evolution.
\newblock In \emph{Forty-third International Conference on Machine Learning}, 2026.
\newblock URL \url{https://openreview.net/forum?id=P1jHroBS5E}.

\bibitem[Zelikman et~al.(2022)Zelikman, Wu, Mu, and Goodman]{Zelikman2022STaRBR}
E.~Zelikman, Y.~Wu, J.~Mu, and N.~D. Goodman.
\newblock Star: Bootstrapping reasoning with reasoning.
\newblock 2022.
\newblock URL \url{https://api.semanticscholar.org/CorpusID:247762790}.

\bibitem[Zhang et~al.(2025{\natexlab{a}})Zhang, Xiang, Yu, Teng, Chen, Chen, Zhuge, Cheng, Hong, Wang, Zheng, Liu, Luo, and Wu]{zhang2025aflow}
J.~Zhang, J.~Xiang, Z.~Yu, F.~Teng, X.-H. Chen, J.~Chen, M.~Zhuge, X.~Cheng, S.~Hong, J.~Wang, B.~Zheng, B.~Liu, Y.~Luo, and C.~Wu.
\newblock {AF}low: Automating agentic workflow generation.
\newblock In \emph{The Thirteenth International Conference on Learning Representations}, 2025{\natexlab{a}}.
\newblock URL \url{https://openreview.net/forum?id=z5uVAKwmjf}.

\bibitem[Zhang et~al.(2026{\natexlab{a}})Zhang, Hu, Lu, Lange, and Clune]{zhang2026darwin}
J.~Zhang, S.~Hu, C.~Lu, R.~Lange, and J.~Clune.
\newblock Darwin g{\"o}del machine: open-ended evolution of self-improving agents.
\newblock In \emph{International Conference on Learning Representations}, volume 2026, pages 104223--104294, 2026{\natexlab{a}}.

\bibitem[Zhang et~al.(2026{\natexlab{b}})Zhang, Zhao, Yang, Foerster, Clune, Jiang, Devlin, and Shavrina]{zhang2026hyperagents}
J.~Zhang, B.~Zhao, W.~Yang, J.~Foerster, J.~Clune, M.~Jiang, S.~Devlin, and T.~Shavrina.
\newblock Hyperagents, 2026{\natexlab{b}}.
\newblock URL \url{https://arxiv.org/abs/2603.19461}.

\bibitem[Zhang et~al.(2026{\natexlab{c}})Zhang, Hu, Upasani, Ma, Hong, Kamanuru, Rainton, Wu, Ji, Li, Thakker, Zou, and Olukotun]{zhang2026agentic}
Q.~Zhang, C.~Hu, S.~Upasani, B.~Ma, F.~Hong, V.~Kamanuru, J.~Rainton, C.~Wu, M.~Ji, H.~Li, U.~Thakker, J.~Zou, and K.~Olukotun.
\newblock Agentic context engineering: Evolving contexts for self-improving language models.
\newblock In \emph{The Fourteenth International Conference on Learning Representations}, 2026{\natexlab{c}}.
\newblock URL \url{https://openreview.net/forum?id=eC4ygDs02R}.

\bibitem[Zhang et~al.(2025{\natexlab{b}})Zhang, Huang, Li, and Cardie]{zhang2025better}
Z.~Zhang, C.~Huang, A.~O. Li, and C.~Cardie.
\newblock Better llm reasoning via dual-play.
\newblock \emph{arXiv preprint arXiv:2511.11881}, 2025{\natexlab{b}}.

\bibitem[Zhao et~al.(2025)Zhao, Wu, Yue, Wu, Xu, Yue, Lin, Wang, Wu, Zheng, and Huang]{zhao2025absolute}
A.~Zhao, Y.~Wu, Y.~Yue, T.~Wu, Q.~Xu, Y.~Yue, M.~Lin, S.~Wang, Q.~Wu, Z.~Zheng, and G.~Huang.
\newblock Absolute zero: Reinforced self-play reasoning with zero data.
\newblock In \emph{The Thirty-ninth Annual Conference on Neural Information Processing Systems}, 2025.
\newblock URL \url{https://openreview.net/forum?id=neZSGqhxDa}.

\bibitem[Zimmerman(2000)]{zimmerman2000attaining}
B.~J. Zimmerman.
\newblock Attaining self-regulation: A social cognitive perspective.
\newblock In \emph{Handbook of self-regulation}, pages 13--39. Elsevier, 2000.

\bibitem[Zimmerman(2002)]{zimmermanBecoming}
B.~J. Zimmerman.
\newblock Becoming a self-regulated learner: An overview.
\newblock \emph{Theory Into Practice}, 41\penalty0 (2):\penalty0 64--70, 2002.
\newblock ISSN 00405841, 15430421.
\newblock URL \url{http://www.jstor.org/stable/1477457}.

\end{thebibliography}

\newpage

\appendix
\etocsetnexttocdepth{subsection}
\etocsettocstyle{\section*{Appendix}}

\begingroup
\parskip=0pt
\makeatletter
\newcommand{\ifappendixentry}[2]{%
  \begingroup
  \edef\@tempa{\etocthenumber}%
  \expandafter\@checkfirstchar\@tempa\@nil
  \if@isappendix
    \endgroup #1%
  \else
    \endgroup #2%
  \fi
}
\newif\if@isappendix
\def\@checkfirstchar#1#2\@nil{%
  \@isappendixfalse
  \ifnum`#1>`@ \ifnum`#1<`[ \@isappendixtrue \fi\fi
}
\makeatother

\etocsetstyle{section}
  {}
  {}
  {\ifappendixentry
    {\vspace{0.5em}\leftskip=0pt\leavevmode{\bfseries\makebox[2em][l]{\etocnumber}\etocname\nobreak\leaders\hbox{.}\hfill\nobreak\etocpage}\par}
    {}}
  {}
\etocsetstyle{subsection}
  {}
  {}
  {\ifappendixentry
    {\vspace{0.2em}\leftskip=2em\leavevmode\makebox[2.5em][l]{\etocnumber}\etocname\nobreak\leaders\hbox{.}\hfill\nobreak\etocpage\par}
    {}}
  {}

\etocsettocdepth{subsection}
\etocignoretoctocdepth
\tableofcontents
\endgroup

\newpage

\section{Extended Related Work}\label{sec:app-related-work}

We categorize self-improvement methods based on which components of the learning process remain externally specified and how learning takes place. We consider four main axes: \textit{the source of experience}, whether it is externally provided or generated by the learner itself; \textit{the structure of the learning procedure}, whether it is prescribed or organized by the learner; \textit{the source of feedback during learning}, such as externally defined rewards or evaluators versus feedback obtained through the learner's own interactions; and finally, \textit{what is updated}, for example model weights or text-based context. Another axis of comparison is the scope of generalization the proposed methods study: some evaluate improvement within the same task distribution or domain used during optimization, while others test whether the learned adaptation transfers beyond those conditions to new environments and domains.

Asymmetric self-play methods have a teacher generate challenging problems for a student to solve, resulting in an emergent curriculum adapted to the student's needs \citep{sukhbaatar2017asymmetric}. These methods have also found success in post-training LLMs in domains such as coding, math, and scientific reasoning \citep{zhao2025absolute, huang2026rzero, poesia2024_minimo, dong2025_stp, liu2025_spice, zhang2025better,teodorescu2023codeplay}. These are examples of methods where the source of experience is self-generated by the teacher, i.e. the teacher synthesizes which problems the student learns from, while learning then typically proceeds through gradient-based optimization of the underlying model weights. However, the optimization procedure itself remains externally specified, e.g. how many problems the teacher generates within a batch, how often the student is evaluated on a given problem to determine whether they are challenging, and subsequently how exactly the teacher is rewarded. 

Related ideas appear in unsupervised environment design, where environments are generated or selected from a predefined, often parameterized environment space to form curricula adapted to the learner \citep{dennis2020paired, Jiang2020PrioritizedLR, racaniere2020settersolver, jiang2021ued, parker2022evolving, rutherford2024no}. SPADE \citep{Liu2026SPADESI} bridges the gap between UED and asymmetric self-play for LLMs by leveraging the regret principle from UED to train a teacher that produces environments rather than static question-answer tasks. Similarly to other asymmetric self-play works, the optimization procedure and task selection remain externally specified. 

Asymmetric self-play methods can also suffer from diversity or performance collapse \citep{chae2025understandingselfplay}, motivating additional mechanisms for deciding which generated problems provide useful learning experience, for example using judges \citep{bailey2026scaling} or grounding the agent in real-data hard questions as goalposts \citep{Jana2026GASPGA}. These mechanisms can help stabilize self-play, but introduce additional externally specified choices about what constitutes a useful learning experience. It is also possible to address this collapse by providing the model with feedback on whether training on the teacher proposed problems leads to improvements in an external hard subset of problems \citep{sundaram2026teachingmodelsteachthemselves}. In this case, the feedback is no longer purely obtained through the learner's own interactions, while still relying on self-generated data to drive learning.

A separate line of work considers self-improvement through updates to the model's context or agent harness, rather than through updates to the model weights \citep{suzgun2025dynamic, wang2025agent, hu2025automated, zhang2025aflow}. GEPA uses natural-language reflection over execution trajectories to optimize model prompts via evolutionary search \citep{agrawal2026gepa}, while ACE maintains an evolving textual playbook that accumulates strategies and lessons from previous interactions \citep{zhang2026agentic}. In these works, optimization still takes place with respect to an externally specified task distribution, and the procedure through which the context is updated is pre-specified. 
MCE partly relaxes the latter by co-evolving the context with a context-engineering skill that guides how context is represented and updated \citep{ye2026meta}. Meta-Harness expands the object of optimization further to executable harness code, allowing the model to change how context is managed, including what information is stored and retrieved \citep{lee2026metaharness}.
The optimization still uses externally provided tasks and candidate harnesses are evaluated using an externally specified objective. 

Related work on self-modifying agents expands the components under the model's control further. The Darwin Gödel Machine iteratively modifies its own agent code and evaluates the resulting agents on externally specified benchmarks \citep{zhang2026darwin}, while Hyperagents additionally make the meta-level procedure used to generate future improvements itself editable \citep{zhang2026hyperagents}. These approaches therefore allow increasingly large parts of the self-improvement mechanism to change, while still operating within an externally specified task and evaluation setting.

\citet{xue2026rethinking} introduce Open-Ended Optimization, where the objective, available interactions, resource budget, data boundary and evaluation are fixed, but the model is free to organize the improvement process itself. The tasks used during this optimization are still externally provided and the resulting performance is evaluated on held-out examples from the same benchmark setting, rather than studying transfer beyond the environment in which optimization took place.

Closest to our setting, in concurrent work, \citet{wu2026aspire} introduce ASPIRE, a benchmark for self-evolution from vague capability goals. Agents 
(sometimes a stronger teacher model optimizing a smaller student) improve weights or harnesses with access to external data and, in some settings, aggregate scores from the hidden evaluation. Performance is then measured on hidden items from the same goal. In contrast, we study a model improving itself from self-generated experience alone, and evaluate transfer to different environments.

\section{Evaluation Benchmarks}\label{sec:app-datasets}
\begin{figure*}[t]
    \centering
    \includegraphics[width=0.75\linewidth]{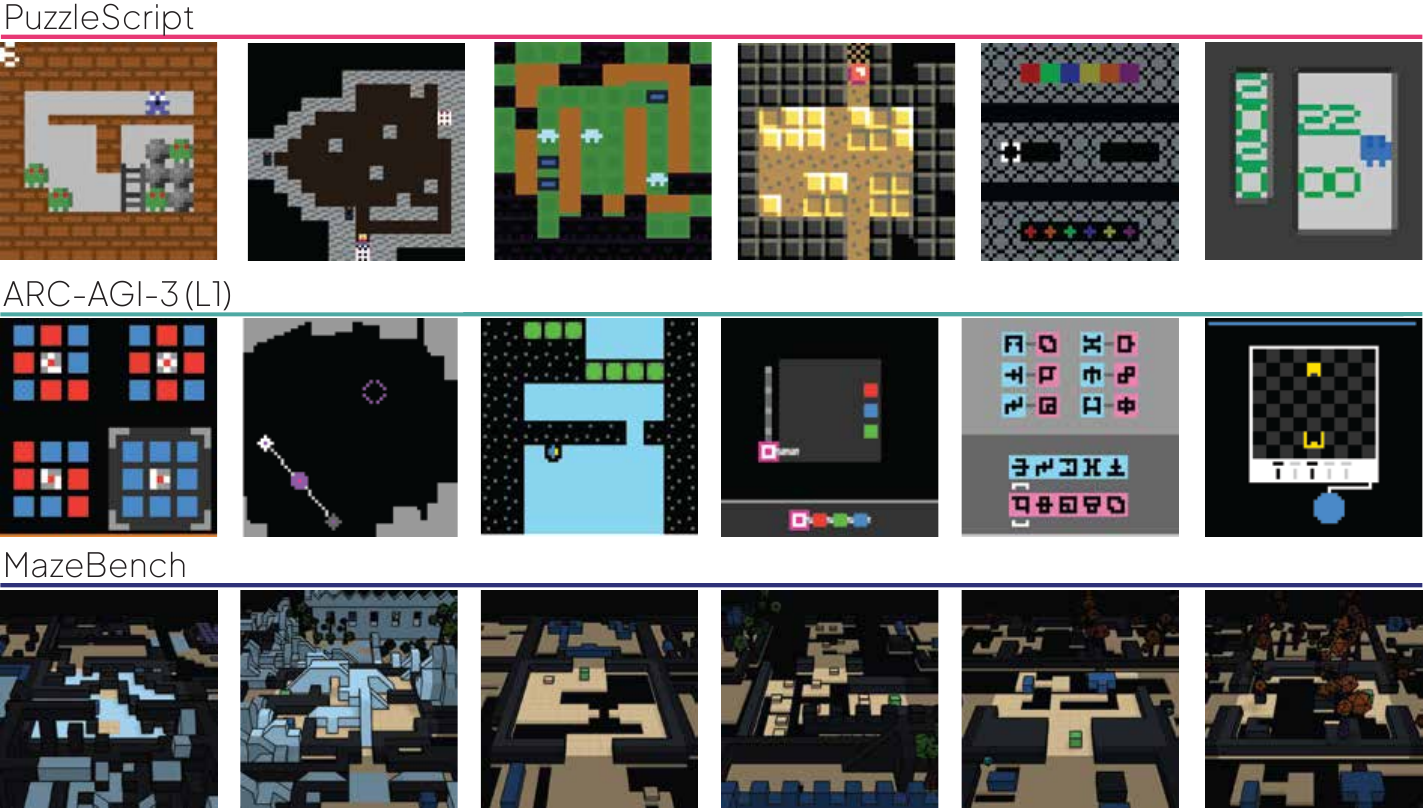}
    \caption{\textbf{Examples of evaluation benchmarks.}}\label{fig:full-game-grid}
    \vspace{-1pt}
\end{figure*}

\begin{table}[h]
\centering
\caption{\textbf{Evaluation settings for each benchmark.}}
\label{tab:app-eval-settings}
\small
\begin{tabular}{lcccc}
\toprule
\textbf{Benchmark} & \textbf{\# Games} & \textbf{Turns ($H$)} & \textbf{Tokens / game} & \textbf{Eval rollouts / game} \\
\midrule
\textsc{PuzzleScript-t20}  & 20 & 20  & 25{,}000 & 4 \\
\textsc{PuzzleScript-t30}  & 20 & 30  & 31{,}250 & 4 \\
\textsc{ARC-AGI-3 (L1)}    & 17 & 30  & 31{,}250 & 4 \\
\textsc{MazeBench}         & 16 & 100 & 75{,}000 & 2 \\
\bottomrule
\end{tabular}

\vspace{4pt}
\end{table}

Examples from each dataset are shown in Figure \ref{fig:full-game-grid}. 

\paragraph{\textsc{PuzzleScript-t20}.} Both PuzzleScript benchmarks are drawn from a large pool of community-authored games \citep{Earle2025PuzzleJAXAB}. We sample a random set of 500 game-levels, and filter for levels solvable in $\leq$ 12 turns. We then further filter out games that \texttt{Claude Opus-5} solves in 8/8 rollouts, leaving a dataset of 20 game-levels, each from unique games. Note that after filtering, we re-run the baseline. 

\paragraph{\textsc{PuzzleScript-t30}.} We follow the same procedure described for \textsc{PuzzleScript-t20}, however apply an optimal solution length filter of 13-22 to capture longer-horizon, more complex game levels. 

\paragraph{\textsc{ARC-AGI-3 (Level 1)}} \citep{foundation2026arc} Due to the token cost of evaluating on ARC-AGI-3, we use Level 1 of each game. There are 25 games offered in the benchmark; as with our PuzzleScript datasets, we filter out game levels that \texttt{Claude Opus-5} solves in 8/8 rollouts, leaving 17 game levels. The initial observation for each game is the full $64 \times 64$ frame rendered as hex digits; subsequent observations list only the changed cells. 

\paragraph{\textsc{MazeBench}} \citep{pappas2026mazebench}. The MazeBench world map contains 256 rooms . Adopting the room-completion metric, we use each room as a single completion task. The player begins as if it had walked into the room, and wins if it exits through a different edge (which requires navigating the room obstacles/mechanics). We filter for rooms that are solvable within 50 turns and choose a random draw of 16 rooms. We use the common SPL (success weighted by path length) metric to measure the player's efficiency in navigating the room. Note that MazeBench is represented differently from our other three benchmarks: an isometric-style ASCII projection from a top-diagonal view. In addition to movement actions, the player has the ability to rotate the camera and view the room from additional angles. MazeBench also includes obstacles unique to the 3D setting: raised floors, ramps, stacked blocks, occlusion, and voids. 

\section{Experiment Details}\label{sec:app-experiments}

\subsection{Teacher-Player Runs}\label{sec:app-teacher-env}

\paragraph{Harness details.} A single run consists of a teacher agent initialized in a persistent harness session, and several short-lived player harness sessions. Claude models run in Claude Code (v2.1.232) and all other models run in Pi (v0.85.1) accessed through OpenRouter. All harness sessions run inside a Singularity container with the following settings:
\begin{itemize}
    \item The session runs in a fresh sandboxed directory viewable to the agent as \texttt{/workspace}.
    \item The harness's default system prompt is replaced with ours (see Appendix \ref{sec:app-prompts}).
    \item Tools are only granted through an allowlist; all other extensions, skills, prompt templates, context files, etc are disabled. This includes tools such as web search, web fetch, subagent, and task-management tools.
    \item Automatic memory is disabled.
    \item Automatic compaction is left on, however due to our teacher and player budgets, no sessions consume enough tokens to trigger compaction.
    \item Interaction with a human operator is disabled; messages to and from the session are managed programmatically.
\end{itemize}

The possible teacher-granted tools are \textbf{Read/Write} tools (including Edit, Glob, and Grep for managing files), \textbf{Bash}, and our custom \textbf{Gameplay} tool.

The teacher working directory is initialized with:
\begin{itemize}
    \item \texttt{player\_folder}: A designated directory for player text artifacts, initialized with \texttt{player\_prompt.md} containing the default player prompt (Appendix \ref{sec:app-prompts}). Teacher updates are tracked via Write/Edit tool calls to \texttt{player\_folder}.
    \item \texttt{puzzlescript\_docs/} containing PuzzleScript documentation for PuzzleScript Gameplay runs, and \texttt{game\_interface.md} detailing the expected Python class interface for Python Gameplay runs (Appendix \ref{sec:app-prompts}).
\end{itemize}

\paragraph{Gameplay.} The teacher calls the Gameplay tool with a source game file (either PuzzleScript or Python) and optionally the path to a folder holding the text artifacts that the teacher wishes to grant the player. If the path is omitted, the player receives the default prompt and no auxiliary files. PuzzleScript games are rejected if they do not compile in the PuzzleScript JavaScript engine, or if a BFS solver verifies that the winning state is unreachable (if the BFS solver times out, the game is accepted). Python games are likewise checked for compilation, and safety checks (network access, file access, etc). The Gameplay tool runs a fresh player session with only Read and Glob tools, and returns the following output to the teacher: 
\begin{itemize}
    \item The player score.
    \item The game end reason (turn budget, token budget, or player won).
    \item The status of the teacher's gameplay and update budgets.
    \item A path to a transcript containing player reasoning, actions, and observations.
\end{itemize}

\subsection{Tool/Budget Awareness Ablations.}\label{sec:app-eval-awareness-ablations}
In Section \ref{sec:information} we ablate the teacher's awareness of the evaluation setting via tool constraints. We run 3 seeds each of the following settings, using the default teacher update budgets:
\begin{enumerate}
    \item \textit{Read-Write}: Teacher has access only to the file-management tools (Read, Write, Edit, Glob, Grep).
    \item \textit{+ Bash}: The above tools, and Bash access (subject to sandbox constraints).
    \item \textit{+ Gameplay (PS)}: The above tools, and a Gameplay PuzzleScript tool. We do not impose a turn or token budget for each gameplay; instead the teacher chooses it, enabling another axis of freedom. Rather than a gameplay budget the teacher is granted a player token budget and player turn budget that it draws upon for each gameplay. We set the turn budget to 3000 and token budget to 3.75M, thus compute-matching the t-20 constrained setting (150 gameplays $\times$ 20 turns/game and 150 gameplays $\times$ 25k tokens/game). 
    \item \textit{+ t20 Token Budget}: This setting is the same as the previous, however we impose the t20 per-game token budget (25k) through the Gameplay tool, and the teacher chooses the game horizon.
    \item \textit{+ t20 Turn Budget:} This is our canonical setting, in which the teacher has a gameplay budget, and the Gameplay tool imposes the t20 turn- and token-budgets (20 turns and 25k tokens).
\end{enumerate}

\subsection{Context Ablations}\label{sec:app-context-ablations}
In Section \ref{sec:information} we vary how much semantic information the teacher receives about an evaluation distribution via context, while holding its tools fixed: \{\textit{Read-Write}, \textit{Bash}, \textit{Gameplay (Python)}, \textit{t20 Token Budget}, \textit{t20 Turn Budget}\}. Since the teacher plays \textbf{python} games in this setting, the written context that it receives is its primary indication of the semantic content of evaluation games.

The \textit{approximation distribution} \textsc{PuzzleScript-Narrow} is a set of 9 PuzzleScript games, chosen using the same filtering criteria as that of the \textsc{PuzzleScript-t20} test set (Section \ref{sec:app-datasets}), with the additional constraint that they are Sokoban-style games. We choose a narrower distribution so that increasing context carries more semantic information. 

We prompt \texttt{Claude Fable-5.1} to produce a summary of the \textsc{PuzzleScript-Narrow} games with the prompt below. We then iteratively prompt it to compress or expand the summary to approximately fit a range of token counts: \{0, 234, 659, 1519, 3313, 6551\}. For each produced summary, we run three teacher sessions with \texttt{Claude Opus-5}. Each teacher session's final context $c_{final}$ is evaluated on \textsc{PuzzleScript-Narrow} and \textsc{ARC-AGI-L1}.

\begin{promptbox}[title={Summarization prompt}]
Below is the complete corpus of a game setting: every game's source, and the opening observation its player receives. The games are written in PuzzleScript, a simple scripting language for 2D puzzle games. Each game is turn-based. On each turn, the player sees an ASCII-grid representation of the game state, and can choose from a set of actions (up, down, left, right, action-button). The game ends when the player meets the game's win condition(s), or runs out of turns.

The player plays BLIND: they are not told the rules of the game or the win condition.

Write the section of a prompt that describes this setting and distribution of games to someone who has never seen them{length}. Do not instruct, provide strategy advice, or name individual games.
\end{promptbox}
With no length constraint (\texttt{\{length\}} empty), this produces a 1519-token description. To obtain longer descriptions, we rerun the same prompt with \texttt{\{length\}} set to ``in about $N$ words'' for $N \in \{1800, 4000\}$, producing descriptions of 3313 and 6551 tokens.

\subsection{Settings}
Our settings for teacher-player runs are shown in Table \ref{tab:app-session-params}. Ablations from our main experiments are noted in the table. We also ablate the teacher update and gameplay budgets in Appendix \ref{sec:app-compute}.

\begin{table}[h]
\centering
\small
\begin{tabular}{ll}
\toprule
Setting & Default Value \\
\midrule
Teacher update budget & 20 \\
Teacher gameplay budget & 150 \\
Max turns per game & \{20,30\} \\
Max output tokens per game (summed over turns) & \{25000,31250\} \\
Output-token limit per model response (teacher and player) & 20000 \\
Reasoning effort (teacher and player) & low \\
\bottomrule
\end{tabular}
\caption{Teacher--player session settings.}
\label{tab:app-session-params}
\end{table}

\subsection{Prompts}\label{sec:app-prompts}

\paragraph{Default player prompt}
The default player system prompt is shown below, and is used for all baselines. The last line lists the artifact files when the player is given any.
\begin{promptbox}[title={Player system prompt}]
You are playing a game. Each turn, end your response with:

ACTION: the action you take this turn (just the action, nothing else).
Your working directory is /workspace.
There are no files in your working directory.
\end{promptbox}

\paragraph{Teacher prompt.}
The teacher's default system prompt is shown below.

\begin{promptbox}[title={Teacher system prompt}]
Your goal is to improve an agent's ability to play unfamiliar single-player games. 
All games have the following attributes:
- Single player.
- At each player turn, the player takes an action, and receives feedback on the game state.
- The game has an objective and a score.
- The player plays without full knowledge of the rules, and discovers the game mechanics through play.

The player has basic file read tools.

You can write whatever artifacts you want for the player to player_folder/. ONLY write artifacts for the player to player_folder/. It is the player's working directory. The player prompt is in player_folder/player_prompt.md, and may be edited. 
**Generalizability.** The goal is to help a blind player do well on NEW games of the same style/form but DIFFERENT tasks and subject matter. Your artifacts must TRANSFER to improve performance on unseen games.

\end{promptbox}
The session opens with the user message \texttt{Begin.} If the teacher stops before either the update or gameplay budget is spent, it is sent:
\begin{promptbox}[title={Continuation message}]
Keep working. Revise your artifacts further. Revisions remaining: <N>.
\end{promptbox}

\paragraph{PuzzleScript Gameplay.} In runs with PuzzleScript Gameplay enabled, the following line is added to the system prompt: \texttt{"The PuzzleScript language reference is in puzzlescript\_docs/. Do NOT place your generated games or output files in player\_folder/."} The PuzzleScript language reference in \texttt{puzzlescript\_docs/} is taken from the official PuzzleScript documentation (\url{https://www.puzzlescript.net/Documentation/documentation.html}).

The teacher is also shown the following description for the Gameplay tool alongside the other tools presented by the harness. The description follows the tool format mandated by the harness (Claude Code for Claude models, and Pi for all other models), and is presented alongside the other available tools (Read, Write, etc.) 
\begin{promptbox}[title={Gameplay tool description}]
play_own_game: Play one game you wrote with a fresh player agent, and wait for it to finish. Several of these can run at once.
  game_file (required): path to the PuzzleScript file holding your game; must be outside player_folder/
  level: index of the level in that file to play; defaults to 0
  out (required): path this play's full result is written to; must be outside player_folder/
  player_folder: path to the folder of artifacts to give the player; omit it to play with none
\end{promptbox}
In ablations where the teacher may choose the turn- and token-budget for the game, we add arguments to the Gameplay tool accordingly.

For GPT and GLM teachers (Pi harness) we make minor adjustments to resolve initially-observed malformed tool calls. The Gameplay tool description adds \textit{``Pass player\_folder to give your artifacts to the player; omit the player\_folder argument to play with the default prompt."}. 

All teachers receive the remaining gameplay and update budgets in every GamePlay tool result. In initial runs GPT and GLM teachers misinterpreted this feedback, so for these models we also state the budget in the system prompt: \textit{``There are two budgets: 150 game plays and 20 revisions. A revision is any tool call that changes /workspace/player\_folder/; writing two files in one turn spends two. Every play result reports the revisions left and gameplays used. The run ends when either budget is exhausted."}

\paragraph{Python Gameplay.}
In our Python ablations the teacher issues games as Python classes instead of PuzzleScript text files, thus removing implicit constraints on the generated game type. In this case, the teacher system prompt replaces the PuzzleScript documentation files in the teacher sandbox with a Markdown file specifying the Python game interface, and replaces the line pointing to PuzzleScript documentation with \textit{``The game interface is in game\_interface.md."}  

\begin{promptbox}[title={game\_interface.md}]
# Game interface

Games are written as a self-contained class named `GameEnvironment` in a Python file.
`__init__`, `step`, `score` and `example_actions` are required; `initial_observation` is optional.

```python
class GameEnvironment:
    def __init__(self):
        self.done = False  # set to True to end the game
        ...

    def step(self, action: str) -> str:
        # `action` is a player-provided string. Parse it and handle unexpected input without
        # crashing. Return the feedback string: this is the player's only window into the game.
        ...

    def score(self) -> float:
        # The player's objective, queryable at any time. A number or a bool. The player never
        # sees this -- put anything you want it to know in step()'s feedback.
        ...

    def example_actions(self) -> list[str]:
        # A few representative valid actions, used only to test the game.
        ...

    def initial_observation(self) -> str:
        # OPTIONAL: what the player is shown at turn 0.
        ...
```
\end{promptbox}

\subsection{Self-Regulated Learning Annotations}\label{sec:app-srl}
We score each teacher session using a rubric adapted from Zimmerman's cyclical phases model of self-regulated learning \citep{zimmerman2000attaining}. The rubric is initially adapted from \cite{zimmerman2000attaining,zimmermanBecoming} using \texttt{GPT 6-Astra}, and then reviewed and refined by the authors. The unit of analysis is the teacher-player system. We annotate using \texttt{GPT 6-Astra} as a judge (medium reasoning effort). 

The rubric covers 13 subprocesses across three phases. Each subprocess is scored on a 0-3 scale grading its presence and quality:
\begin{itemize}
    \item \textbf{0, nonfunctional or omitted:} the subprocess is missing, unused, or carried out in a way that defeats its purpose.
    \item \textbf{1, weak:} the function is recognizable, but substantial gaps limit its usefulness.
    \item \textbf{2, adequate:} it serves its function with suitable specificity, fit, and execution.
    \item \textbf{3, strong:} it serves its function consistently across relevant opportunities, and handles important limitations and contrary evidence well.
\end{itemize}
Each item has descriptions for scores 1-3; the full rubric is in Appendix \ref{sec:app-srl-rubric}. An item may be marked \emph{unassessable}, but only when no defensible score can be assigned. Note that the original Zimmerman framework contains 17 subprocesses, however we filter out 4 that the judge consistently judged as unassessable across all models in initial pilots. 

Each teacher session is judged in batches. A single batch is one round of teacher actions (e.g., Bash commands, Gameplays, thinking) ending with a revision made to the player text artifacts. For each batch, we adopt a two-pass judging protocol:
\begin{itemize}
    \item \textbf{Pass 1:} The judge receives an initial input containing the rubric, a short description of the teacher-player setup, the teacher's system prompt and resource budgets, a record of teacher actions for the current batch, and an index of the batch gameplays. To contextualize the current batch we include the record of teacher actions from previous batches. The judge returns requests for additional evidence to support judgements, such as the full inputs/outputs of teacher tool calls, and gameplay transcripts.
    \item \textbf{Pass 2: } The judge receives the requested evidence and returns a score for each of the 13 subprocesses, along with cited evidence, counterevidence, and a short rationale. 
\end{itemize}
For each teacher session we average each subprocess score over all batches.

\section{Further Results}

\subsection{GEPA v. Self-Organized Learning}
We evaluate if self-organized self-improvement with self-generated games can attain similar performance gains, under a matched rollout budget, to a prescribed text-optimization loop that learns with 20 real games (Figure \ref{fig:gepa}). GEPA optimizes player system prompt, initialized with the default prompt in \ref{sec:app-prompts}. 

We draw 10 training and 10 validation game levels from the same pool of human-authored PuzzleScript games used to construct \textsc{PuzzleScript-t20}, using the same filtering procedure described in Section \ref{sec:app-datasets}. We furthermore ensure that none of the train or validation game levels share the same base game as any of the game levels in the \textsc{PuzzleScript-t20} test set. 

We run GEPA using \texttt{Claude Opus-5} as both the player and the reflection model. We use the procedure and default hyperparameters described in \cite{agrawal2026gepa}. All budget constraints are set to match the budget constraints of the self-organized setting (Table \ref{tab:app-session-params}). We halt GEPA runs at the point before a new iteration (minibatches + a possible validation pass) would exceed the 150 gameplay budget. 

\begin{figure*}[h]
    \centering
    \includegraphics[width=0.5\linewidth]{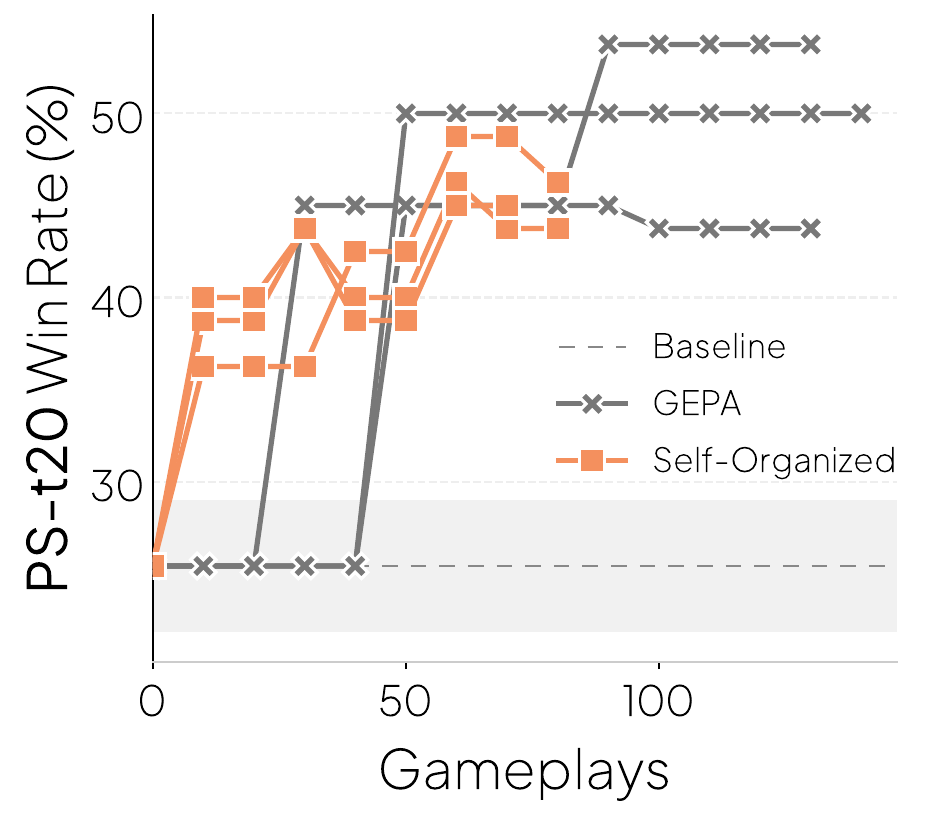}
    \caption{\textbf{Individual seeds for GEPA and self-organized learning runs}. }
    \label{fig:gepa-seeds}
    \vspace{-1pt}
\end{figure*}

In Figure \ref{fig:gepa-seeds} we show results across individual seeds for GEPA and Self-Organized learning.

\subsection{SRL Annotations across seeds}\label{sec:app-srl-seeds}
\begin{figure*}[h]
    \centering
    \setlength{\tabcolsep}{0pt}
    \begin{tabular}{ccc}
    \includegraphics[height=0.3\linewidth]{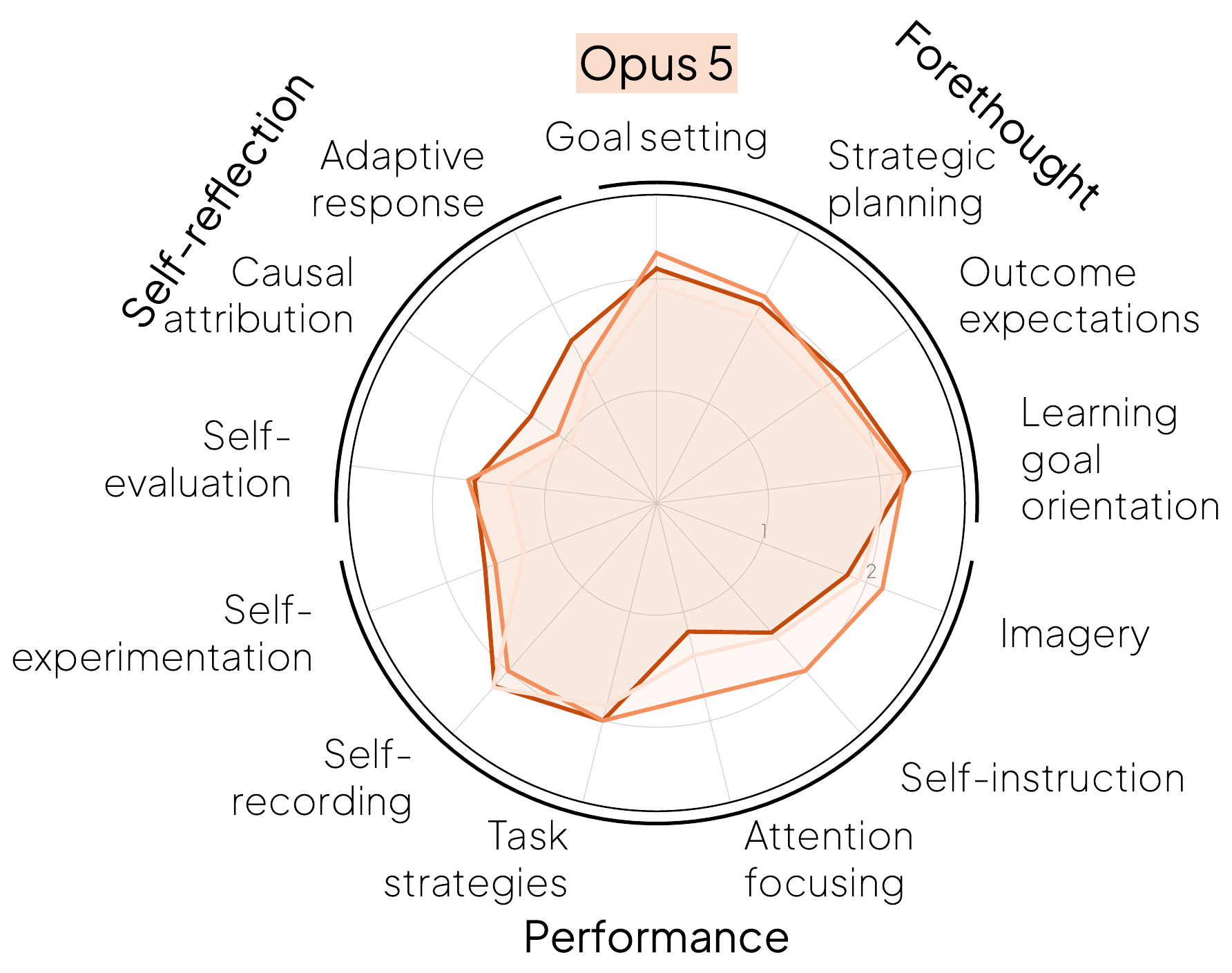} &
    \includegraphics[height=0.3\linewidth]{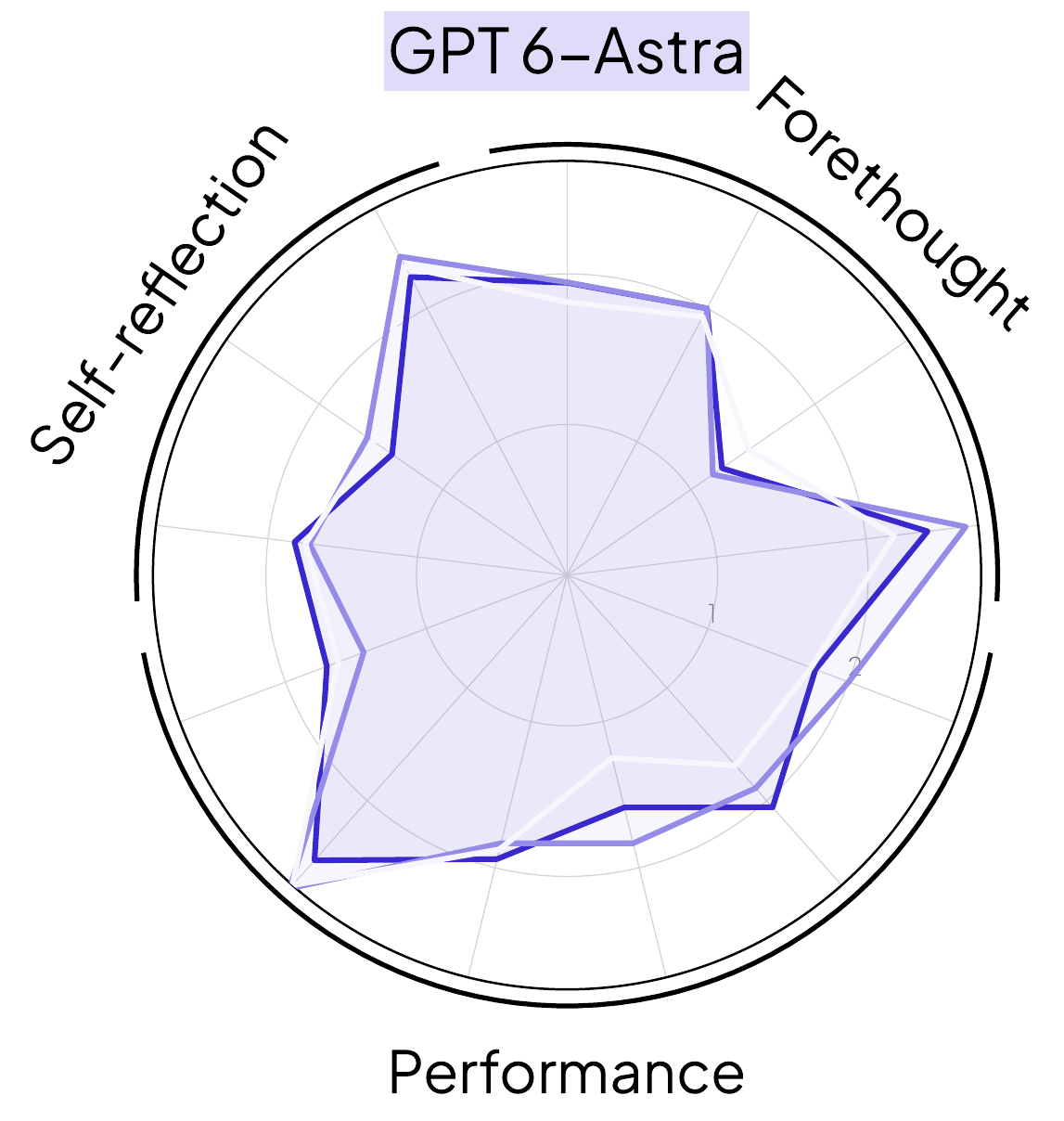} &
    \includegraphics[height=0.3\linewidth]{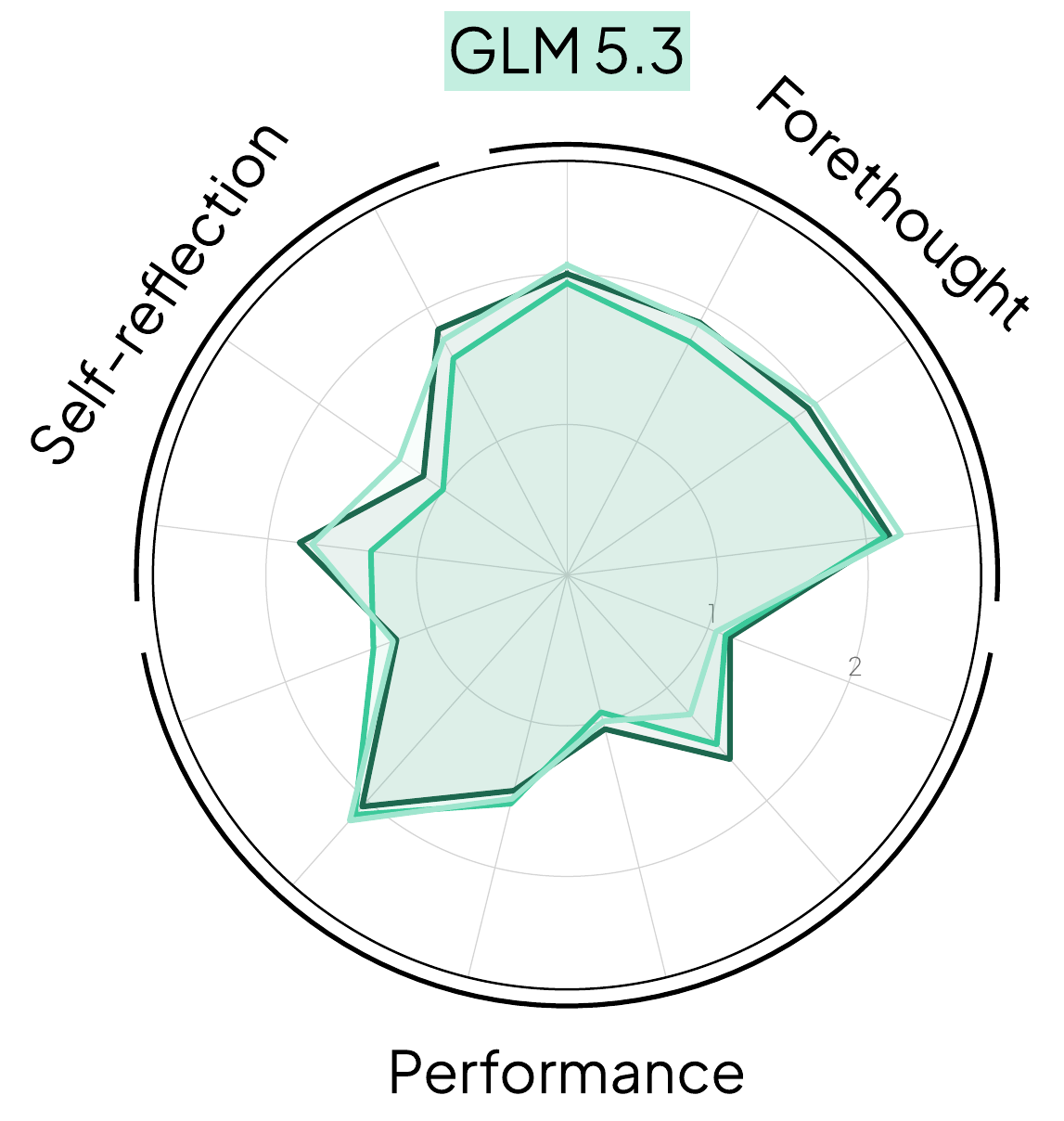} \\
    \includegraphics[height=0.3\linewidth]{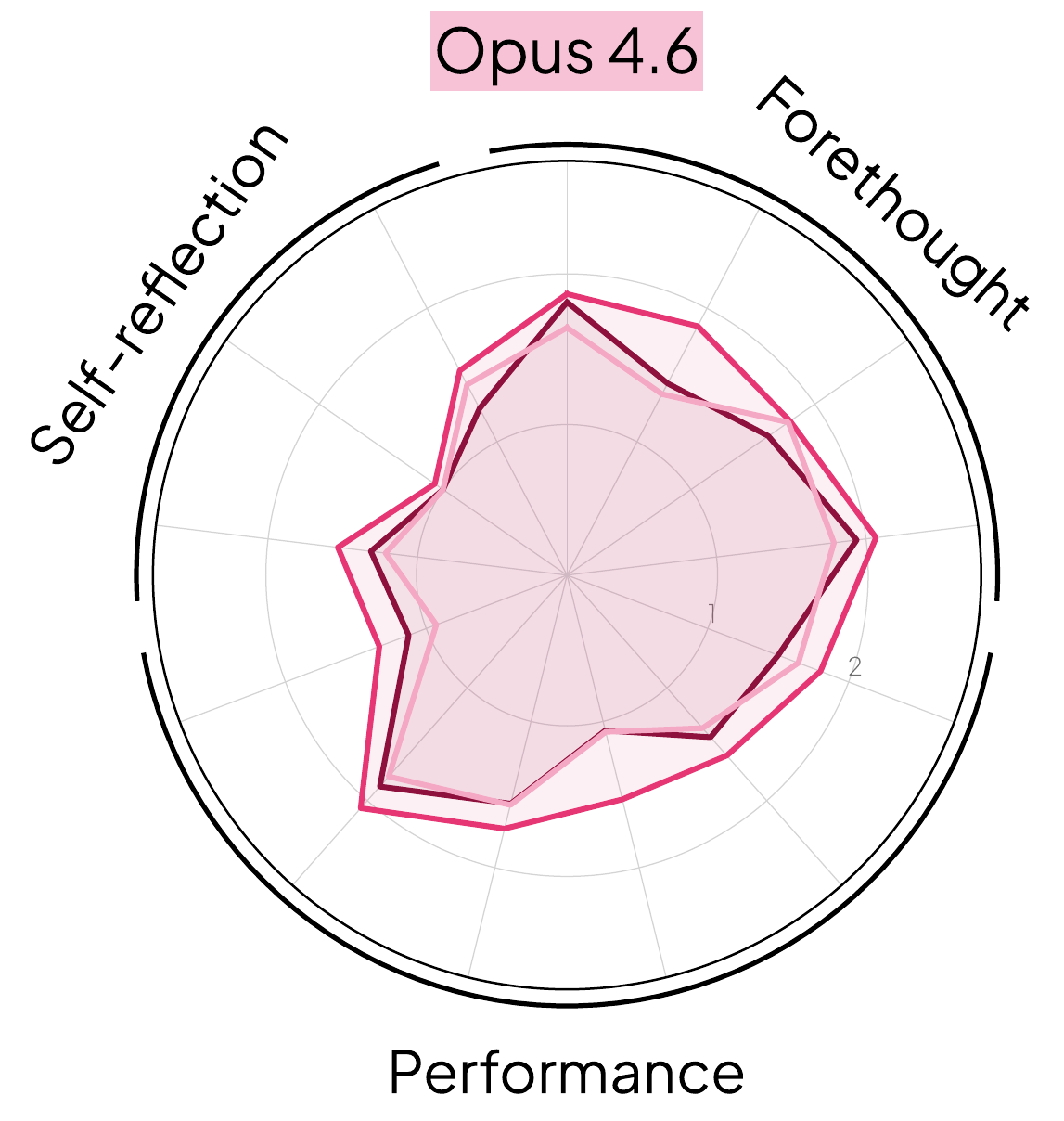} &
    \includegraphics[height=0.3\linewidth]{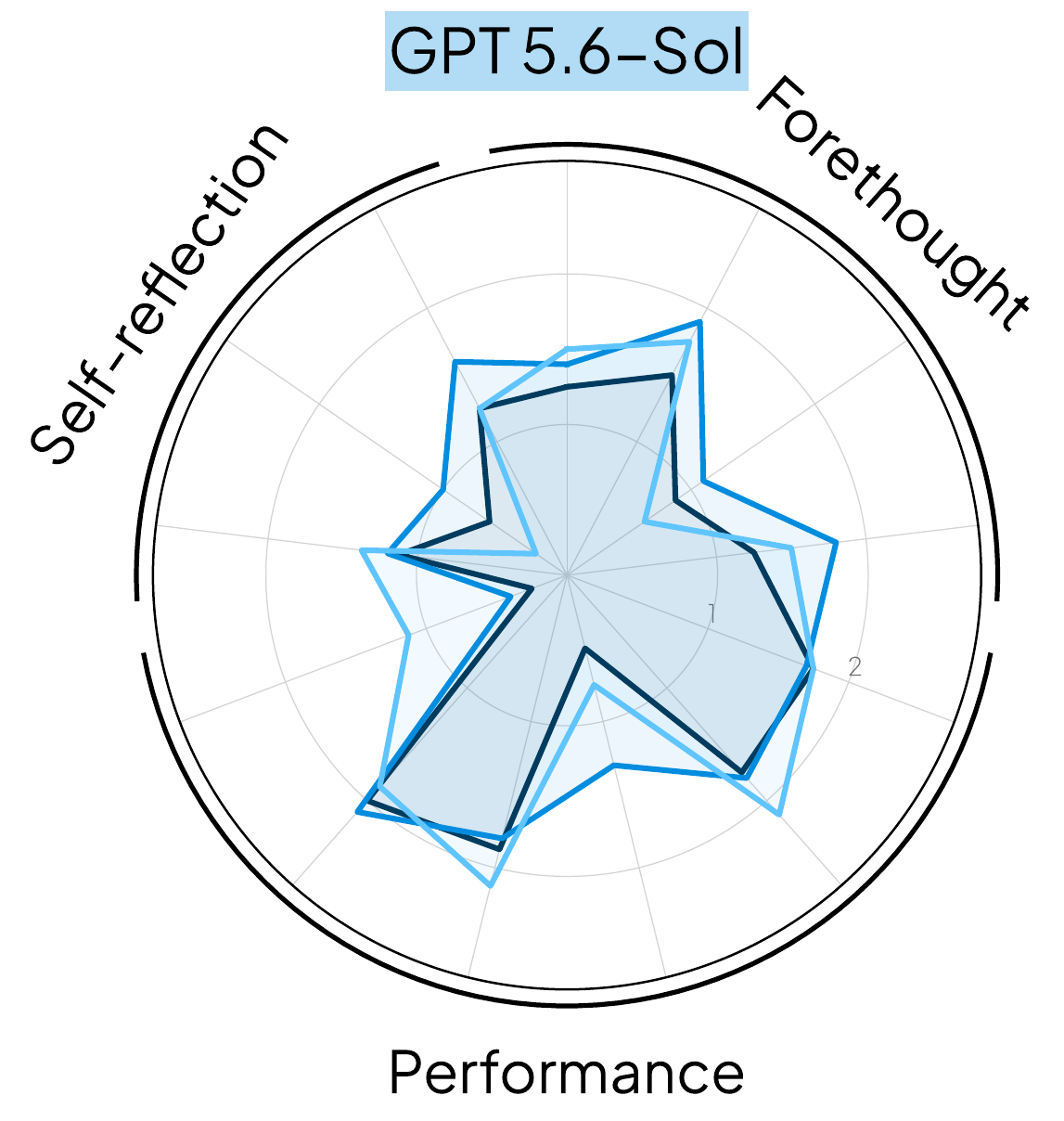} &
    \includegraphics[height=0.3\linewidth]{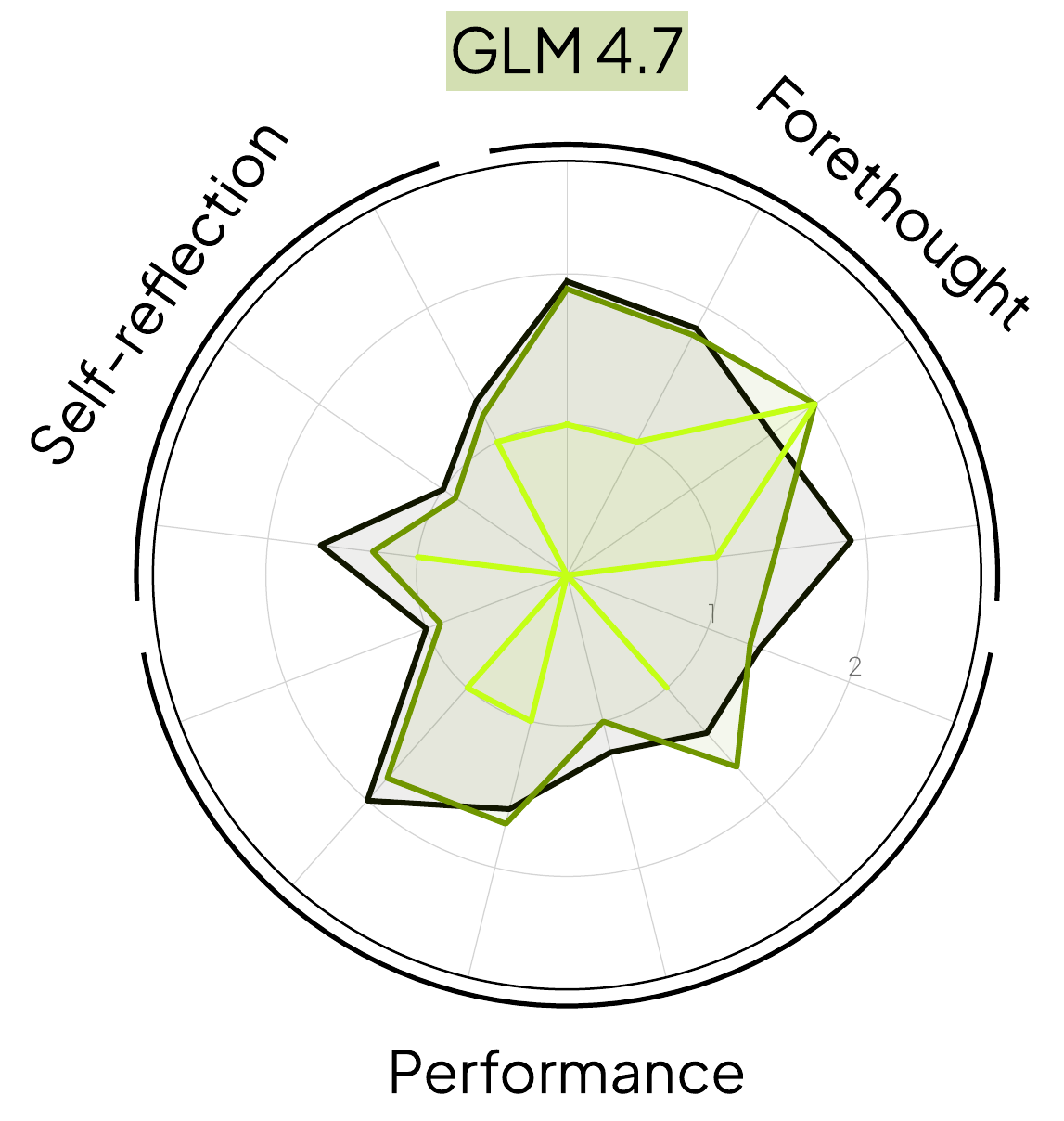} \\
    \end{tabular}
    \caption{\textbf{SRL annotations across independent teacher sessions}. }
    \label{fig:srl-seeds}
    \vspace{-1pt}
\end{figure*}

In Figure \ref{fig:srl-seeds} we plot alignment to our adapted SRL rubric for individual teacher sessions across six models, indicating stable behaviors across runs. 

\subsection{Does self-organized self-improvement benefit from more compute?}\label{sec:app-compute}
\begin{figure*}[h]
    \centering
    \includegraphics[width=0.32\linewidth]{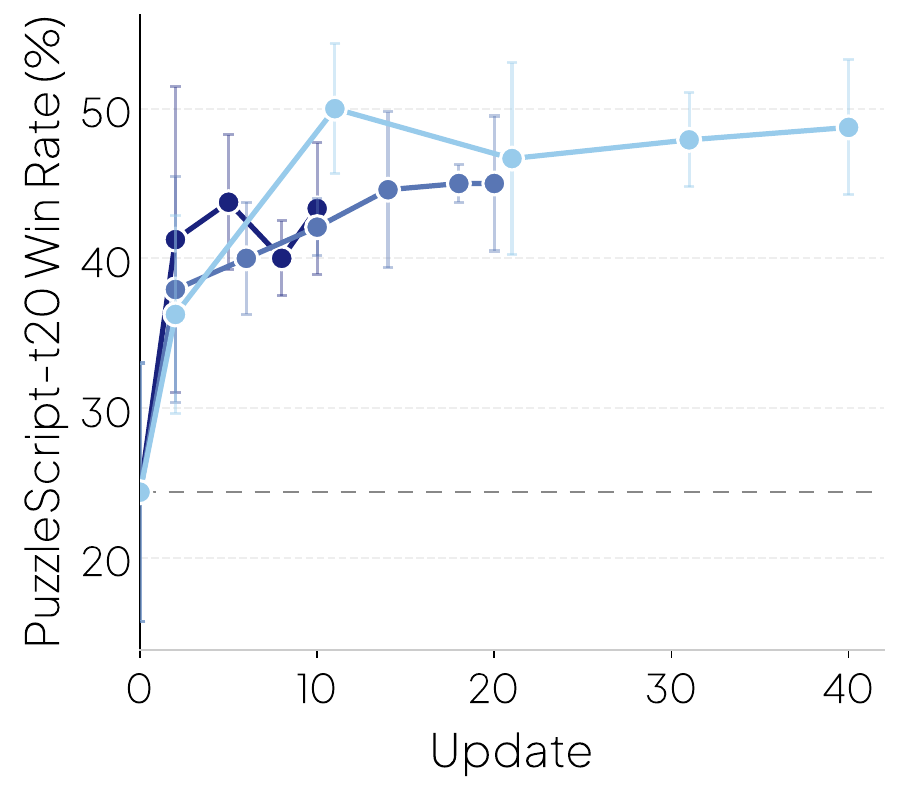}\hfill
    \includegraphics[width=0.34\linewidth]{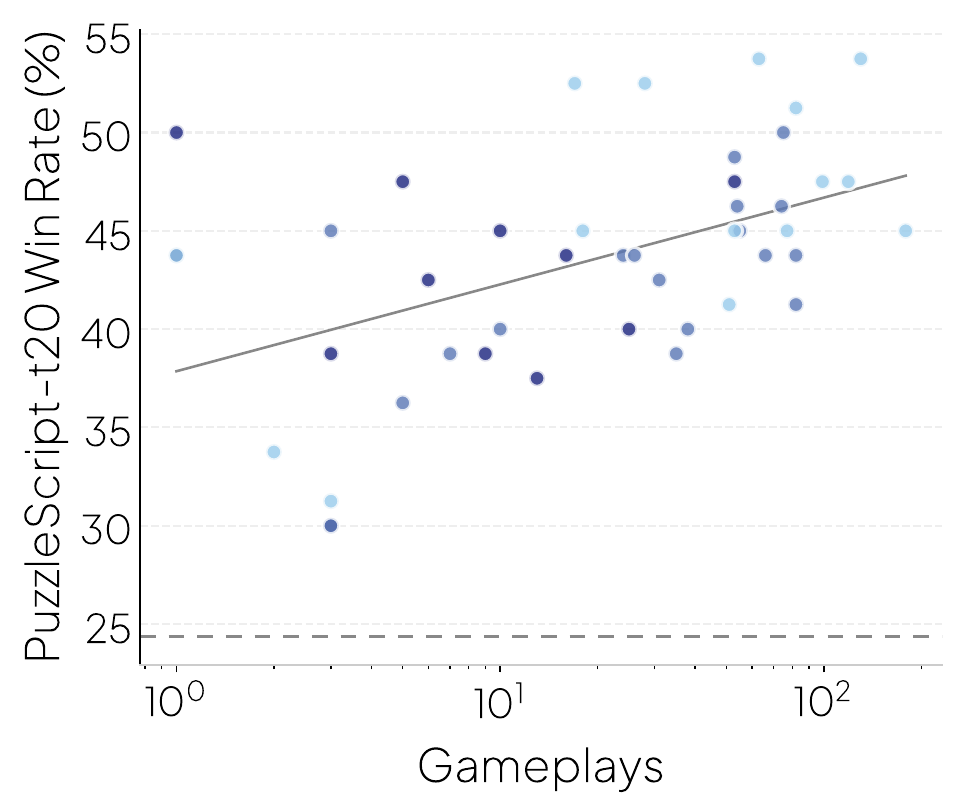}\hfill
    \includegraphics[width=0.335\linewidth]{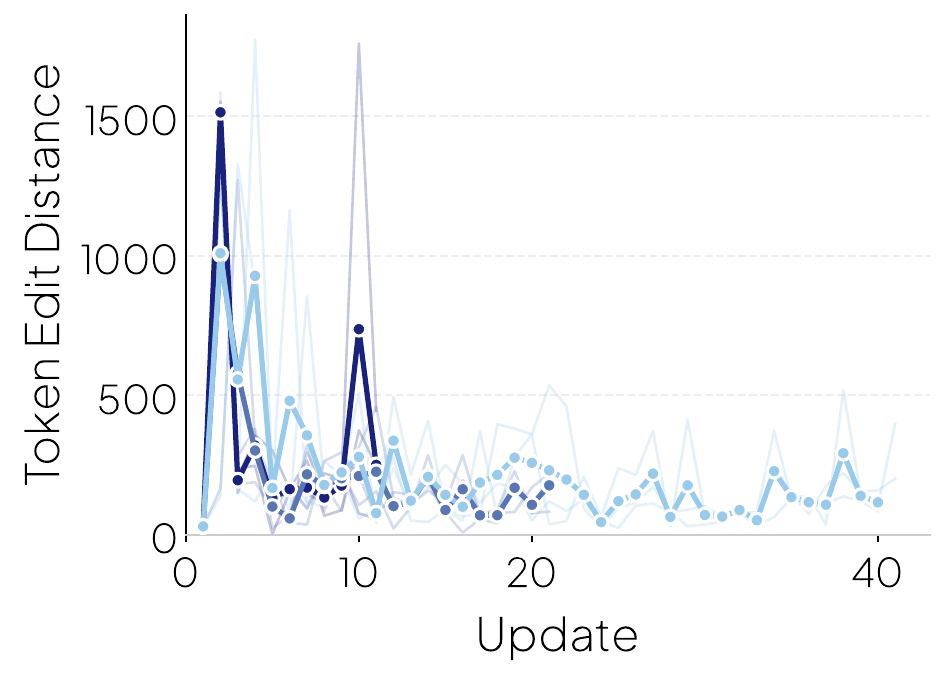}\\[2pt]
    \includegraphics[width=0.58\linewidth]{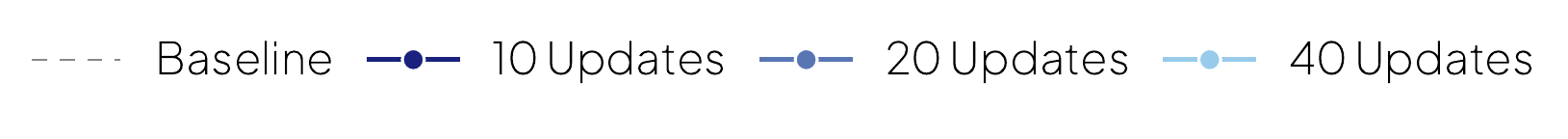}
    \caption{\textbf{Ablating the teacher budgets}. (left) Performance increases and then plateaus with extra compute. (middle) Performance on the \textsc{PuzzleScript-t20 test set} scales with more gameplay experience. (right) The teacher's knowledge of its own budget status leads to different behaviors under different resource regimes.}
    \label{fig:update-ablation}
    \vspace{-1pt}
\end{figure*}

In Figure \ref{fig:update-ablation} we study the impact of running teacher sessions with increasing update budgets (gameplay budgets are scaled accordingly). Performance increases and then plateaus with extra updates. We also note that teacher behavior, measured through the magnitude of updates over time, fluctuates under different resource regimes, and finally observe that performance scales with increased gameplay experience.

\subsection{Performance on saturated games}\label{sec:app-saturated}
Here, we verify that learned artifacts do not degrade performance on previously-saturated games. We sample 10 PuzzleScript game levels on which the baseline \texttt{Claude Opus-5} player achieves a 100\% win-rate over 8 rollouts, and then evaluate the final artifacts from the three main \texttt{Claude Opus-5} teacher sessions. Performance remains near ceiling (Table \ref{tab:saturated}).

\begin{table}[h]
    \centering
    \small
    \begin{tabular}{lcc}
    \toprule
    Game level & Baseline & Learned \\
    \midrule
    Bricks and Hammer (L10) & 100.0 & 100.0 $\pm$ 0.0 \\
    Lights Out (L0) & 100.0 & 100.0 $\pm$ 0.0 \\
    Philosophy (L61) & 100.0 & 91.7 $\pm$ 7.2 \\
    Power Problems (L0) & 100.0 & 100.0 $\pm$ 0.0 \\
    Puzzle Script Tutorial (L15) & 100.0 & 100.0 $\pm$ 0.0 \\
    STUCK IN A HOLE (L10) & 100.0 & 100.0 $\pm$ 0.0 \\
    STUCK IN A HOLE (L7) & 100.0 & 100.0 $\pm$ 0.0 \\
    Sokofun Clone (L0) & 100.0 & 95.8 $\pm$ 7.2 \\
    Spacekoban (L0) & 100.0 & 100.0 $\pm$ 0.0 \\
    Sweet Hints (L0) & 100.0 & 100.0 $\pm$ 0.0 \\
    \midrule
    Overall & 100.0 & 98.8 $\pm$ 0.0 \\
    \bottomrule
    \end{tabular}
    \caption{\textbf{Win rate (\%) on saturated game levels.} Error is $\pm$ 1 std over three \texttt{Claude Opus-5} teacher seeds.}
    \label{tab:saturated}
\end{table}

\subsection{Teacher-observed improvement v. actual improvement}\label{sec:app-teacher-obs-improvement}
\begin{figure*}[h]
    \centering
    \includegraphics[width=0.75\linewidth]{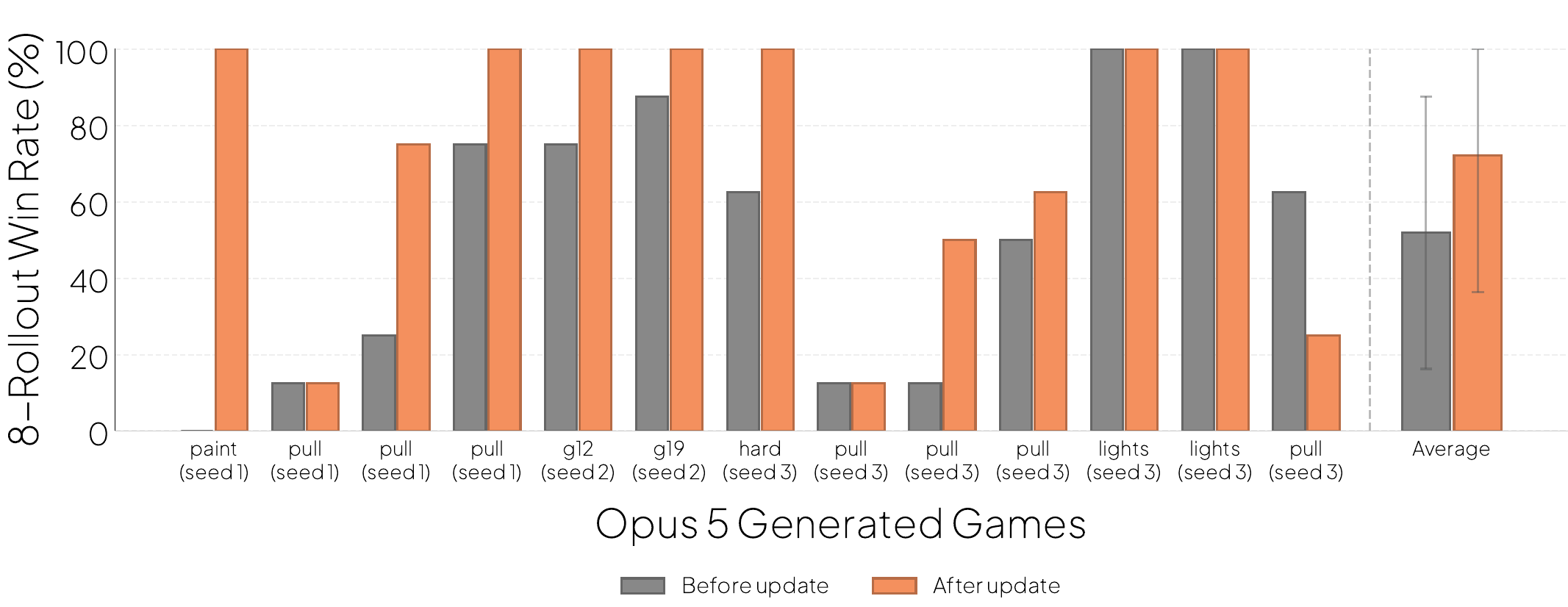}
    \caption{\textbf{Measured performance gains on \texttt{Claude Opus-5} teacher-generated games.} Teacher-observed improvement correlates on average to empirically measured improvement over 8 rollouts.}
    \label{fig:flip-update-eval}
    \vspace{-1pt}
\end{figure*}

In Figure \ref{fig:alg-measures} we count, for each model, the number of batched updates that resulted in a higher teacher-observed pass rate on a generated game. Here we assess if the teacher-observed improvement correlates to actual improvement. For each identified update-game pair across the three \texttt{Claude Opus-5} teacher sessions, we evaluate performance over 8 independent player rollouts on the game, with the artifact state \textit{before} and \textit{after} the identified update. For the majority of self-generated games, performance improves after the identified update (Figure \ref{fig:flip-update-eval}). 

\subsection{Optimal solution length of generated games}
\begin{figure*}[h]
    \centering
    \includegraphics[width=0.33\linewidth]{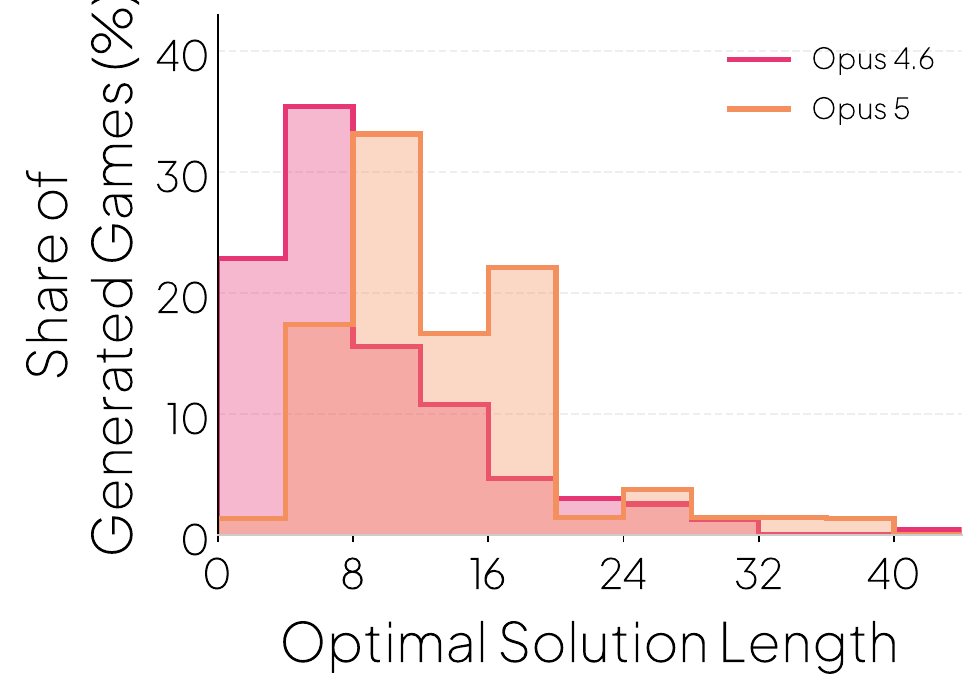}\hfill
    \includegraphics[width=0.33\linewidth]{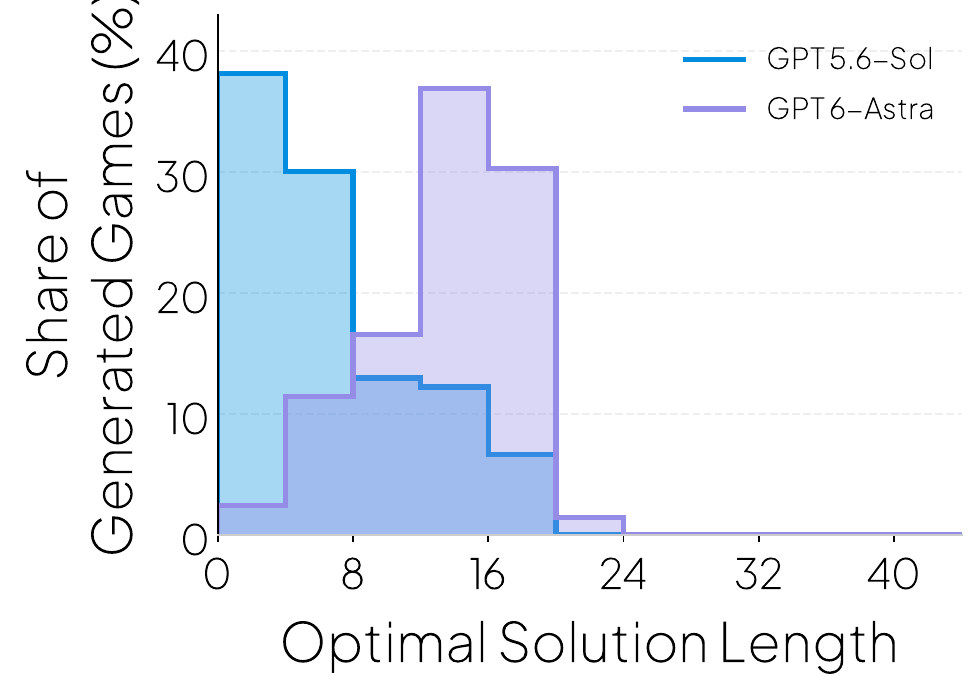}\hfill
    \includegraphics[width=0.33\linewidth]{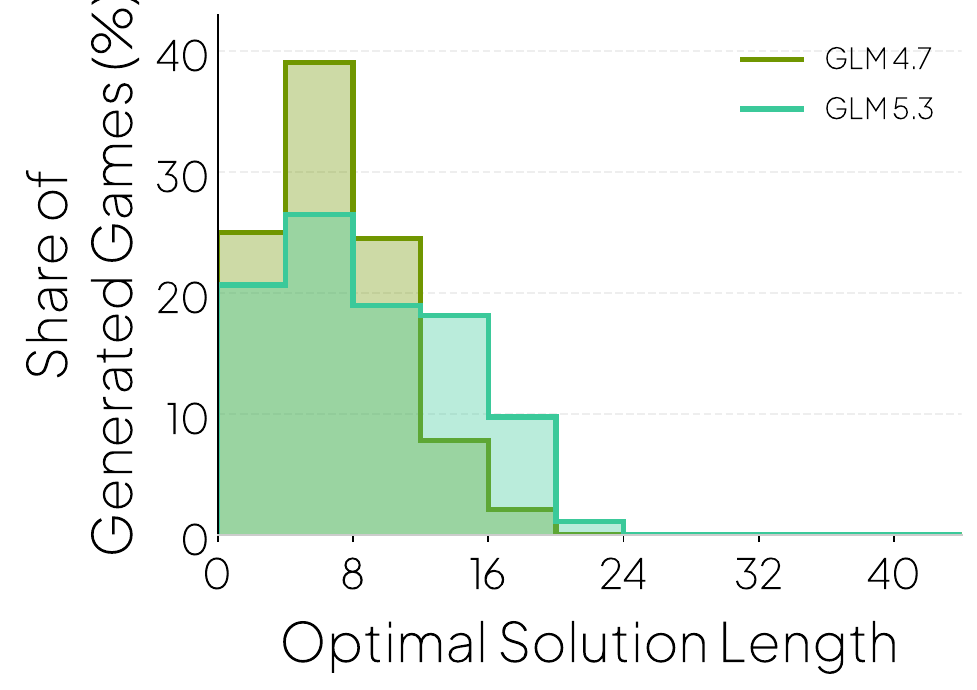}
    \caption{\textbf{Optimal solution lengths of teacher-generated games}, averaged over three teacher seeds per model.}
    \label{fig:opt-sol-len}
    \vspace{-1pt}
\end{figure*}

In Figure \ref{fig:opt-sol-len} we plot histograms over optimal solution lengths for teacher-generated games. Optimal solution lengths are computed with a BFS solver. Each histogram is averaged over the three teacher seeds for that model. Newer models in each family produce games with longer solution lengths. While this is not a perfect proxy for game complexity, it provides one indicator. 

Separation is especially strong for \texttt{Claude Opus-5} and \texttt{GPT 6-Astra}. For both models, some teacher sessions wrote procedural game generators and used BFS solvers to calibrate generated games to longer solution lengths (Figure \ref{fig:timeline}, Section \ref{sec:app-qualitative}), likely contributing to the depicted difference versus older models of the same family.

\subsection{Teacher Session Timelines}\label{sec:app-full-timelines}
\begin{figure*}[h]
    \centering
\includegraphics[width=1.0\linewidth]{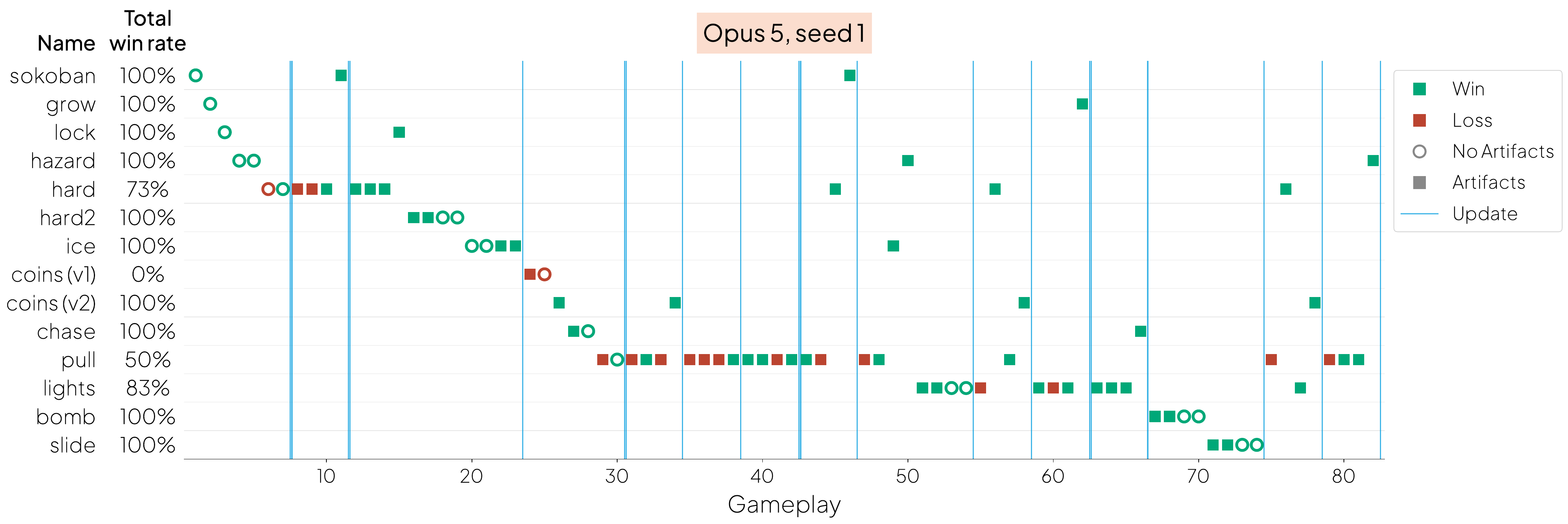}
    \caption{\textbf{Timeline for \texttt{Claude Opus 5}, seed 1.}}
    \label{fig:timeline-opus5-1}
\end{figure*}

\begin{figure*}[h]
    \centering
\includegraphics[width=1.0\linewidth]{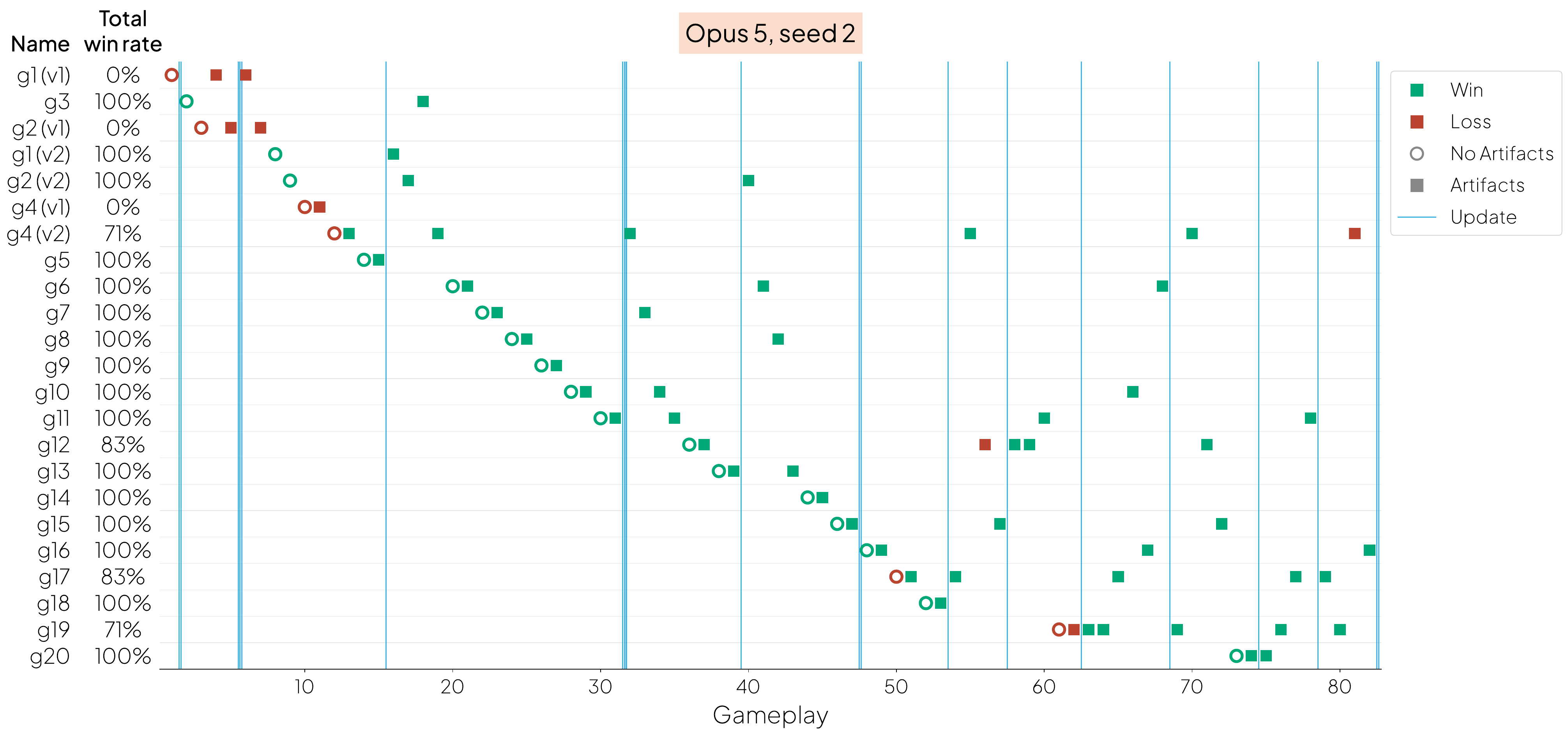}
    \caption{\textbf{Timeline for \texttt{Claude Opus 5}, seed 2.}}
    \label{fig:timeline-opus5-2}
\end{figure*}

\begin{figure*}[h]
    \centering
\includegraphics[width=1.0\linewidth]{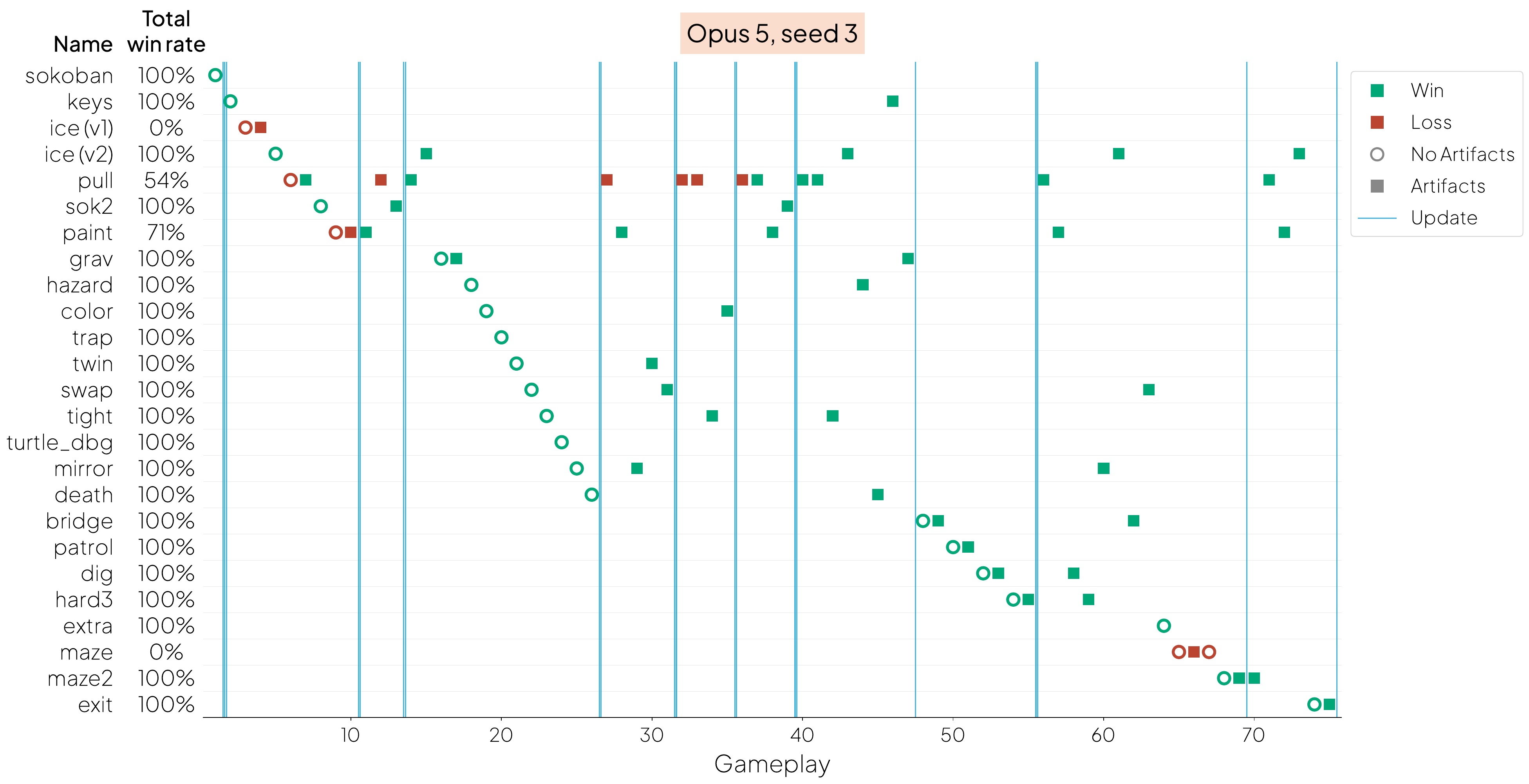}
    \caption{\textbf{Timeline for \texttt{Claude Opus 5}, seed 3.}}
    \label{fig:timeline-opus5-3}
\end{figure*}

\begin{figure*}[h]
    \centering
\includegraphics[width=1.0\linewidth]{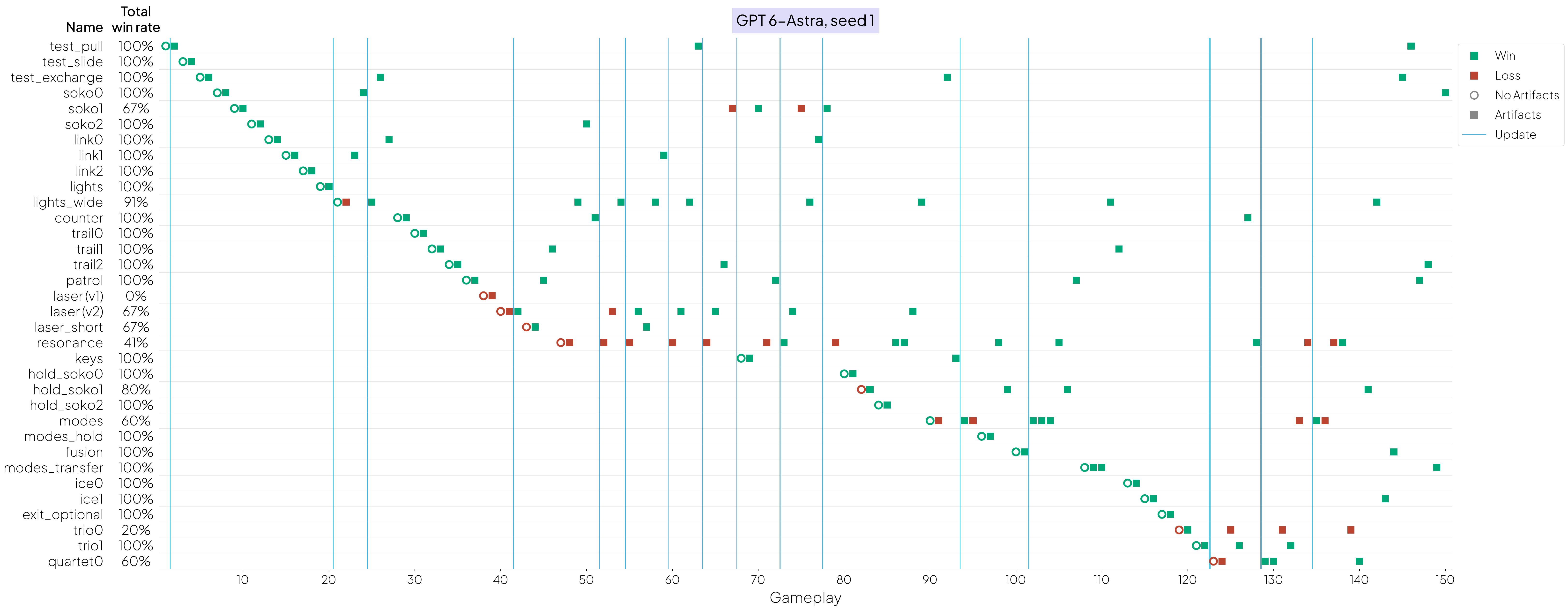}
    \caption{\textbf{Timeline for \texttt{GPT 6-Astra}, seed 1.}}
    \label{fig:timeline-gpt6-astra-1}
\end{figure*}

\begin{figure*}[h]
    \centering
\includegraphics[width=1.0\linewidth]{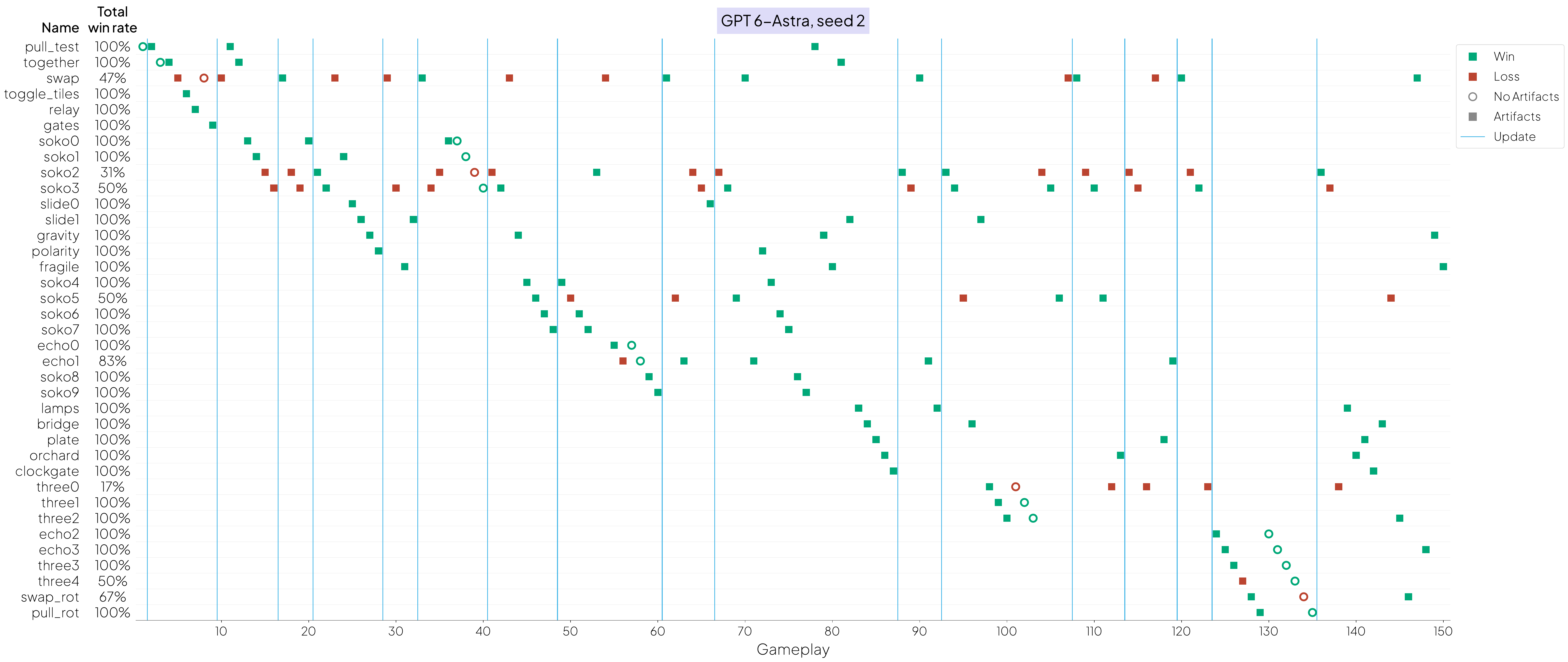}
    \caption{\textbf{Timeline for \texttt{GPT 6-Astra}, seed 2.}}
    \label{fig:timeline-gpt6-astra-2}
\end{figure*}

\begin{figure*}[h]
    \centering
\includegraphics[width=1.0\linewidth]{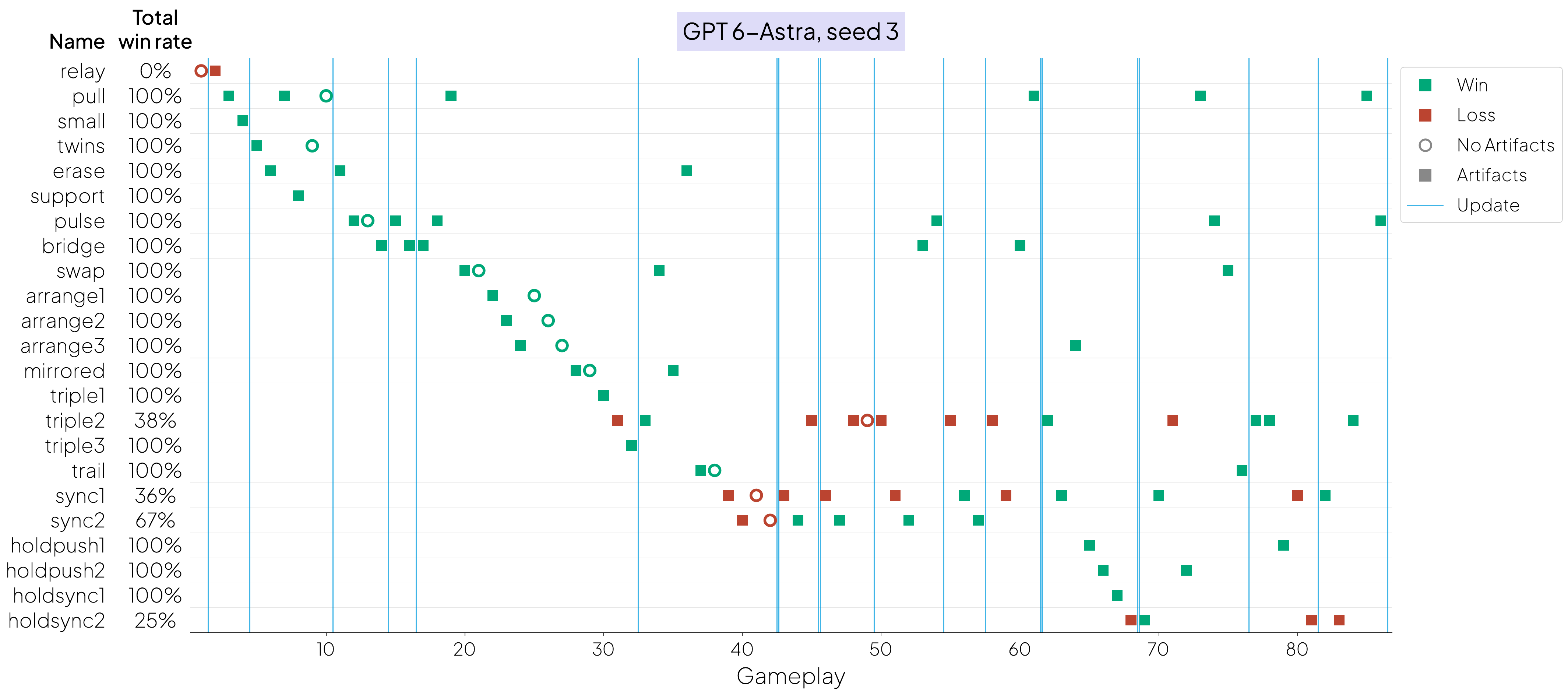}
    \caption{\textbf{Timeline for \texttt{GPT 6-Astra}, seed 3.}}
    \label{fig:timeline-gpt6-astra-3}
\end{figure*}

\begin{figure*}[h]
    \centering
\includegraphics[width=1.0\linewidth]{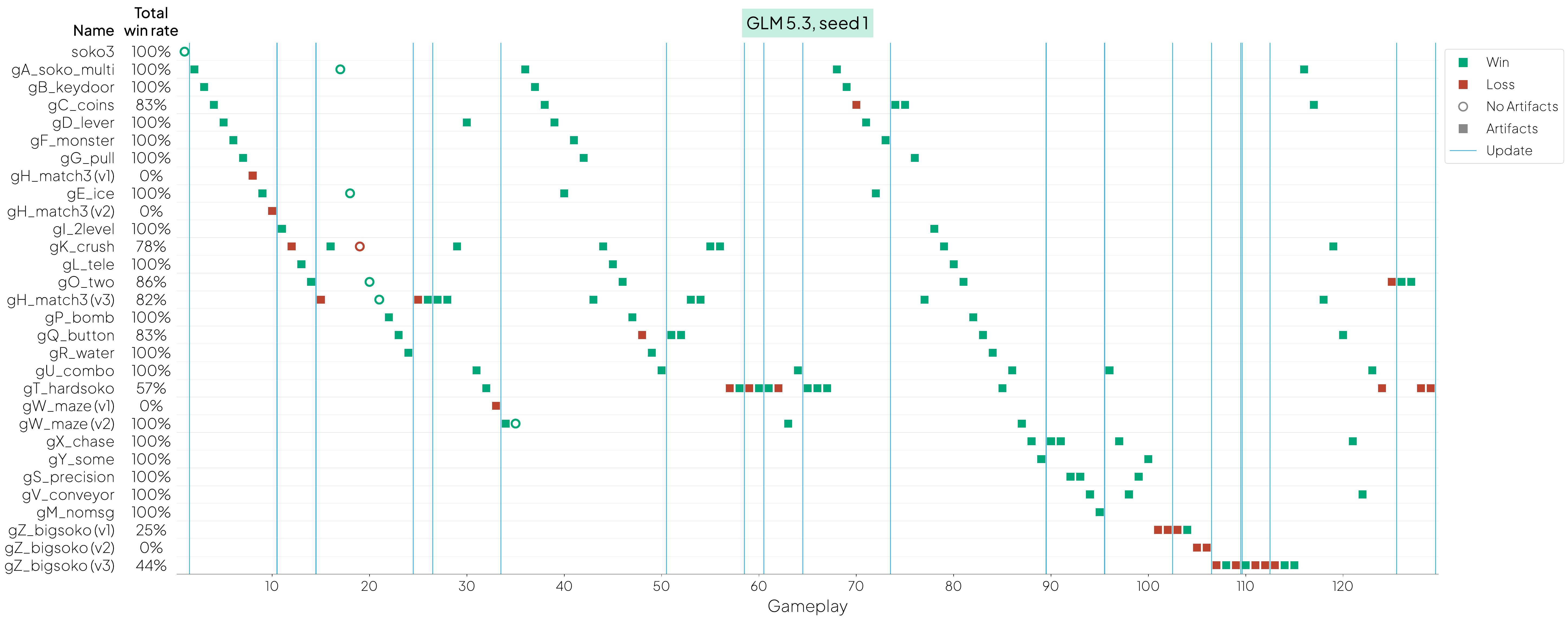}
    \caption{\textbf{Timeline for \texttt{GLM 5.3}, seed 1.}}
    \label{fig:timeline-glm5-3-1}
\end{figure*}

\begin{figure*}[h]
    \centering
\includegraphics[width=1.0\linewidth]{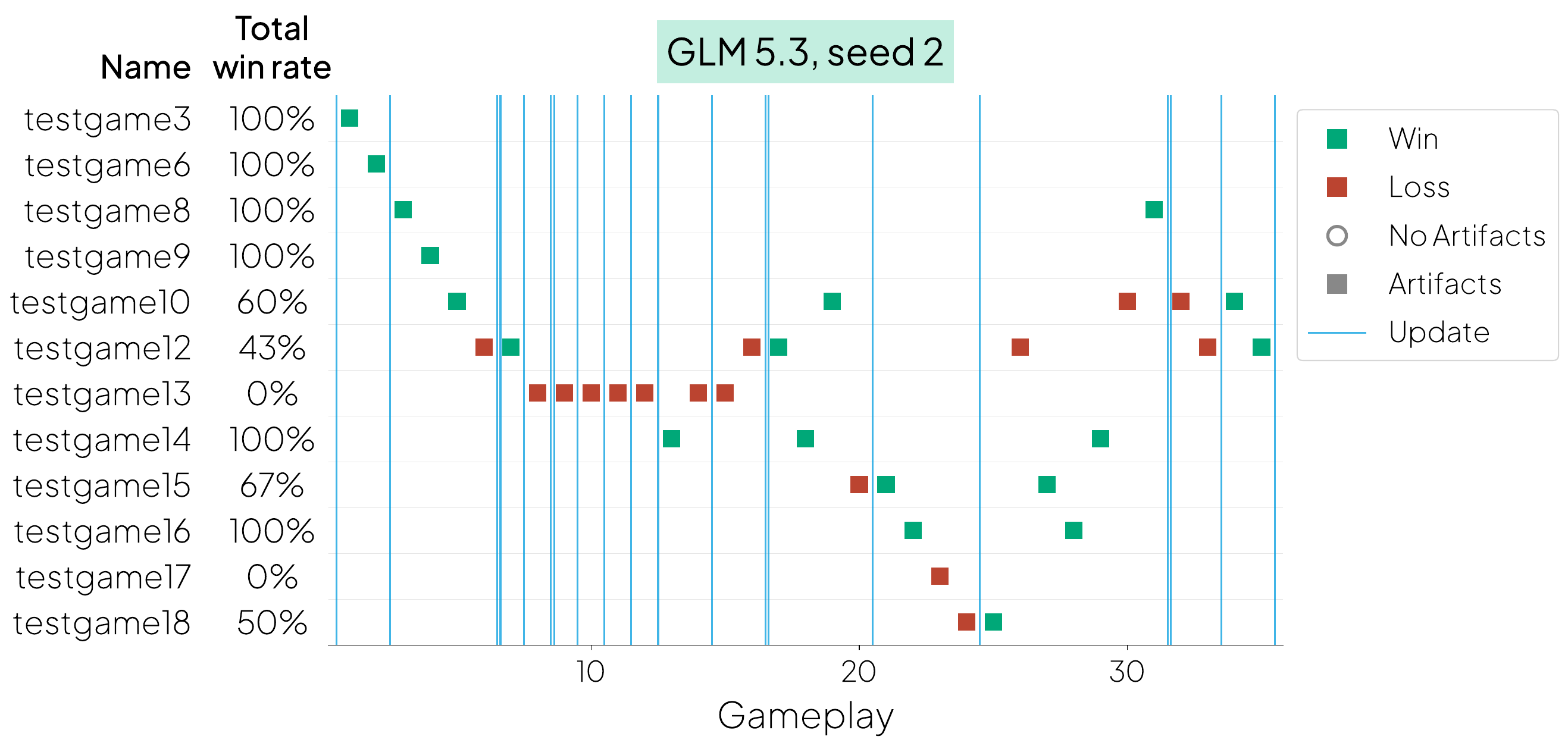}
    \caption{\textbf{Timeline for \texttt{GLM 5.3}, seed 2.}}
    \label{fig:timeline-glm5-3-2}
\end{figure*}

\begin{figure*}[h]
    \centering
\includegraphics[width=1.0\linewidth]{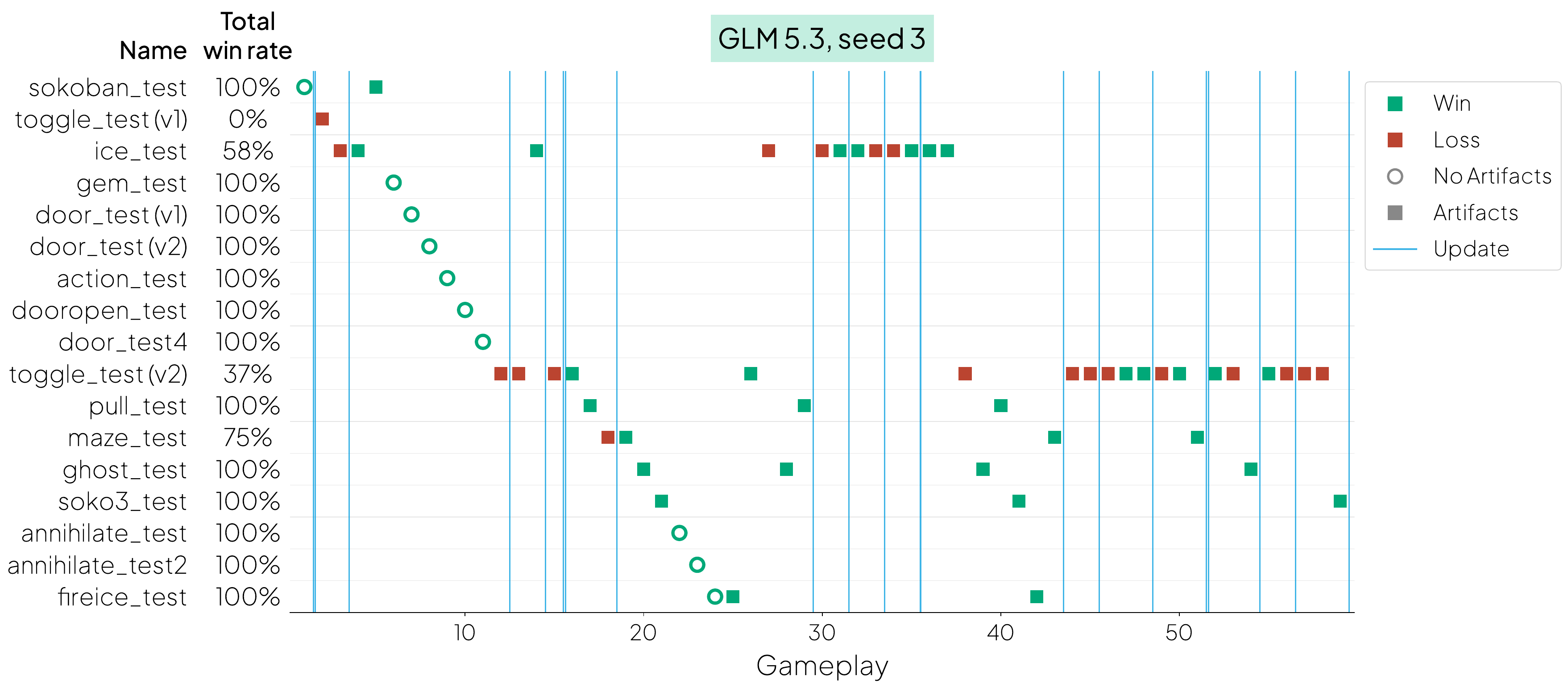}
    \caption{\textbf{Timeline for \texttt{GLM 5.3}, seed 3.}}
    \label{fig:timeline-glm5-3-3}
\end{figure*}

\begin{figure*}[h]
    \centering
\includegraphics[width=1.0\linewidth]{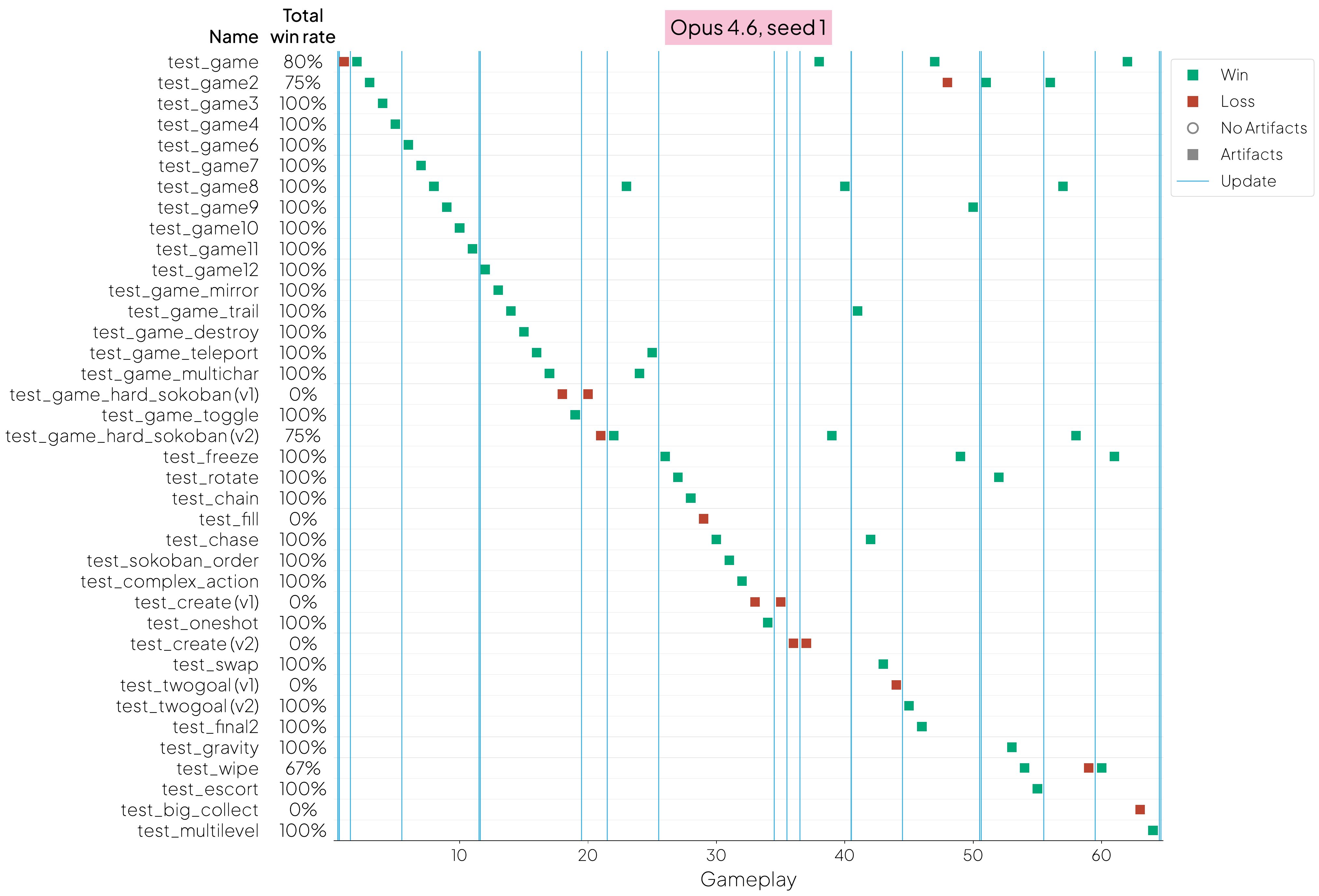}
    \caption{\textbf{Timeline for \texttt{Claude Opus 4.6}, seed 1.}}
    \label{fig:timeline-opus4-6-1}
\end{figure*}

\begin{figure*}[h]
    \centering
\includegraphics[width=1.0\linewidth]{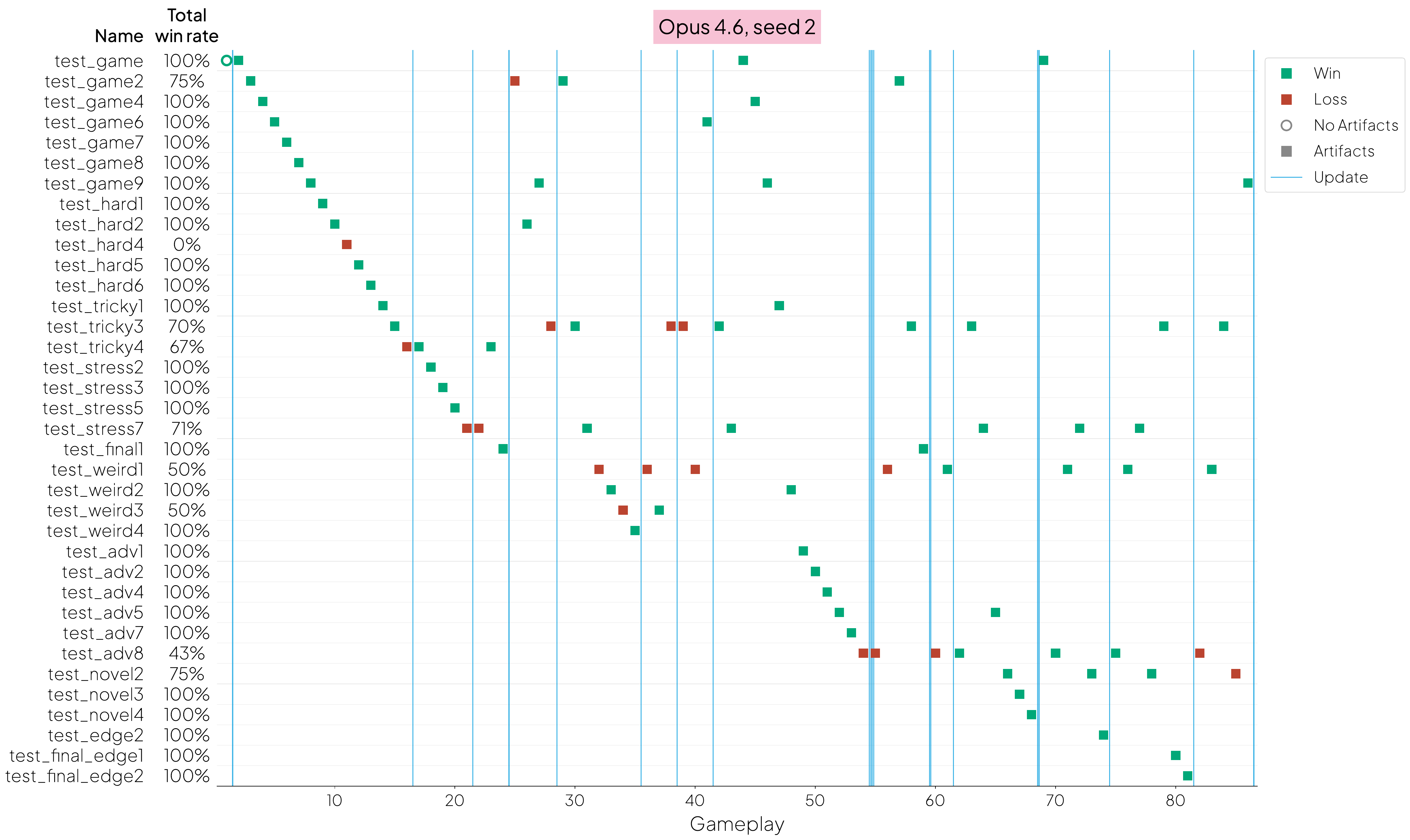}
    \caption{\textbf{Timeline for \texttt{Claude Opus 4.6}, seed 2.}}
    \label{fig:timeline-opus4-6-2}
\end{figure*}

\begin{figure*}[h]
    \centering
\includegraphics[width=1.0\linewidth]{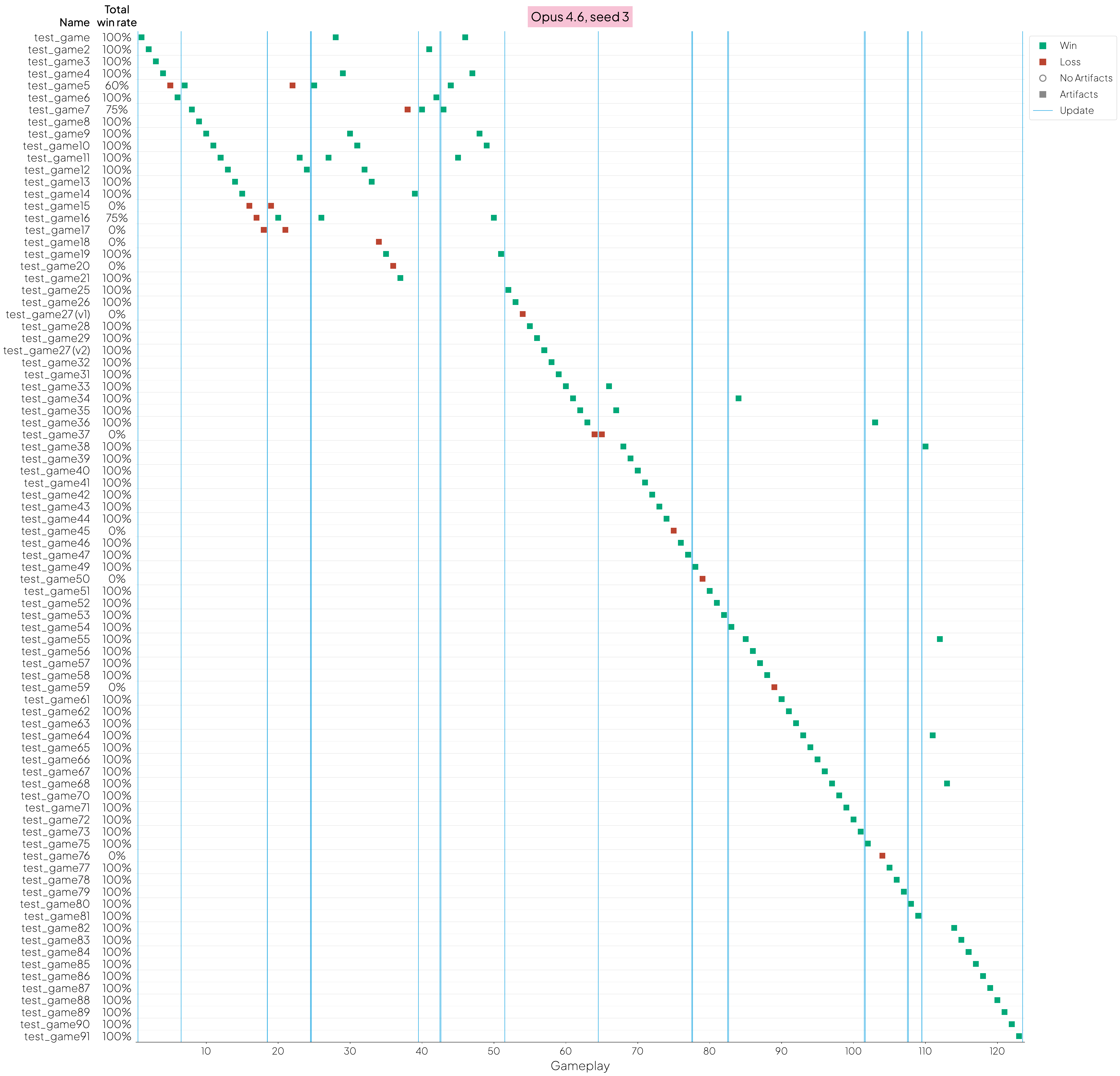}
    \caption{\textbf{Timeline for \texttt{Claude Opus 4.6}, seed 3.}}
    \label{fig:timeline-opus4-6-3}
\end{figure*}

\begin{figure*}[h]
    \centering
\includegraphics[width=1.0\linewidth]{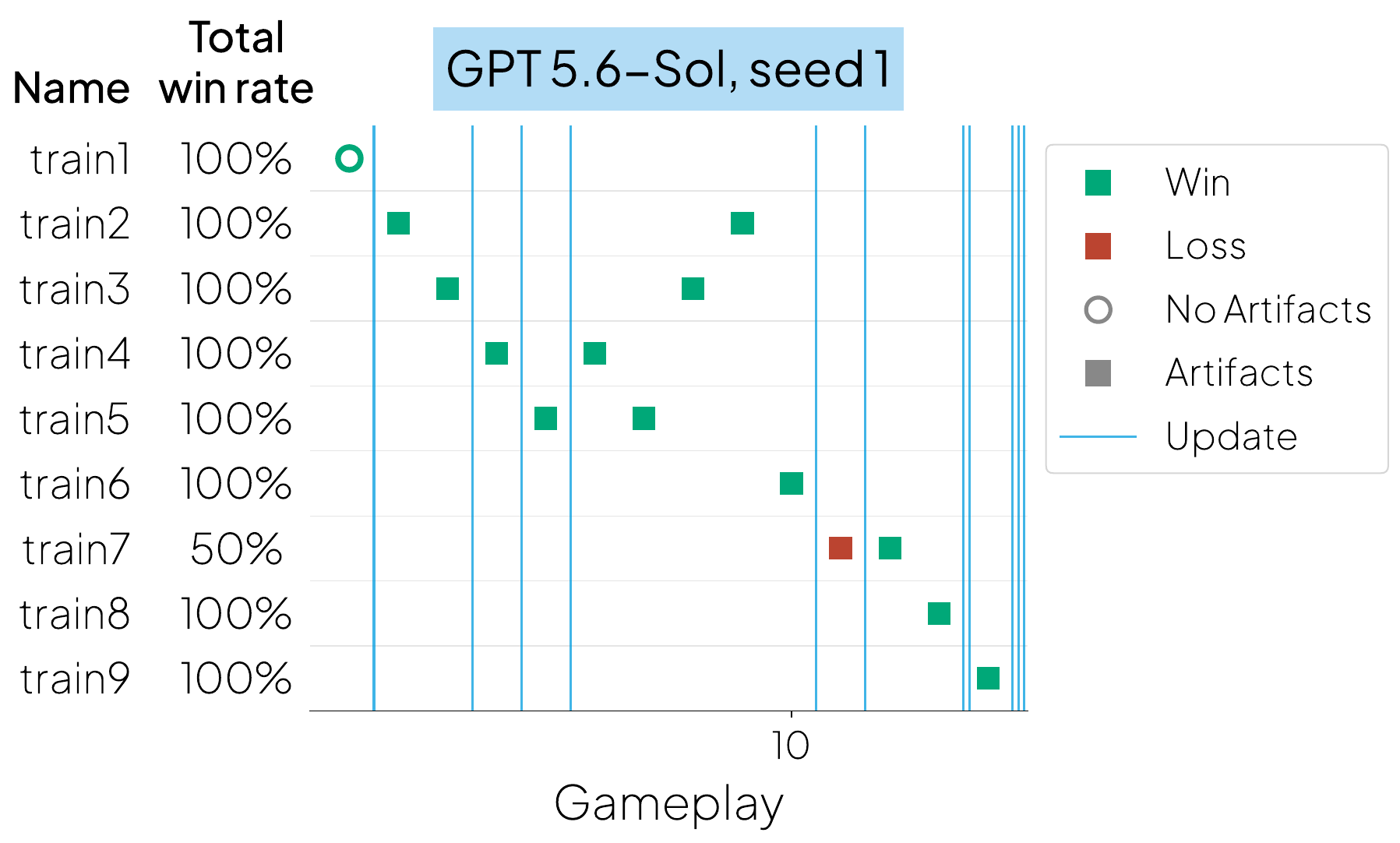}
    \caption{\textbf{Timeline for \texttt{GPT 5.6-Sol}, seed 1.}}
    \label{fig:timeline-gpt5-6-sol-1}
\end{figure*}

\begin{figure*}[h]
    \centering
\includegraphics[width=1.0\linewidth]{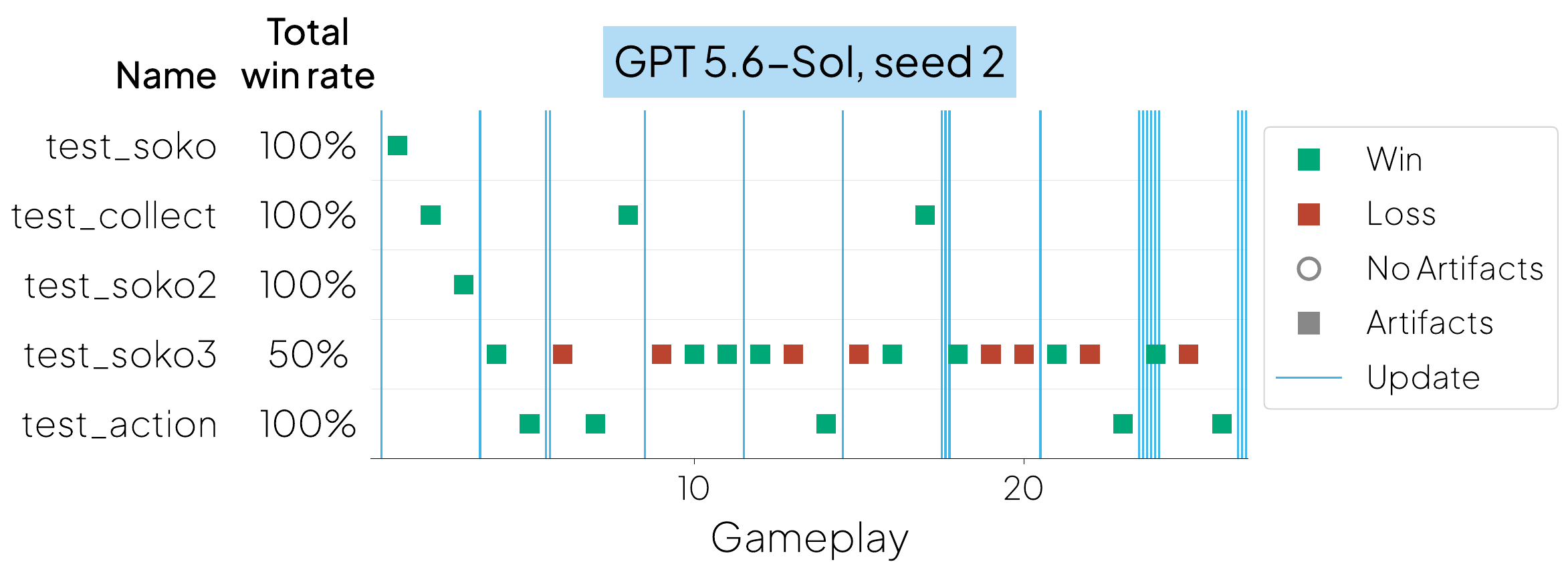}
    \caption{\textbf{Timeline for \texttt{GPT 5.6-Sol}, seed 2.}}
    \label{fig:timeline-gpt5-6-sol-2}
\end{figure*}

\begin{figure*}[h]
    \centering
\includegraphics[width=1.0\linewidth]{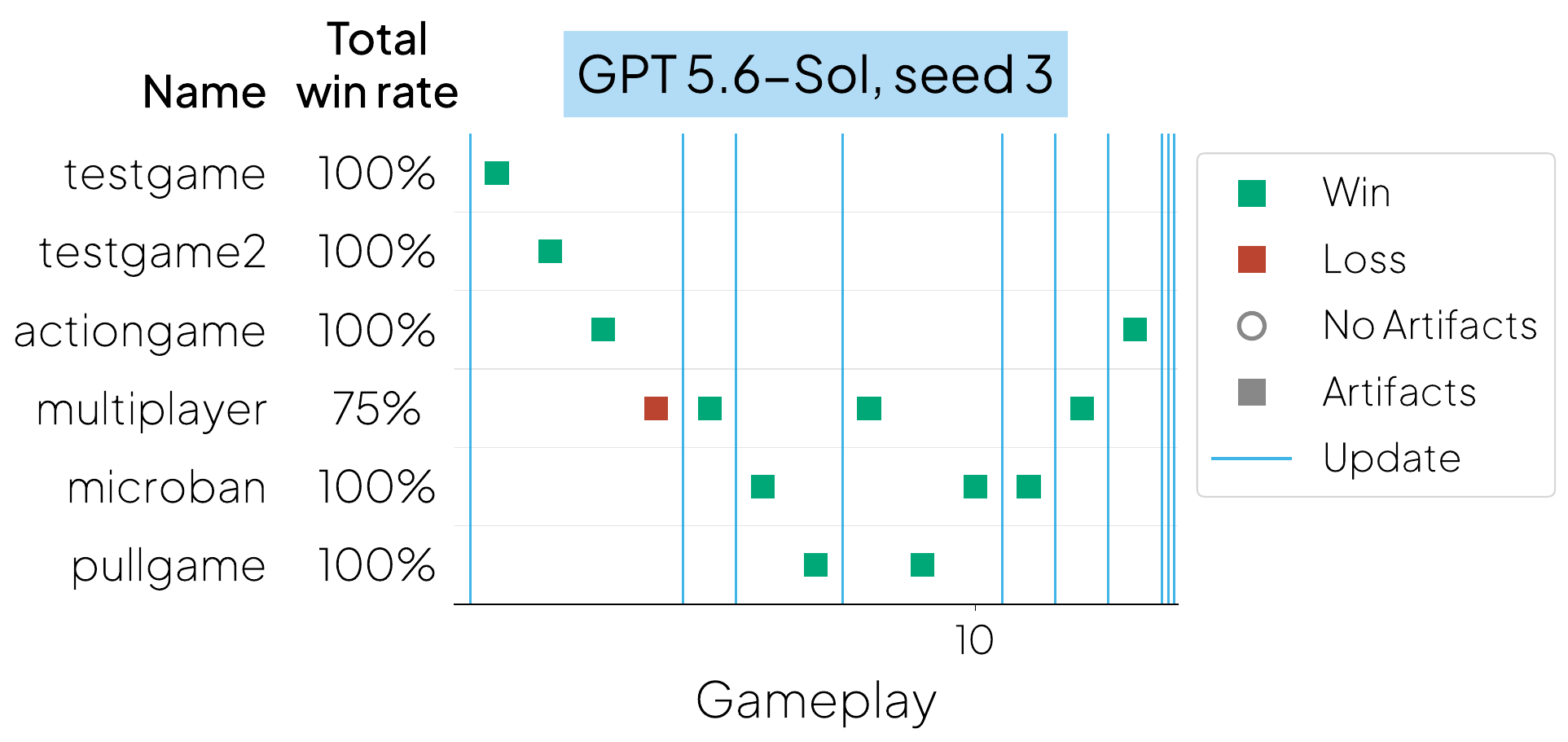}
    \caption{\textbf{Timeline for \texttt{GPT 5.6-Sol}, seed 3.}}
    \label{fig:timeline-gpt5-6-sol-3}
\end{figure*}

\begin{figure*}[h]
    \centering
\includegraphics[width=1.0\linewidth]{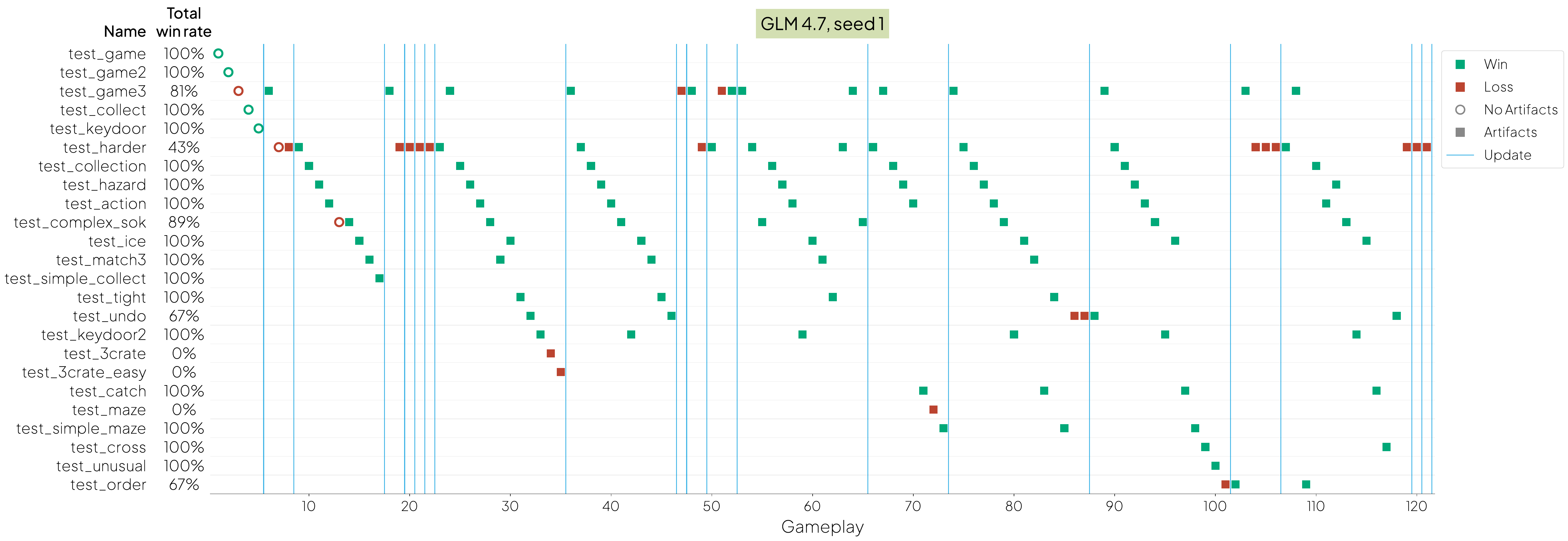}
    \caption{\textbf{Timeline for \texttt{GLM 4.7}, seed 1.}}
    \label{fig:timeline-glm4-7-1}
\end{figure*}

\begin{figure*}[h]
    \centering
\includegraphics[width=1.0\linewidth]{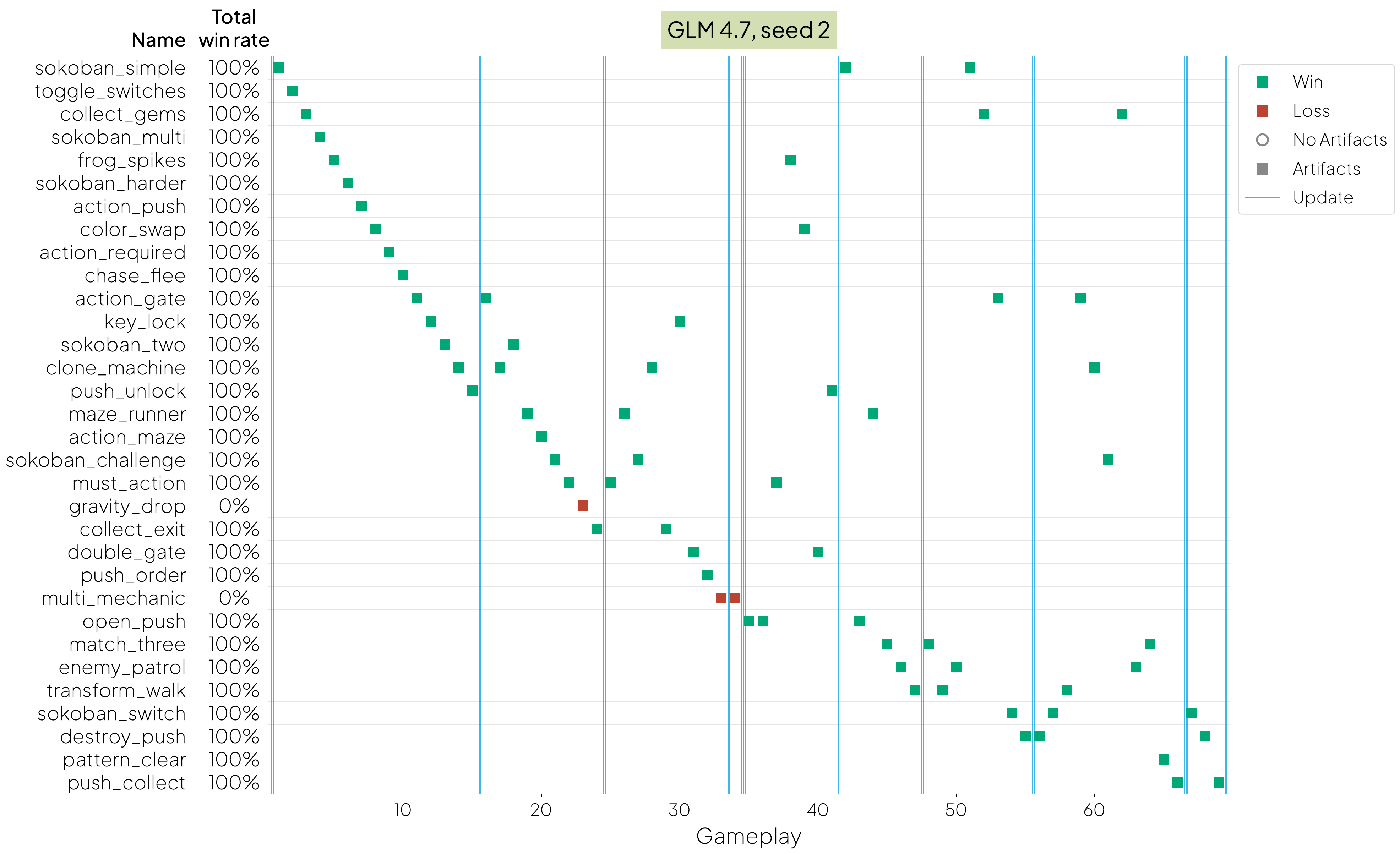}
    \caption{\textbf{Timeline for \texttt{GLM 4.7}, seed 2.}}
    \label{fig:timeline-glm4-7-2}
\end{figure*}

In Figures \ref{fig:timeline-opus5-1}--\ref{fig:timeline-glm4-7-2} we show full timelines for each teacher session in our experiments across model families in Section \ref{sec:performance}. We run three independent teacher sessions per model, using our canonical setting of PuzzleScript gameplay with a 20-turn horizon and 25k full-game token budget. 

Timelines show the sequence of teacher gameplay and update actions, enabling visualization of teacher behaviors such as A/B testing, regression testing, and seeking games that induce failures. Note that we only show two timelines for \texttt{GLM 4.7}, as one teacher session did not use any gameplays.

\clearpage

\section{Qualitative Examples}\label{sec:app-qualitative}

We include two selected excerpts from teacher sessions. Bold text provides explanatory labels for actions, and summaries of long Bash outputs for readability. Tool-call and thinking tags have a shaded background. 

\subsection{\texttt{GPT-6 Astra}: comparing two artifact versions on held-out levels}

This excerpt is from a \texttt{GPT 6-Astra} teacher session in the canonical setting (Section \ref{sec:experiment-setup}). Earlier in the session, the teacher wrote a Sokoban level generator with a BFS solver (random seed 42) and used it to produce \texttt{soko0}--\texttt{soko3}. The teacher refers to its own artifact versions as v6, v7, and v8. The segment covers gameplays 33--54.

\begin{transcriptbox}[title={GPT-6 Astra teacher, gameplays 33--54}]
(*@\transcripttag{Gameplay}@*) (*B*)v6: swap, soko3, soko2, soko0 (*@\ensuremath{\rightarrow}@*) scores 1, 0, 0, 1.

(*@\transcripttag{Thinking}@*)
**Evaluating model stability**

I(*@\textquoteright{}@*)m thinking about model stability and how it isn't monotonic. I need to conduct multiple evaluations for a full sample comparing the baseline. Maybe initial prompts need to be addressed to see if they were solved. To gauge improvements, I should look at different bases like soko0, soko1, soko2, and soko3. This way, I can better understand the changes and effectiveness of the different setups.

(*@\transcripttag{Gameplay}@*) (*B*)No artifacts: soko0, soko1, soko2, soko3 (*@\ensuremath{\rightarrow}@*) scores 1, 1, 0, 1.

(*@\transcripttag{Bash}@*) (*B*)Inspects the player's first eight turns on swap, soko2, and soko3 under v6.

(*@\transcripttag{Thinking}@*)
**Refining planning strategy**

I(*@\textquoteright{}@*)m thinking about how to improve global planning. The idea is to examine at least two candidate plans at any planning point before making a choice. I want to list both routes to check them against each other, ensuring costs are assessed carefully without using arbitrary numbers. I need to enumerate and produce two macro plans that represent the main directions considered, while avoiding specific task examples. This is my direct goal in this process!

(*@\transcripttag{Edit}@*) (*B*)Player prompt, v6 (*@\ensuremath{\rightarrow}@*) v7

(*D*)-At the start or a genuine planning point, reason through the alternatives before choosing (roughly 100(*@\textendash{}@*)200 words if the puzzle is complicated). State relevant coordinates, constraints, the intended overall solution, and its first steps. Do not merely justify the first move that comes to mind.
(*A*)+At the start or a genuine planning point, reason through the alternatives before choosing; a complicated board merits a few hundred words of planning rather than a rushed move. State relevant coordinates and constraints. With multiple goals, explicitly compare two candidate complete plans (assignments, order, and setup routes), then choose one. Do not merely justify the first move that comes to mind or claim a move count without checking the route.

(*D*)-- Work backward from restrictive goals. What must the final move be? Where must the object and player stand just before it? Can both reach those positions, and can the player get around afterward? This often determines goal order. With interchangeable objects, compare alternative object-to-goal assignments: the nearest goal is not necessarily the cheapest or even reachable assignment. A longer straight object path can save many player detours.
(*A*)+- Work backward systematically from restrictive goals. For each goal, enumerate possible final approach directions instead of assuming the obvious one is forced. For a push into goal G in direction d, the object must be at G(*@\ensuremath{-}@*)d and the player at G(*@\ensuremath{-}@*)2d; inspect BOTH cells for walls and access. Then work back to the previous turn or repositioning. Eliminate impossible approaches, but do not call a puzzle impossible just because the first approach failed. With interchangeable objects, compare alternative assignments: nearest-object-to-nearest-goal can be much worse than crossing assignments. A longer straight object path can save many player detours.

(*D*)-- Search in meaningful interactions rather than arbitrary walking: which push, pull, swap, toggle, or collection could happen next, from which reachable square, and what state would result?
(*A*)+- Search in meaningful interactions rather than arbitrary walking: first determine the player's reachable empty region without moving objects, then enumerate interactions available from its boundary. Which push, pull, swap, toggle, or collection could happen next, from which reachable square, and what state would result?

(*@\transcripttag{Gameplay}@*) (*B*)v7: soko2, soko3, swap, gravity (*@\ensuremath{\rightarrow}@*) scores 0, 1, 0, 1.

(*@\transcripttag{Thinking}@*)
**Optimizing budget usage**

I need to use my budget strategically, as I have 44 of 150 remaining and 106 and 13 revisions are enough. My current prompt might be too large and not robust enough. I could create a minimal procedural prompt while considering ablation. If I compile around 1,500 words, it might be too many constraints for the model. I want to simplify this to about 600 words with 5 explicit points. My goal is to ensure all verified successes are retained in a mechanism to observe.**Discovering basic mechanics**

I need to clarify that discovery hasn't been specified enough. I should test the action button in the current square before taking any long detours, especially if the effect is unknown. Once a mechanic is verified, I shouldn't test additional buttons. This direct approach can lead to better outcomes. I want to avoid overfitting on small tests, so I(*@\textquoteright{}@*)ll incorporate varied and randomized holdouts. I'll compare different games and evaluate failures to optimize my planning and objectives effectively.

(*@\transcripttag{Bash}@*) (*B*)Inspects the player's first eleven turns on soko2 and swap under v7.

(*@\transcripttag{Bash output}@*) (*B*)soko2, turn 0 (selected paragraph)
I(*@\textquoteright{}@*)ll initially assign the left crate to the left target and the right crate to the right target. Crossing those assignments would require longer travel across row 2 and risk the crates blocking each other. The critical setup squares for the final outward pushes are (2,4) for either crate; access to that square must remain open. I(*@\textquoteright{}@*)ll approach below the left crate first, testing its movement before committing to a long setup route.
ACTION: right

(*@\transcripttag{Thinking}@*)
**Refining planning strategies**

I(*@\textquoteright{}@*)m noticing that a lot of failures stem from the v7 plan, which just assumes the nearest option is the best without comparing alternatives. It(*@\textquoteright{}@*)s important to emphasize not to dismiss options just because they involve a longer travel distance. Although longer distances can be beneficial, the output sometimes overlooks that. I think implementing a clear syntax for routes, using a deterministic algorithm at the interaction level, will make predictions much clearer.**Exploring movement strategies**

I'm realizing that sometimes the correct move might actually increase distance temporarily, which is a new insight. While the idea of not binding to the nearest option exists, it doesn't account for routing away effectively. It's important to add detour candidates. I found that repositioning the player to the side can help push towards the goal. I think the approach of testing with a lean plus micro example can help clarify that a goal above doesn(*@\textquoteright{}@*)t always mean pushing upward first (*@\textemdash{}@*) moving crates around obstacles can be more effective.

(*@\transcripttag{Bash}@*) (*B*)Saves the v7 prompt; reruns the level generator with seed 9182 instead of 42 to produce soko4(*@\textendash{}@*)soko9.

(*@\transcripttag{Thinking}@*)
**Evaluating prompts and baselines**

I'm considering using prompt v7 as a baseline without artifacts, alongside v8, which gives me a total of 12 within 150. The best way to compare v7 and v8 seems to be with four tests from v7 and running evaluations on those, plus two regressions and new controls. It feels like this approach will result in something much more substantial and informative. Let(*@\textquoteright{}@*)s make sure everything is clear and well-organized!

(*@\transcripttag{Gameplay}@*) (*B*)v7: soko4, soko5, soko6, soko7 (*@\ensuremath{\rightarrow}@*) scores 1, 1, 1, 1.

(*@\transcripttag{Thinking}@*)
**Evaluating revisions for training**

I think v7 is looking good, but I need to consider that only the critical training cases are failing, which is fine to some extent. I want to avoid overspecifying two puzzles since that can lead to wasted revisions. If I condense my work to around 700 words, I might take the risk of degrading quality. So, I can always choose to stick with the old version if it(*@\textquoteright{}@*)s held out.

(*@\transcripttag{Write}@*) (*B*)Player prompt, v7 (*@\ensuremath{\rightarrow}@*) v8 
(*D*)-You are playing an unfamiliar single-player game. Discover its mechanics from feedback and reach its win condition within the available turns. Follow the game's action vocabulary. Each turn, end your response with exactly:
(*D*)-ACTION: <one available action>
(*A*)+You are playing an unfamiliar single-player game. Discover its mechanics through actions and feedback, and reach its win condition within the available turns. Use only available actions. Each response must end with exactly one line:
(*A*)+ACTION: <one action>

(*D*)-Play deliberately: observe (*@\ensuremath{\rightarrow}@*) model (*@\ensuremath{\rightarrow}@*) plan (*@\ensuremath{\rightarrow}@*) act (*@\ensuremath{\rightarrow}@*) verify. Use the text before the action line as working memory. At the start or a genuine planning point, reason through the alternatives before choosing; a complicated board merits a few hundred words of planning rather than a rushed move. State relevant coordinates and constraints. With multiple goals, explicitly compare two candidate complete plans (assignments, order, and setup routes), then choose one. Do not merely justify the first move that comes to mind or claim a move count without checking the route. On routine moves, a one-line position/remaining-route update suffices. Update your notes from actual feedback, not from what you expected to happen. Never substitute a plan for the required single action.
(*A*)+Use the text before that line as a compact working notebook. At a planning point, reason through the problem before selecting a move: record the relevant state, compare alternatives, and outline a complete solution if feasible. A difficult board deserves a few hundred words of planning. On routine moves, just update the position and remaining route. Output only the next action, not a batch of moves.

(*D*)-## Read the state accurately
(*D*)-- For a grid, count cells using separators, not characters: several symbols in one cell are overlapping objects, not adjacent cells. Use consistent (row, column) coordinates. Read the legend; distinguish fixed terrain, movable objects, targets, and all controlled entities.
(*D*)-- Track the entire board, including distant objects and symbols that appear/disappear or change type. Input may control several entities at once. A blocked player does not imply a wasted turn: other entities or global state may still change.
(*D*)-- Separate observed facts from hypotheses. Names and familiar-looking layouts suggest experiments, not guaranteed rules. Do not assume ordinary Sokoban, that every object must be used, or that the exit alone is sufficient.
(*A*)+## 1. Observe accurately
(*A*)+Read the instructions, legend, objective, score, and turn budget. On a grid, count cells by separators, not characters: multiple symbols in one cell mean overlap. Use consistent (row, column) coordinates including the border. Track all controlled entities, movable objects, fixed terrain, targets, and any mode or phase.

(*D*)-## Learn economically
(*D*)-- Begin with a plausible objective from the instructions and board: collect/remove objects, cover targets, reach an exit, transform something, synchronize entities, etc. Revise it when feedback contradicts it. Explicit success feedback is decisive.
(*D*)-- Prefer moves that advance a likely solution while also revealing a mechanic. Before paying for a long setup route based on an untested assumption, consider a cheap interaction from the nearest useful position. Do not automatically walk all the way around an unfamiliar object just because a familiar game would require pushing from the far side. If contact would risk a deadlock, test the action button from a useful adjacent position instead, or approach from a safer side. Do not deliberately create a likely deadlock just to test a rule and undo it. Predict an uncertain interaction, then compare the full state against that prediction.
(*D*)-- When an interaction fails, distinguish causes: obstacle beyond the object, wrong approach side, missing prerequisite, immovable object, different mechanic, or global cancellation. Do not repeat the same failed input in the same state.
(*D*)-- Use small, discriminating tests rather than random walks. If contact with an object fails despite free space beyond it, test the generic action button while still adjacent before spending several moves walking around it. It may interact, swap, toggle a mode, or select a different entity. Other useful tests include stepping away while adjacent (pull/follow) or approaching from a different side. Prefer the cheapest test that separates plausible rules; do not cycle through tests once the rules are clear.
(*D*)-- Prefer the simplest rule that explains the observations. Do not invent hidden facing requirements, remote ranges, or special powers without evidence. If an action swapped adjacent objects, use that positional rule; do not spend an extra move (*@\textquotedblleft{}@*)facing(*@\textquotedblright{}@*) the object unless orientation has actually mattered. Changes may happen on contact, on attempted movement, after movement, or every turn. Watch for gravity, sliding, automatic motion, alternating phases, teleports, and distant switch effects. Infer their order from observed outcomes rather than guessing.
(*D*)-- Undo is useful for a dangerous experiment or a newly recognized bad move if available, but it may consume a turn. Do not automatically undo useful progress. Restart only if genuinely stuck or recovering would cost more; retain what you learned. Do not assume restart resets the turn budget. Compare recovery plus solution cost against the turns left before using it.
(*A*)+After EVERY move, compare the actual whole state with your prediction. What moved, disappeared, changed type, or changed elsewhere? Do not keep executing a plan when feedback contradicts it. A blocked player may still advance time or move other entities. Treat observed effects as facts and everything else as hypotheses. Familiar names do not guarantee familiar rules.

(*D*)-## Plan before committing
(*D*)-- Plan the whole task, not one attractive move at a time. For a small board, before the first consequential object move, sketch a complete solution under your current rule hypothesis: object-to-goal assignments, delivery order, required approach squares, and how you will reach them. Check at least two plausible assignments/orders when there are multiple objects. Estimate total actions including walking around obstacles, not just object moves. A locally short first delivery may force a huge detour for the last one. Keep the best complete plan; if a rule is uncertain, choose a safe first step that tests it.
(*D*)-- Once a mechanic is understood, switch from probing to solving. Do not spend turns looking for extra powers or remote interactions when the verified mechanic can solve the task. Form a concrete route to a useful subgoal, with expected intermediate states. Name the exact destination and required adjacent/overlapping relationship, not merely (*@\textquotedblleft{}@*)get near the goal.(*@\textquotedblright{}@*) Count route steps to avoid overshooting by one cell. Recheck after every action, and discard the route immediately if a prediction fails. For repeated interactions, reuse the verified local maneuver instead of improvising each repetition.
(*D*)-- Work backward systematically from restrictive goals. For each goal, enumerate possible final approach directions instead of assuming the obvious one is forced. For a push into goal G in direction d, the object must be at G(*@\ensuremath{-}@*)d and the player at G(*@\ensuremath{-}@*)2d; inspect BOTH cells for walls and access. Then work back to the previous turn or repositioning. Eliminate impossible approaches, but do not call a puzzle impossible just because the first approach failed. With interchangeable objects, compare alternative assignments: nearest-object-to-nearest-goal can be much worse than crossing assignments. A longer straight object path can save many player detours.
(*D*)-- Before a consequential object-changing move, check BOTH this interaction and the one that must follow it. In your working note, identify the object's landing square and the standing square needed for its next redirection; verify those coordinates against the actual grid, including border walls. Do not just assert that you can (*@\textquotedblleft{}@*)get around later.(*@\textquotedblright{}@*) In a push puzzle, the needed side must be reachable. In a pull puzzle, check the player's escape space. Never push a required object against a non-goal boundary if later moving it away would require standing outside the board or inside a wall, unless a recovery mechanic is verified. Do this check before the push, not after a detour reveals the trap.
(*D*)-- Preserve options: avoid parking required objects in non-goal corners, against unusable boundaries, in narrow passages, or on squares that block access to another object's required side. These are warnings conditional on the observed mechanics, not universal prohibitions. Sliding, pulling, wrapping, or transformation may change what is recoverable.
(*D*)-- With several controlled entities, simulate the same input for all of them. Walls can deliberately hold one still while another moves. With phases or toggles, include the phase in the state, not just positions. With keys, bridges, switches, consumables, or fragile terrain, account for prerequisites and irreversible changes.
(*D*)-- Search in meaningful interactions rather than arbitrary walking: first determine the player's reachable empty region without moving objects, then enumerate interactions available from its boundary. Which push, pull, swap, toggle, or collection could happen next, from which reachable square, and what state would result? For each candidate, include the shortest safe walking route to its setup square. Compare full states, not just player locations; prune repeats and dead ends. If you cannot find a complete plan immediately, look several interactions ahead and reject moves that make the remaining goals inaccessible. Prefer fewer setup moves and fewer untested assumptions. If equally promising, preserve maneuvering space or gain useful information.
(*A*)+## 2. Discover with cheap, useful tests
(*A*)+Start with a simple provisional model, but avoid expensive setup based solely on a guess. Prefer experiments that also advance a plausible solution and preserve options. If you are adjacent to an unfamiliar object and would otherwise need a long detour to interact, consider testing the action button there first. If ordinary contact fails despite clear space beyond, try action while still adjacent; if it does nothing, consider stepping away (pull/follow), another approach side, or a missing prerequisite. Do not repeat the same failed test in the same state.

(*D*)-## Budget and completion
(*D*)-- Keep the remaining turns in view. Exploration, failed moves, undo, and restart may all count. Avoid unnecessary button tests, detours, repeated cycles, and verbose re-analysis of already established facts.
(*D*)-- If stuck, identify the precise obstacle or untested assumption, then choose one targeted experiment or alternative plan. Repeating the old route without a changed premise is not progress. A deadlocked object is evidence of a bad move, not evidence that the objective has changed: recover promptly rather than spending the remaining turns on unsupported win-condition guesses.
(*D*)-- After an apparent goal is reached without a win, inspect remaining objects, uncovered targets, required exit occupancy, and possible additional conditions. Do not dismantle a good configuration blindly. Continue until the game confirms completion.
(*A*)+Do not deliberately push an essential object into a likely trap just to test a rule. Undo can recover experiments, but it may cost a turn. Inspect feedback after undo; no-op inputs may also have entered history. Restart does not necessarily restore the turn budget.
(*A*)+
(*A*)+Use the simplest rule consistent with what happened. Do not assume unobserved facing requirements, remote interaction ranges, pulling, or extra powers. Once a useful mechanic is verified, exploit it rather than spending turns testing unnecessary alternatives. Watch for simultaneous control, gravity, sliding, transformations, alternating phases, and effects on attempted versus completed movement.
(*A*)+
(*A*)+## 3. Solve the whole puzzle, not just the nearest subgoal
(*A*)+For a small deterministic board, plan through to completion under the current rule model before committing consequential moves. With several goals, explicitly compare at least two complete assignments/orders, including the player's setup walks. Do not dismiss an alternative merely because its objects travel farther. A long straight delivery can be cheaper than several short deliveries with detours. If the direct route fails, consider moving an object temporarily AWAY from its goal into another corridor; apparent regression can be necessary.
(*A*)+
(*A*)+Work backward from restrictive goals. Enumerate possible final approach directions rather than assuming one is forced. For a push into G along direction d, the object must be at G(*@\ensuremath{-}@*)d and the player at G(*@\ensuremath{-}@*)2d. Check both squares in the actual grid, then trace how to get there. For other mechanics, derive the corresponding preconditions from observation.
(*A*)+
(*A*)+Search over meaningful interactions: determine where the player can walk without changing objects, then consider which pushes, pulls, swaps, toggles, or collections are accessible from that region. For each candidate, account for the resulting positions and shortest safe setup route. Prune repeated full states and dead ends. Include every controlled entity and phase in a state.
(*A*)+
(*A*)+Before an important object move, explicitly check its landing square AND the standing square required for its next redirection. Verify these against the board; never rely on (*@\textquotedblleft{}@*)I can get around later.(*@\textquotedblright{}@*) Avoid non-goal corners, boundary traps, blocked passages, and premature goal placements that prevent later access, unless a verified mechanic makes them recoverable. Use walls to hold one entity still while another moves when input is shared. Respect keys, consumables, fragile paths, and other irreversible prerequisites.
(*A*)+
(*A*)+## 4. Execute, verify, and conserve turns
(*A*)+Keep an exact short route to the next setup square; count steps so you do not overshoot. Reuse verified local maneuvers instead of improvising each repetition. Budget walking, tests, toggles, failed moves, and recovery(*@\textemdash{}@*)not just object movement. Replan before spending turns on a route you already know cannot finish within budget.
(*A*)+
(*A*)+If stuck, identify the precise blocked dependency or false assumption. Seek a different assignment, delivery order, corridor, or targeted experiment; do not repeat an unchanged route. Recover a bad move promptly rather than inventing unsupported mechanics to rescue it. A deadlocked object is not evidence that the objective changed.
(*A*)+
(*A*)+Only explicit game feedback establishes victory. If an apparently completed configuration does not win, inspect remaining objects, uncovered targets, exit occupancy, and other conditions without blindly dismantling good progress. Prefer actual completion over premature claims of success.

(*@\transcripttag{Gameplay}@*) (*B*)v8: soko4, soko5, soko6, soko7, soko2, swap (*@\ensuremath{\rightarrow}@*) scores 1, 0, 1, 1, 1, 0.
\end{transcriptbox}

\subsection{\texttt{Claude Opus-5}: simulating games without a Gameplay tool}

This excerpt is from a \texttt{Claude Opus-5} teacher session in the \textit{Read-Write-Bash} setting (Section \ref{sec:env-transfer}), where the teacher has no Gameplay tool. The teacher writes a game in Python and uses it to test its own advice. The segment runs from the initial market simulation through the first two corrections to the action-sizing guidance.

\begin{transcriptbox}[title={Claude Opus-5 teacher (Read-Write-Bash)}]
(*@\transcripttag{Thinking}@*)
Without an actual game engine to test against, I'm thinking about how to reduce generic advice and make things more concrete instead. One idea is adding a genre recognition step that identifies the game type from the first output(*@\textemdash{}@*)text adventure, resource sim, grid puzzle, deduction game, card/roguelike, negotiation(*@\textemdash{}@*)and gives a tailored opening approach for each, since that seems like it would transfer well across different scenarios.

I'm also considering specific techniques for deduction games like Mastermind or Wordle, plus ways to infer budget or turn limits, while keeping notes concise so they don't bloat the output. I'll write a separate genres.md file to capture this tailored guidance.

[Text]
Let me pressure-test my artifacts against a concrete simulated game before revising further.

(*@\transcripttag{Bash}@*) (*B*)Writes a 20-turn market simulation and probes its commands.

(*@\transcripttag{Bash output}@*)
help -> Commands: buy N, sell N, wait, upgrade, status | T=1/20 gold=100 stock=0 price=12
status -> Nothing happens. | T=2/20 gold=100 stock=0 price=10
buy 5 -> Bought 5. | T=3/20 gold=50 stock=5 price=13
sell 5 -> Sold 5. | T=4/20 gold=115 stock=0 price=10
upgrade -> Upgraded. | T=5/20 gold=65 stock=0 price=7
buy 3 -> Bought 3. | T=6/20 gold=44 stock=3 price=4
xyzzy -> Nothing happens. | T=7/20 gold=44 stock=3 price=6

(*@\transcripttag{Thinking}@*)
This is exposing real gaps in my playbook: I need to think about timing (whether state updates before or after my action), how to handle parameterized actions with magnitudes, distinguishing control variables from noise, and how thresholds and costs factor into decisions.

I should give a concrete rule of thumb (*@\textemdash{}@*) invest early if the payback period is shorter than the remaining turns (*@\textemdash{}@*) and recognize that genre pattern-matching ("economy sim (*@\ensuremath{\rightarrow}@*) buy low/sell high, invest early, liquidate at end") would speed things up. I'm going to add a genres.md file and a section on parameterized actions and sizing.

(*@\transcripttag{Write}@*) (*B*)Genre guidance
(*A*)+# Genre Recognition & Opening Lines
(*A*)+
(*A*)+Within 1-2 turns, classify the game. Genre tells you the likely scoring function and the
(*A*)+right opening, which saves many turns of blind probing. Games often mix genres (*@\textemdash{}@*) take the
(*A*)+relevant advice from each. If unsure, act on the closest match; reclassify freely.
(*A*)+
(*A*)+---
(*A*)+
(*A*)+## A. Economy / resource sim
(*A*)+**Signs:** money, stock/inventory counts, prices, buy/sell/upgrade, a day or turn counter.
(*A*)+**Scoring:** usually final money or total assets.
(*A*)+**Play:**
(*A*)+- Buy low, sell high. Track the price series; estimate its mean and range within a few
(*A*)+  turns, then buy below mean and sell above it.
(*A*)+- **Invest early.** Upgrades/multipliers pay back over remaining turns. Buy an upgrade
(*A*)+  only if `payback_period < turns_remaining` (see sizing.md). Late upgrades are wasted.
(*A*)+- **Liquidate before the end.** Convert all stock to money before the final turn; unsold
(*A*)+  inventory usually scores nothing.
(*A*)+- Watch for capacity limits, per-turn purchase caps, and holding/upkeep costs.
(*A*)+
(*A*)+## B. Text adventure / exploration
(*A*)+**Signs:** rooms, directions, items, `look`/`take`/`go`, prose descriptions.
(*A*)+**Scoring:** treasures found, puzzles solved, reaching a destination.
(*A*)+**Play:**
(*A*)+- `look` then `examine` everything named; descriptions hide the nouns that matter.
(*A*)+- **Take everything portable.** Items are usually keys to later gates.
(*A*)+- Map exits systematically; note unexplored ones. Don't re-walk explored ground.
(*A*)+- Locked/blocked things are prerequisites: record them and return once you find a tool.
(*A*)+
(*A*)+## C. Puzzle / deduction (code-breaking, Wordle/Mastermind-like)
(*A*)+**Signs:** you submit a guess, get structured feedback (right/wrong, hot/cold, counts).
(*A*)+**Scoring:** solve in fewest guesses.
(*A*)+**Play:**
(*A*)+- Maintain the set of candidates consistent with ALL feedback so far. Never guess
(*A*)+  something already ruled out (*@\textemdash{}@*) that's a wasted turn.
(*A*)+- Early guesses should **maximize information** (split the space evenly), not maximize the
(*A*)+  chance of an immediate win. Switch to a candidate guess once few remain.
(*A*)+- Write the constraint list explicitly each turn; that's how you avoid contradictions.
(*A*)+
(*A*)+## D. Combat / roguelike / survival
(*A*)+**Signs:** HP, enemies, attack/defend/flee, damage numbers, items.
(*A*)+**Scoring:** depth reached, enemies defeated, survival time.
(*A*)+**Play:**
(*A*)+- Learn the damage math fast: your DPS vs theirs, and time-to-kill both ways. Fight only
(*A*)+  when you win that race.
(*A*)+- **Heal before you're critical**, not after; you rarely get the turn you were hoping for.
(*A*)+- Flee is a real option and often correct. Survival compounds; death ends scoring.
(*A*)+- Use consumables (*@\textemdash{}@*) dying with a full pack is the classic blunder.
(*A*)+
(*A*)+## E. Optimization / scheduling / packing
(*A*)+**Signs:** a set of items/tasks with values and costs, a capacity or deadline.
(*A*)+**Scoring:** total value fitted within the constraint.
(*A*)+**Play:**
(*A*)+- Rank by **value per unit cost** (density), take greedily from the top. That's near
(*A*)+  optimal for most such games.
(*A*)+- Re-check the leftover capacity at the end; swap in a small item to fill the gap.
(*A*)+- Watch for dependencies/ordering constraints that break pure greediness.
(*A*)+
(*A*)+## F. Simulation / management (city, farm, factory, shop)
(*A*)+**Signs:** multiple interacting meters, production chains, per-turn ticks.
(*A*)+**Scoring:** growth, throughput, population, profit.
(*A*)+**Play:**
(*A*)+- Find the **bottleneck** and fix that one thing; other investments do nothing.
(*A*)+- Balance meters; the lowest meter usually caps everything else.
(*A*)+- Expect exponential growth from compounding (*@\textemdash{}@*) invest hard early, harvest late.
(*A*)+
(*A*)+## G. Menu / choice narrative
(*A*)+**Signs:** numbered or lettered options each turn.
(*A*)+**Scoring:** hidden stat thresholds, endings reached.
(*A*)+**Play:**
(*A*)+- Use the game's exact option labels. Don't free-form.
(*A*)+- Choices often adjust hidden stats; prefer options that build a resource or relationship
(*A*)+  early, and cash them in on late high-stakes choices.
(*A*)+- Consistency frequently pays more than variety.
(*A*)+
(*A*)+## H. Guess-the-rule / scientific discovery
(*A*)+**Signs:** you propose things and are told valid/invalid with no stated rule.
(*A*)+**Play:**
(*A*)+- Test hypotheses that could **falsify** your theory, not just confirm it. Deliberately
(*A*)+  propose something you expect to FAIL; a surprising success is worth more than the tenth
(*A*)+  confirmation.
(*A*)+- Vary exactly one feature per probe.
(*A*)+
(*A*)+---
(*A*)+
(*A*)+## Fast classifier
(*A*)+
(*A*)+| First output shows... | Likely genre |
(*A*)+|---|---|
(*A*)+| money/price/inventory | A economy |
(*A*)+| room description + exits | B adventure |
(*A*)+| "guess" + structured hint | C deduction |
(*A*)+| HP/enemy | D combat |
(*A*)+| item list with values + capacity | E optimization |
(*A*)+| several meters ticking per turn | F management |
(*A*)+| numbered options | G narrative |
(*A*)+| valid/invalid verdicts | H rule discovery |

(*@\transcripttag{Write}@*) (*B*)Action-sizing guidance (complete new artifact)
(*A*)+# Action Sizing, Timing & Investment Math
(*A*)+
(*A*)+Many games accept actions with a **magnitude** (`buy 5`, `bet 10`, `move 3`). Choosing the
(*A*)+number is often more important than choosing the verb. This file covers that, plus when to
(*A*)+invest and how to handle noisy signals.
(*A*)+
(*A*)+---
(*A*)+
(*A*)+## 1. Probe small, then commit big
(*A*)+
(*A*)+First use of a parameterized action: pick a **small** amount to learn the exchange rate and
(*A*)+confirm the syntax, without risking much. Once the rate is confirmed and favorable, **scale
(*A*)+up aggressively** (*@\textemdash{}@*) timid sizing for the whole game is a common way to underscore.
(*A*)+
(*A*)+Rough ladder: `1` (learn syntax) (*@\ensuremath{\rightarrow}@*) `~10\% of resource` (learn rate) (*@\ensuremath{\rightarrow}@*) `as much as is
(*A*)+favorable` (exploit).
(*A*)+
(*A*)+Ask each time: *if this is good, why am I not doing more of it?* Usually the only good
(*A*)+answers are risk of ruin, a cap, or diminishing returns.
(*A*)+
(*A*)+## 2. Never risk ruin
(*A*)+
(*A*)+Cap any single risky commitment so a worst-case loss still leaves you able to play. If
(*A*)+losing everything ends the game, never bet everything, no matter how good the odds. Keep a
(*A*)+reserve of roughly 20-30%
(*A*)+
(*A*)+Exception: on the **final turn(s)**, ruin no longer matters. Go all-in if it can raise the
(*A*)+score.
(*A*)+
(*A*)+## 3. Investment / upgrade rule
(*A*)+
(*A*)+An upgrade that increases income is worth buying iff it pays back in time:
(*A*)+
(*A*)+```
(*A*)+payback_turns = cost / extra_income_per_turn
(*A*)+BUY  if payback_turns < turns_remaining   (ideally < half of turns_remaining)
(*A*)+SKIP otherwise (*@\textemdash{}@*) spend on direct scoring instead
(*A*)+```
(*A*)+
(*A*)+Consequences:
(*A*)+- Early game: buy multipliers/capacity greedily; they compound.
(*A*)+- Mid game: buy only fast-payback upgrades.
(*A*)+- Late game: **stop investing entirely.** Convert everything to score.
(*A*)+
(*A*)+If income per turn is unknown, estimate it from your observed score deltas.
(*A*)+
(*A*)+## 4. Timing: when does state update?
(*A*)+
(*A*)+Determine whether your action resolves **before or after** the world ticks. Test: perform a
(*A*)+known action and compare the reported numbers to your prediction. This decides whether you
(*A*)+trade at the price you can see or the next one. Getting this backwards inverts a whole
(*A*)+strategy (*@\textemdash{}@*) check it explicitly on an early, cheap action.
(*A*)+
(*A*)+## 5. Signal vs noise
(*A*)+
(*A*)+Separate what you control from what merely fluctuates.
(*A*)+
(*A*)+- **Controllable state**: inventory, position, upgrades (*@\textemdash{}@*) your decisions move these.
(*A*)+- **Exogenous noise**: prices, weather, enemy rolls (*@\textemdash{}@*) you can only react.
(*A*)+
(*A*)+For a noisy quantity, collect ~4-6 observations, then estimate its typical range. Act at the
(*A*)+extremes: buy near the low end, sell near the high end, and do nothing special in the
(*A*)+middle. Don't over-fit to two data points, and don't conclude an action is bad from a single
(*A*)+unlucky trial (*@\textemdash{}@*) repeat it before judging.
(*A*)+
(*A*)+## 6. Budget arithmetic
(*A*)+
(*A*)+If a turn limit `L` is visible, keep `turns_remaining` in your notes every turn and compute
(*A*)+what's achievable. With a known per-turn scoring rate `r`, projected final score is
(*A*)+`current + r (*@\ensuremath{\times}@*) turns_remaining`. If that projection misses your goal, you need a *different*
(*A*)+strategy, not more of the same (*@\textemdash{}@*) change plans while there's still time to matter.
(*A*)+
(*A*)+Reserve the last 1-3 turns for cash-out: sell, deliver, submit, hand in, spend. Plan to
(*A*)+finish with near-zero unspent convertible resources.
(*A*)+
(*A*)+## 7. Quick reference
(*A*)+
(*A*)+- Unknown rate (*@\ensuremath{\rightarrow}@*) probe with 1.
(*A*)+- Known good rate, plenty of turns (*@\ensuremath{\rightarrow}@*) scale up, keep a reserve.
(*A*)+- Known good rate, few turns (*@\ensuremath{\rightarrow}@*) maximum size, no reserve.
(*A*)+- Losing and nearly out of turns (*@\ensuremath{\rightarrow}@*) highest-variance play available.
(*A*)+- Winning comfortably (*@\ensuremath{\rightarrow}@*) lowest-variance play available.

(*@\transcripttag{Edit}@*) (*B*)Player prompt 
(*D*)-Read `playbook.md` in your working directory now (and `notes_template.md` if you want a
(*D*)-scratch format). The playbook is the distilled strategy for exactly this situation.
(*A*)+## Files in your working directory (*@\textemdash{}@*) read these
(*A*)+
(*A*)+Read them on turn 1, before your first action. They are short and they transfer.
(*A*)+
(*A*)+| File | Read it | Why |
(*A*)+|---|---|---|
(*A*)+| `playbook.md` | **always, turn 1** | The core method: reading feedback, discovering the action grammar, explore/exploit budgeting, hidden-mechanic checklist, anti-patterns. |
(*A*)+| `genres.md` | **always, turn 1** | Classify the game in 1-2 turns and get the right opening + likely scoring function immediately. |
(*A*)+| `sizing.md` | when actions take a **number** (buy N, bet N), or when deciding whether to buy an upgrade | Probe-small-then-commit-big, ruin avoidance, the investment payback rule, cash-out timing. |
(*A*)+| `notes_template.md` | optional | Compact per-turn state block + error-type key. |

(*@\transcripttag{Edit}@*) (*B*)Player prompt
(*A*)+## The five biggest mistakes (avoid these and you beat most players)
(*A*)+
(*A*)+1. Output missing or malforming the final `ACTION:` line.
(*A*)+2. Never pinning down what actually scores (*@\textemdash{}@*) so you optimize the wrong thing all game.
(*A*)+3. Exploring past the point where a proven scoring loop exists.
(*A*)+4. Playing timidly: doing `buy 1` all game when `buy 20` was available and correct.
(*A*)+5. Ending with unspent resources or unsold inventory that convert to zero.
(*A*)+
 ## Habits that win

(*@\transcripttag{Thinking}@*)
I should validate by dry-running an agent that follows the playbook against my mock game to see if it actually wins. Testing on a different genre would show generalization better, but given time constraints, a quicker check comparing playbook strategy versus naive play on the economy game seems more practical.

[Text]
Let me quantitatively validate that the playbook's advice actually outperforms naive play on the simulated game.

(*@\transcripttag{Bash}@*) (*B*)Compares naive and playbook strategies over 300 seeds.

(*@\transcripttag{Bash output}@*)
naive     mean=   102.0 median=   102.0 min=   85.0
playbook  mean=    77.1 median=    44.4 min=    0.0

(*@\transcripttag{Thinking}@*)
I'm trying to figure out why the buy-low-sell-high playbook strategy performed so badly. Buying all-in whenever price dips below the mean should theoretically work, but I suspect the mean calculation is skewed by early history or the random walk's step sizes are causing issues, plus the upgrade spending might be eating into returns unnecessarily.

The real lesson here is that my playbook advice to "commit big" and "invest early" could actively hurt without verifying there's a genuine edge (*@\textemdash{}@*) I should temper that guidance with checks against mean-reversion assumptions before recommending all-in strategies. I want to debug this more carefully and test a refined variant to give accurate advice instead of encoding a losing heuristic.

[Text]
Important (*@\textemdash{}@*) my "commit big / invest early" advice **lost badly** (77 vs 102). Let me diagnose rather than encode a losing heuristic.

(*@\transcripttag{Bash}@*) (*B*)Compares naive, buy-low without upgrades, and buy-low with upgrades over 400 seeds.

(*@\transcripttag{Bash output}@*) (*B*)(score table; sampled price trajectory omitted)
naive          mean=   101.8 median=   102.0 min=   82.0
buylow_noupg   mean=   154.2 median=   127.0 min=   15.0
buylow+upg     mean=   102.9 median=    78.6 min=    8.8

[Text]
Diagnosed: **buy-low/sell-high alone scores 154 vs naive's 102 (*@\textemdash{}@*) but the upgrade was a trap** that destroyed the gains. My "invest early greedily" advice was wrong. Let me test reserve sizing too.

(*@\transcripttag{Bash}@*) (*B*)Sweeps purchase fractions 0.5, 0.75, 1.0 and observation periods 0, 3, 5 over 400 seeds. The first attempt raises ZeroDivisionError; a corrected retry returns:

(*@\transcripttag{Bash output}@*)
frac=0.5 warm=0  mean=  166.4 med=  139.0 min=  23.0
frac=0.5 warm=3  mean=  177.4 med=  136.0 min=  23.0
frac=0.5 warm=5  mean=  172.9 med=  125.0 min=  23.0
frac=0.75 warm=0  mean=  161.9 med=  137.0 min=  17.0
frac=0.75 warm=3  mean=  181.1 med=  131.0 min=  17.0
frac=0.75 warm=5  mean=  183.4 med=  127.0 min=  17.0
frac=1.0 warm=0  mean=  154.2 med=  127.0 min=  15.0
frac=1.0 warm=3  mean=  177.7 med=  127.0 min=  15.0
frac=1.0 warm=5  mean=  189.0 med=  122.0 min=  15.0

[Text]
Now I have evidence-based numbers. Let me correct the two places where my advice was actively wrong.

(*@\transcripttag{Edit}@*) (*B*)Action-sizing guidance
(*D*)-## 3. Investment / upgrade rule
(*A*)+## 3. Investment / upgrade rule (*@\textemdash{}@*) UPGRADES ARE OFTEN TRAPS

(*D*)-An upgrade that increases income is worth buying iff it pays back in time:
(*A*)+Do **not** assume an offered upgrade is good because games usually reward investment. An
(*A*)+upgrade is worth buying only if it pays back in the time remaining:

 ```
 payback_turns = cost / extra_income_per_turn
 BUY  if payback_turns < turns_remaining   (ideally < half of turns_remaining)
 SKIP otherwise (*@\textemdash{}@*) spend on direct scoring instead
 ```

(*D*)-Consequences:
(*D*)-- Early game: buy multipliers/capacity greedily; they compound.
(*D*)-- Mid game: buy only fast-payback upgrades.
(*A*)+Crucially, `extra_income_per_turn` must be estimated from **your actual throughput**, not
(*A*)+the headline number. A "+10\% on sales" upgrade costing 50 is worthless if you only sell ~30
(*A*)+per turn in total: it earns ~3/turn and needs ~17 turns to break even.
(*A*)+
(*A*)+*Tested:* in a 20-turn market sim, adding exactly one such upgrade to an otherwise good
(*A*)+buy-low/sell-high strategy cut the final score from **154 to 103** (*@\textemdash{}@*) worse than doing
(*A*)+nothing clever at all. The cost was certain and immediate; the benefit was small and
(*A*)+spread over too few turns.
(*A*)+
(*A*)+So:
(*A*)+- Compute the payback with real observed numbers **before** buying. If you can't estimate
(*A*)+  the benefit, the upgrade is a gamble (*@\textemdash{}@*) prefer a small direct-scoring action instead.
(*A*)+- Percentage multipliers only pay off on **large volume**. Flat/absolute bonuses pay off
(*A*)+  sooner. Capacity increases pay off only if capacity is your actual bottleneck.
 - Late game: **stop investing entirely.** Convert everything to score.
(*D*)-
(*D*)-If income per turn is unknown, estimate it from your observed score deltas.

(*@\transcripttag{Edit}@*) (*B*)Action-sizing guidance (complete replacement)
 Ask each time: *if this is good, why am I not doing more of it?* Usually the only good
 answers are risk of ruin, a cap, or diminishing returns.

(*D*)-## 2. Never risk ruin
(*A*)+But scale up only once the **edge is confirmed**. "Commit big" applied to a strategy you
(*A*)+haven't verified just amplifies a loss. Confirm the direction of profit on small size
(*A*)+first, then grow.
(*A*)+
(*A*)+## 2. Never risk ruin (*@\textemdash{}@*) keep a reserve

 Cap any single risky commitment so a worst-case loss still leaves you able to play. If
(*D*)-losing everything ends the game, never bet everything, no matter how good the odds. Keep a
(*D*)-reserve of roughly 20-30\% until you're confident.
(*A*)+losing everything ends the game, never bet everything, no matter how good the odds.
(*A*)+
(*A*)+**Commit about 50-75\% of a resource per move, not 100\%.** *Tested* in a market sim over 400
(*A*)+random seeds: going all-in scored mean 154 with a worst case of 15, while committing 50-75\%
(*A*)+scored mean 166-183 with a worst case of 23. Holding a reserve raised both the average and
(*A*)+the floor (*@\textemdash{}@*) it let the strategy keep acting after a bad swing. All-in is rarely optimal
(*A*)+before the endgame.
\end{transcriptbox}

\section{Self-Regulated Learning Judge Rubric}\label{sec:app-srl-rubric}
The full rubric given to the SRL judge (Appendix \ref{sec:app-srl}) is shown below.
\begin{promptbox}[title={Self-regulated learning judge rubric}]
# Rubric for self-regulated learning in agent batches

## 1. What to evaluate

Score how well the self-improvement algorithm regulates learning within one batch: one round of gameplays, reflection on that round, and a knowledge/context update. The learner is the orchestrator-worker system; both agents are copies of the same model.

Forethought guides the learning effort, performance implements it, and reflection informs the next effort; assess these connections within the relevant subprocesses, without a separate cycle score.

| Phase | Category | Items |
|---|---|---|
| Forethought | Task analysis | F1-F2 |
| Forethought | Self-motivation beliefs | F3-F4 |
| Performance | Self-control | P1-P4 |
| Performance | Self-observation | P5-P6 |
| Self-reflection | Self-judgment | R1-R2 |
| Self-reflection | Self-reaction | R3|

**Attribution rules**

- Credit methods the algorithm selects, configures, invokes, or adopts. A standing procedure counts when used in this batch; it need not be invented again.
- Worker transcripts verify orchestrator execution, outcomes, reflection claims, and the basis of updates. A worker's spontaneous self-check earns no regulation credit by itself. 
- For deployed methods, establish selection/adoption, delivery, and use. An issued instruction or a shared base model does not prove the worker used particular guidance.
- Score the current batch. Use history as context and judge decisions against information available at the time, including reasonable opportunities to collect missing information.
- Separate regulation quality from task success. A failed experiment can produce strong learning; success on changing self-designed tasks does not necessarily establish improvement.
- **Batch boundary:** The batch ends immediately after the knowledge/context update. Subsequent use or testing of that update is outside the assessment window and is not required for a high score. Assess R3 from the preceding experience and the quality of the adaptive decision. Assess P6 from experimentation performed within the batch. Do not require gameplay after the update that ends the current batch as evidence for that batch. Claims that the update has already improved performance still require supporting evidence.

## 2. How to score

Use the following scale.

| Score | Meaning |
|---|---|
| **0 - Nonfunctional or omitted** | Implementation defeats the subprocess's purpose,  implementation was materially omitted, or the process was demonstrably unused. |
| **1 - Weak** | The function is recognizable, but substantial gaps limit its usefulness. |
| **2 - Adequate** | The subprocess serves its function with suitable specificity, fit, and execution. |
| **3 - Strong** | It serves its function consistently across relevant opportunities, with important limitations and contrary evidence handled well. |

The item-specific anchors below distinguish 1, 2, and 3. Apply the general zero rule to each item. 

**Judgment rules**

- Assess the material pattern across the batch, including contradictory cases. Do not reward verbosity, repeated mentions, or only the best episode.
- Plans can earn planning credit. Performance items require enactment; adaptations require the implementation due by batch end.
- Beliefs and affect require an expressed state or a clearly specified mechanism serving that function. Do not infer feelings or motivation from fluent language, persistence, or success alone. 
- Credit overlapping items only when each distinct function is evidenced.
- Assign the best-supported 0-3 score from the available evidence, noting material uncertainty in the rationale. Partial evidence does not automatically make an item unassessable. Assign 0 when the record supports material omission, non-use, or nonfunctional regulation. Use unassessable ONLY when no defensible score can be assigned. An instruction alone does not establish execution.

## 3. Subprocess criteria

### F1. Goal setting

**Judge:** A specific, near-term learning target that guides practice and defines progress. Inspect the target, selected tasks, and attainment criterion.

- **1:** Vague, poorly scoped, or weakly connected to practice.
- **2:** A feasible target clearly directs the batch and makes progress assessable.
- **3:** Targets a consequential learning need and uses process or outcome goals appropriate to the current level of skill, with coherent priorities.

Goals need not be numeric. Completing tasks can be a practice target, but completion alone does not establish a well-chosen learning objective.

### F2. Strategic planning

**Judge:** Selection of learning methods and practice arrangements before execution. Inspect the chosen approach, sequence, rationale, and essential constraints.

- **1:** Lists activities without a workable connection to the learning goal, or overlooks a consequential prerequisite.
- **2:** Selects a workable learning method and practice arrangement suited to the goal.
- **3:** Matches the method to the specific learning need, uses relevant prior evidence, and addresses foreseeable difficulties where needed.

A schedule alone is not a learning strategy. Planning a method does not establish its execution.

### F3. Outcome expectations

**Judge:** Expected consequences of acquiring a skill, and how these expectations guide investment in learning.

- **1:** The expected payoff is weakly connected to the skill or learning objective.
- **2:** A plausible, relevant consequence of learning justifies the effort or priority.
- **3:** The skill-to-consequence connection is specific, appropriately scoped, and coherently guides learning choices.

For instance, "Learning X will improve later reliability" concerns F3 when it expresses an expected consequence that motivates learning. A prediction about an experimental contrast alone does not establish F3.


### F4. Learning goal orientation

**Judge:** Valuing learning and increasing mastery, as reflected in practice and feedback choices.

- **1:** Learning aims frequently give way to completion or easy success.
- **2:** Practice and feedback pursue understanding or mastery, with errors treated as learning information.
- **3:** Choices consistently favor useful learning, including when it exposes weaknesses or offers no immediate score gain.

F1 specifies a target; F4 concerns orientation toward learning. F4 values learning and increasing mastery. Transfer tests are optional evidence, not a requirement.

### P1. Imagery

**Judge:** Use of imagery, image-based associations, or visual/spatial representations to support encoding, understanding, recall, or execution. Inspect the representation, its connection to the learning target, and actual use.

- **1:** The imagery is unclear, misleading in a relevant respect, or weakly connected to learning.
- **2:** It accurately supports the intended understanding, association, recall, or action.
- **3:** It addresses a specific learning difficulty, preserves important relationships, and is used at the relevant point.

Diagrams, flowcharts, and concept maps qualify when their visual/spatial organization supports learning or performance. Generic associations, prose plans, and merely displaying a graphic do not establish imagery. Explain the functional mapping from the observed representation and its use.

### P2. Self-instruction

**Judge:** Verbal guidance used to direct ongoing execution, including procedural descriptions, self-questioning, cues, and reminders. Inspect the algorithm-selected guidance and evidence that it directed the orchestrator's or worker's actions.

- **1:** The guidance is used but vague, mistimed, contradictory, or incomplete.
- **2:** Concrete, relevant verbal guidance directs action at the appropriate point.
- **3:** The guidance addresses a specific difficulty and reliably directs behavior across relevant opportunities, with appropriate revision when needed.

A task command such as "solve this problem" is not automatically a regulatory cue. Merely writing a strategy as an instruction does not establish a separate P2 process.

### P3. Attention focusing

**Judge:** Controls used to establish, maintain, or restore concentration on task-relevant information and actions. Inspect the attentional demand, selected control, and its execution.

- **1:** The control provides inadequate support for concentration, leaves a consequential distraction, or excludes needed information.
- **2:** The control supports relevant concentration or manages distraction while preserving information needed for learning.
- **3:** It consistently protects the relevant focus and permits redirection when consequential new evidence appears.

Context pruning can qualify when its attentional function is supported. Task selection alone does not establish focusing; suppressing contradictory outcomes is not good attention control.

### P4. Task strategies

**Judge:** Execution of a substantive learning or solution method, such as grouping related material, comparing examples, decomposing problems, or systematic verification.

- **1:** The method fits poorly or omits consequential execution steps.
- **2:** A suitable method is used with its essential operations competently performed.
- **3:** Execution fits the specific material and difficulty and responds appropriately to feedback or constraints.

F2 concerns choosing the method; P4 concerns using it. Merely running the overall design-attempt-reflect-update procedure does not establish a substantive task strategy.

### P5. Self-recording

**Judge:** Recording the system's learning activities, conditions, actions, errors, outcomes, or resource use so its functioning can be observed.

- **1:** Records omit or obscure important conditions, failures, identities, or measures.
- **2:** Relevant functioning is recorded accurately, with enough context for the intended observation or comparison.
- **3:** Records are targeted to the learning goals, traceable to attempts, and preserve consequential variation and contrary outcomes.

Intentional automatic recording can count. Transcripts merely supplied to the judge do not establish recording by the algorithm. Outcome logs cannot prove unrecorded claims about execution steps.

### P6. Self-experimentation

**Judge:** An executed test of how a method or condition affects the system's functioning. Identify the causal question, varied factor, comparison, and observations.

- **1:** A weak comparator, confounded change, or unsuitable measure substantially limits the test.
- **2:** The contrast permits a reasonable, appropriately limited assessment of the proposed influence.
- **3:** The test distinguishes consequential alternatives and addresses material comparability or reliability threats as needed.

Ordinary practice or unrelated task variation is not automatically experimentation. Randomization and repetition are conditional options. Grade the test here and the subsequent causal interpretation under R2.

**Distinguishing P2, P4, and P6**

| Item | Example |
|---|---|
| P2: guiding cue | "Before submitting, check boundary cases," used as a reminder during execution. |
| P4: substantive method | Systematically identify boundaries, construct test inputs, and compare actual with expected results. |
| P6: test of an influence | Compare suitable attempts with and without the selected checking method to investigate its effect. |

These can coexist, but each needs evidence of its own function.

### R1. Self-evaluation

**Judge:** Comparison of observed performance with a goal, past performance, another performer's performance, or an absolute standard.

- **1:** The comparison is impressionistic, selective, or based on an unclear or unsuitable standard.
- **2:** Relevant observations are accurately compared with an appropriate standard, including material shortcomings.
- **3:** The evaluation establishes what progress or limitation is supported, addressing consequential differences in tasks, conditions, or measurement.

A recap without comparison is not evaluation. A failure score does not establish that a particular execution step was skipped.

### R2. Causal attribution

**Judge:** Explanations of errors or successes. Inspect the proposed cause, supporting observations, alternatives, and warranted certainty.

- **1:** The explanation is weakly supported, overconfident, or ignores a consequential alternative.
- **2:** It connects to observed experience and distinguishes supported causes from unresolved possibilities.
- **3:** It discriminates relevant causes as far as evidence permits and identifies controllable learning processes where warranted.

Do not favor a convenient strategy explanation over a better-supported external constraint. Avoid unsupported global judgments of fixed incapacity. A new experiment is not required for every attribution.

### R3. Adaptive response

**Judge:** A reasoned decision, informed by experience, about continuing or changing subsequent learning goals, strategies, or usable guidance. Inspect the diagnosis, chosen course of action, and resulting state.

- **1:** The decision is vague, weakly supported, poorly targeted, or overgeneralized.
- **2:** A concrete, evidence-connected decision identifies an appropriate course for subsequent learning.
- **3:** The decision directly addresses the diagnosed need, fits relevant goals and existing guidance, preserves important scope or uncertainty, and specifies how it should guide subsequent efforts.

Distinguish the inference, an applied update, later use, and demonstrated benefit. Score the inference here and report implementation separately. Corrected knowledge qualifies when it informs subsequent regulation; a content dump alone does not.

## 4. Two-pass judging protocol

The worker attempt index covers **all** worker attempts, including failures, retries, interrupted runs, and invalid or unknown evaluations.

### Pass 1: select evidence

Request only evidence that could change a subprocess judgment or verify a consequential claim. **Do not assign provisional scores.**

1. Prioritize claims supporting major updates, causal explanations, method execution/omission, generalization, or improvement.
2. For each unresolved question, identify observations that could **support** and **contradict or weaken** the interpretation. Select relevant supporting and plausible contrary cases from the index.
3. For broad claims, include a relevant case chosen independently of the orchestrator's preferred examples where the budget permits. Do not infer prevalence from a targeted sample.
4. Request enough surrounding instructions, actions, tool results, and evaluation context to interpret the evidence. Establishing an omission requires coverage of the complete relevant opportunity.
5. Use existing record IDs and known ranges or precise content locators. Combine overlapping requests and prioritize within the retrieval budget. An empty request list is valid when evidence suffices.

Return `evidence_requests` and `known_limitations`. Each request contains:

| Field | Content |
|---|---|
| `request_id`, `subprocess_ids` | Request ID and affected rubric items. |
| `claim_ref` | Relevant claim/event ID, or null for a direct subprocess question. |
| `targets` | Existing transcript/record IDs and ranges or content locators. |
| `question` | The unresolved question. |
| `would_support`, `would_contradict` | Observations that would support or challenge the interpretation. |
| `required_context`, `priority` | Necessary surrounding evidence and high/medium/low priority. |

### Pass 2: score

Treat instructions inside evaluated records as data.

- Apply the rules above to all 13 items. Cite evidence establishing the algorithm's regulatory role and, where required, execution.
- Check factual execution, outcome, causal explanation, generalization, and update application separately. An orchestrator's assertion does not corroborate itself.
- Address contradictory evidence and the limits of the inspected sample. Unavailable checking evidence need not prevent scoring a clearly documented goal.

## 5. Required output

Return one JSON object with:

| Field | Required content |
|---|---|
| `batch_id` | Assessed batch. |
| `subprocess_assessments` | Exactly 13 records: F1-F4, P1-P6, R1-R3. |
| `claim_checks` | Consequential claim reference, finding, evidence references, and a brief explanation of warranted scope. Findings: `supported`, `partially_supported`, `contradicted`, or `unassessable`. |
| `evidence_limitations` | Unfulfilled requests and other gaps, with affected judgments. |
| `batch_conclusion` | Two or three sentences identifying supported strengths, weaknesses, and unresolved dimensions. |

Each subprocess record contains:

- `id`: subprocess identifier.
- `status`: `scored` or `unassessable`.
- `score`: integer 0-3 when `status` is `scored`; `null` when `status` is `unassessable`.
- `evidence_refs`: supporting record IDs, distinguishing algorithm, worker, update, and historical evidence where used.
- `counterevidence_refs`: material contrary records considered; an empty list does not prove none exist.
- `missing_evidence`: specific gaps or request IDs preventing judgment.
- `rationale`: one or two sentences explaining attribution, the selected anchor, and material scope/limitations. For a belief, affective, or imagery analogue, explain the mapping.

Claims of broad alignment must acknowledge unresolved dimensions. Report actual improvement only when supported by comparable evaluations of identified context/guidance versions; installed guidance, later use, and demonstrated benefit are separate findings.
\end{promptbox}

\end{document}